# Reliable Off-policy Evaluation for Reinforcement Learning

Jie Wang
School of Science and Engineering, The Chinese University of Hong Kong, Shenzhen
jiewang@link.cuhk.edu.cn

Rui Gao
Department of Information, Risk and Operations Management, The University of Texas at Austin
rui.gao@mccombs.utexas.edu

Hongyuan Zha
School of Data Science, Shenzhen Institute of Artificial Intelligence and Robotics for Society, The Chinese University of Hong Kong, Shenzhen
zhahy@cuhk.edu.cn

In a sequential decision-making problem, off-policy evaluation estimates the expected cumulative reward of a target policy using logged trajectory data generated from a different behavior policy, without execution of the target policy. Reinforcement learning in high-stake environments, such as healthcare and education, is often limited to off-policy settings due to safety or ethical concerns, or inability of exploration. Hence it is imperative to quantify the uncertainty of the off-policy estimate before deployment of the target policy. In this paper, we propose a novel framework that provides robust and optimistic cumulative reward estimates using one or multiple logged trajectories data. Leveraging methodologies from distributionally robust optimization, we show that with proper selection of the size of the distributional uncertainty set, these estimates serve as confidence bounds with non-asymptotic and asymptotic guarantees under stochastic or adversarial environments. Our results are also generalized to batch reinforcement learning and are supported by empirical analysis.

## 1. Introduction

Reinforcement learning (RL) has achieved phenomenal success in games and robotics (OpenAI et al. 2019, Mnih et al. 2015, Kober et al. 2013) in the past decade, which also stimulates the enthusiasm of extending these techniques in other areas including healthcare (Raghu et al. 2017, Gottesman et al. 2019), education (Mandel et al. 2014), autonomous driving (Sallab et al. 2017), recommendation systems (Liu et al. 2018a, Wang et al. 2018), etc. One of the major challenges in applying RL to these real-world applications, especially those involve high-stake environments, is the problem of *off-policy evaluation* (OPE): how one can evaluate a new policy before deployment, using only historical data collected from a different policy, known as the behavior policy. Indeed, for many practical applications, one may not have a faithful simulator of the domain from which sufficient amount of data can be exploited to train the RL system, and it may not always be feasible to try out a new policy without causing unintended harms. For example, consider the problem of finding the best treatment plan for a patient, or testing the performance of an automated driving system, or suggesting a personalized curriculum for a student. In those tasks, conducting experimentation involves interactions with real people, thus it can be costly to collect data. Even worse, a bad policy can be risky or unethical and may result in severe consequences. Therefore, it is important for the RL system to have the ability to predict how well a new policy would perform without having to deploy it first.

While most existing works on OPE aim to provide accurate point estimates for short-horizon problems (Thomas et al. 2017, Precup 2000, Hanna et al. 2017, Jiang and Li 2016) as well as long- or infinite-horizon problems (Liu et al. 2018b, Zhang et al. 2020a, Tang et al. 2019, Farajtabar et al. 2018, Kallus and Uehara 2020, Chen et al. 2020b), it is equally important to quantify the uncertainty of the OPE point estimates for both safe exploration and optimistic planning. On the one hand, in high-stake mission-critical environments as aforementioned, providing a lower confidence bound for OPE enables us to explore policies with safety guarantees and thus help to reduce the risk and circumvent catastrophic events; specifically, in batch reinforcement learning in which a fixed set of logged data is used for policy optimization, this pessimism principle is important to guarantee good performance (Buckman et al. 2021, Jin et al. 2021). On the other hand, for risk-seeking policy optimization algorithms that apply the optimism principle in the face of uncertainty (Munos 2014), an upper confidence bound used to balance the exploration and exploitation trade-off is desirable. Motivated by these problems, the main goal of this paper is to develop reliable confidence interval (CI) estimates for OPE with provable statistical guarantees.

Let us discuss several challenges in deriving a CI for OPE. Arguably, the most straightforward thought is based on the sample mean and sample deviation estimates and invokes some form of concentration inequality to implement. Unfortunately, this could be problematic due to various reasons.

(I) To begin with, most existing approaches (Thomas et al. 2015, Hanna et al. 2017) are based on step-wise importance sampling, which do not work in long- or infinite-horizon setting. This is because the OPE point estimate suffers from the "curse of horizon" (Liu et al. 2018b): it has excessively high variance arising from the multiplication of importance ratios at each time period in the horizon.

(II) To reduce the variance, one may consider approaches based on a marginalized formulation (Liu et al. 2018b) that applies importance sampling to the average state visitation distribution. Essentially this yields a bilevel stochastic optimization (feasibility) problem (see also equations (2)(3) in Section 2), in which the lower-level problem estimates the marginalized importance weight function and the upper-level problem estimates the cumulative discounted reward. Developing a CI for its optimal value is a highly non-trivial task.

(III) Even for asymptotic CIs that are usually easier to obtain than non-asymptotic CIs, existing concentration results for the marginalized formulation (e.g. Kallus and Uehara (2022)) depend crucially on certain mixing conditions, i.e., the finite-sample distribution of the associated Markov chain should be close to its steady-state distribution. For non-asymptotic CIs, Feng et al. (2020) proposes a variational framework that solves an optimization problem over a confidence set containing the underlying true state-action value function with high probability, but assumes the access to i.i.d. state-action transition pairs, and the length of their CI depends on the sample size in a sub-optimal (fourth-root) rate; Dai et al. (2020) applies the generalized empirical likelihood method to the Lagrangian reformulation of the bilevel stochastic program, which assumes i.i.d. state-action-next-state transition tuples (and may be relaxed to certain fast-mixing condition) and involves a nonconvex-concave saddle-point problem that may not have global optimality guarantees. Unfortunately, the i.i.d. or mixing conditions aforementioned generally do not hold in practice. Indeed, the historical trajectory data may not serve as a faithful representation of the steady-state distribution under the behavior policy, because the number of trajectories in the data set is usually limited or even only one, each of which has dependent data sequence with finite length and thus may not be mixing yet.

(IV) In OPE, the environment where the target policy will be deployed may deviate from the past environment where the historical data were collected. For instance, in clinical trials, a medicine or vaccine is initially tested for young and healthy people, but eventually we would like to know its safety and effectiveness on other population such as the elderly people. As such, there is a shift in the age distribution between the training population and the testing population. This is usually referred to as covariate shift or distribution shift in the literature (Si et al. 2020, Uehara et al. 2020b). In this case, it is important to evaluate the risk of a new policy under adversarial scenarios so that the decision-maker would have a sense of how worst it could happen and makes plans accordingly.

To take these concerns into account, we propose a novel framework for computing CI estimates for the infinite-horizon discounted OPE, inspired from distributionally robust Markov Decision Process (MDP) (Iyengar 2005, Nilim and Ghaoui 2005, Wiesemann et al. 2013). In a nutshell, the idea is to develop a distributionally robust/optimistic counterpart of the marginalized formulation as mentioned in the item (II) above (see formulation (P) in Section 3), whose optimal values serve as the lower/upper confidence bounds for the OPE under proper selection of the radius of the uncertainty set. This gives rise to an end-to-end framework that uses logged trajectory data to simultaneously learn the importance weight function, find the worst-case/best-case scenarios, and compute the CI. Specifically, we consider an s-rectangular Wasserstein distributional uncertainty set (Yang 2017) centered around the empirical conditional action-next-state distributions under the behavior policy, which captures the distributional uncertainty (e.g. adversarial data perturbation and distribution shift) of the average visitation distribution, and naturally incorporates the geometry of the state-action space and is suitable for distributions with non-overlapping support. Our main contributions are as follows.

- We derive exact tractable reformulations for the distributionally robust and optimistic OPE (Theorem 1), based on which the CIs can be computed via robust value iteration algorithm (Algorithm 1). We also develop an equivalent Lagrangian formulation that can be solved numerically in a fashion similar to generative adversarial networks (Goodfellow et al. 2014), in which the discriminator is regularized by its Lipschitz norm (Proposition 3).
- In stochastic setting, we develop both asymptotic and non-asymptotic CIs for OPE (Theorem 2 and Remark 1), only assuming that the logged data are collected from (one or more) trajectories under the behavior policy and the underlying Markov chain transition dynamics. In adversarial setting where there is a distribution shift due to changing environments, we provide asymptotic and non-asymptotic CIs for the adversarial reward (Theorems 3 and 4). When applying to on-policy problems, our results provide an end-to-end statistical inference approach for robust MDP that directly uses trajectory data without estimating the transition probability matrix.
- We extend our framework to batch reinforcement learning by developing efficient algorithms and provide finite-sample performance guarantees (Theorems 5 and 6). Our theoretical findings are verified by numerical experiments.
- Our analysis is based on two new results on Wasserstein distributionally robust optimization in discrete finite space, both of which may be of independent interest. The first result is on its equivalence to a discrete form of Lipschitz regularization; and the second result is a new finite-sample guarantee that does not suffer from the curse of dimensionality, which, to the best of our knowledge, is the first result of this kind in discrete settings.

## Related Work

**Uncertainty Quantification for OPE.** Recently, there is a surge of interest in studying uncertainty quantification for OPE. Existing works (Feng et al. 2020, Dai et al. 2020, Kallus and Uehara 2022) assume i.i.d. transitions pairs or mixing condition. The non-asymptotic CI in Dai et al. (2020) is computed from solving a nonconvex-concave optimization involving an $f$-divergence distributional uncertainty set. The non-asymptotic CI derived in Feng et al. (2020) exploits concentration bounds for U/V-statistics, whose length of the CI depends on the sample size in a sub-optimal fourth-root rate. Asymptotic CIs are developed in Kallus and Uehara (2022) for a broad class of MDPs using central limit theorem under various mixing conditions. In addition, by assuming the value function can be approximated by linear functions under certain basis, non-asymptotic and asymptotic CIs are constructed in Duan et al. (2020), Shi et al. (2020), but this approach suffers from model misspecification and may lead to biased estimates. Asymptotic CIs are also derived using bootstrapping (Kostrikov and Nachum 2020) and Bayesian hypothesis testing (Sonabend et al. 2020). Besides, Jiang and Huang (2020) considers a different notion of the CI that does not capture the randomness of data. For finite, short-horizon problems, CIs are proposed based on concentration inequalities (Thomas et al. 2015) and bootstraping (Hanna et al. 2017). However, those bounds become vacuous due to the large variance of estimators in long- or infinite-horizon problems (Liu et al. 2018b). In all above works, it is assumed that the deployment environment remains the same. OPE for bandit learning under distribution shift is investigated in Si et al. (2020).

**(Distributionally) Robust MDP and RL.** Our framework is closely related to robust MDPs and its distributionally robust counterpart and applications in RL. Robust MDPs take account of the uncertainty in transition dynamics by hedging against a family of transition probability matrices specifying the range of transition probabilities, and rectangularity is a useful property to maintain tractability of the resulting optimization problem (Iyengar 2005, Nilim and El Ghaoui 2005, Wiesemann et al. 2013, Mannor et al. 2016, Goyal and Grand-Clement 2022). Distributionally robust MDP exploits a priori distributional information to construct the distributional uncertainty set of transition probability distributions (Xu and Mannor 2010). Different distributional uncertainty sets of transition dynamics have been studied, including the set constructed based on relative entropy (Smirnova et al. 2019), Wasserstein metric (Yang 2017, Tirinzoni et al. 2018, Abdullah et al. 2019, Hou et al. 2020, Song and Zhao 2020), $L_1$-norm (Ho et al. 2021), or general statistical distances together with certain moment conditions (Chen et al. 2019b). Most of these works do not consider off-policy evaluations with exceptions of Tirinzoni et al. (2018) and Petrik and Russel (2019), which consider entropy-based and Bayesian uncertainty sets, respectively. The idea of using distributionally robust optimization for uncertainty quantification has appeared in the context of simulation optimization (Lam and Zhou 2017, Lam and Qian 2017). In RL, pessimistic MDPs based on offline data are considered in Kidambi et al. (2020), Yu et al. (2020), Matsushima et al. (2021), but the way of uncertainty quantification in these works are orthogonal to our work. Our framework is consistent to the pessimism principle introduced in Buckman et al. (2021), but our bound is data-dependent and thus tighter than their state-wise bound.

The rest of this paper is organized as follows. Section 2 presents preliminaries on OPE and robust MDPs. Section 3 outlines our framework for robust and optimistic off-policy evaluation for MDPs with discounted reward. Section 4 derives tractable reformulations and develops algorithms. Section 5 presents the theoretical analysis of the optimistic and robust reward evaluation. Section 6 extends the algorithm to robust batch reinforcement learning, and Section 7 demonstrates some numerical experiments and analysis. The Appendices contain the proofs of all the results.

## 2. Off-Policy Evaluation

We consider an infinite-horizon discounted finite state MDP represented by a tuple $\langle \mathcal{S}, \mathcal{A}, P, R, \gamma, d_0 \rangle$, where $\mathcal{S}, \mathcal{A}$ denote, respectively, the state and the action spaces, which are assumed to be finite; $P = \{P(s'|s,a)\}_{s,s' \in \mathcal{S}, a \in \mathcal{A}}$ is the set of transition probability matrices, where $P(s'|s,a)$ represents the state transition probability from the current state $s$ to the next state $s'$ after taking an action $a$; $R = \{r(s,a)\}_{s \in \mathcal{S}, a \in \mathcal{A}}$ is the reward table with the $(s,a)$-th entry being the reward after taking the action $a$ in the state $s$; $\gamma \in (0,1)$ is the discount factor; and $d_0$ denotes the initial state distribution before executing a policy. A stochastic policy $\pi$ is represented by a conditional distribution that takes action $a$ with probability $\pi(a|s)$ in state $s$. At each time $t$, a decision maker observes the current state $s_t$, takes an action $a_t$ according to a policy $\pi(\cdot|s_t)$, receives a non-negative reward $r_t$ whose expectation is $r(s_t, a_t)$, and transit to the next state $s_{t+1}$ according to transition probability $P(s_{t+1}|s_t, a_t)$. The performance of a policy $\pi$ is measured by the expected discounted cumulative reward $R_\pi$, or its *value*, defined as

$$R_\pi := (1-\gamma) \lim_{T \to \infty} \mathbb{E}\Big[ \sum_{t=0}^{T} \gamma^t r_t \Big],$$

where the expectation is taken with respect to the distribution of the trajectories under the policy $\pi$.

Off-policy evaluation is the problem of estimating $R_\pi$ for a new *target policy* $\pi$ using a set of trajectory data collected from a *behavior policy* $\pi_b$. As is well-known, OPE with long- or infinite-horizon MDPs suffers from the curse of horizon: the variance of the importance sampling-based estimates grows exponentially with respect to the length of the horizon $T$. To address this issue, one may consider an alternative formulation of the target policy value based on the *marginalized importance sampling* of the average visitation distribution of state-action pairs (Liu et al. 2018b). To this end, let us provide an alternative representation of the value $R_\pi$. Let $d_{\pi,t}$ be the distribution of the state $s_t$ at time $t$ when executing a policy $\pi$ with initial distribution $d_0$, and define the *average visitation distribution* as

$$d_\pi(s) := (1-\gamma) \lim_{T\to\infty} \sum_{t=0}^{T} \gamma^t d_{\pi,t}(s), \quad s \in \mathcal{S}, \tag{1}$$

which becomes the steady-state distribution under the policy $\pi$ when $\gamma \to 1$. By making use of (1), the value $R_\pi$ can be expressed in the expectation form

$$R_\pi = \mathbb{E}_{(s,a)\sim d_\pi} [r(s,a)] = \sum_{s\in\mathcal{S},a\in\mathcal{A}} d_\pi(s)\pi(a \mid s)r(s,a),$$

where we have used overloaded notation $d_\pi(s,a) := d_\pi(s)\pi(a \mid s)$. Now define $\beta : \mathcal{S}\times\mathcal{A} \to \mathbb{R}_+$ as

$$\beta_s(a) := \frac{\pi(a \mid s)}{\pi_b(a \mid s)}, \quad s \in \mathcal{S}, a \in \mathcal{A},$$

which is assumed to be known or can be estimated accurately from data. Throughout this paper, we make the following assumptions, which are standard in the literature.

Assumption 1. *The Markov chains induced by $\pi$ and $\pi_b$ are ergodic.*

Assumption 2. $d_{\pi_b}(s)$, $\beta_s(a) > 0$ *for all* $s \in \mathcal{S}, a \in \mathcal{A}$.

Define the *marginalized importance ratio* $w : \mathcal{S} \to \mathbb{R}_+$ as

$$w(s) := \frac{d_\pi(s)}{d_{\pi_b}(s)}, \quad s \in \mathcal{S},$$

to be the density ratio between the average visitation state distributions of the target policy and that of the behavior policy. By the importance sampling technique, the value $R_\pi$ for the target policy $\pi$ can be computed as

$$R_\pi = \mathbb{E}_{(s,a)\sim d_{\pi_b}} \left[ w(s)\beta_s(a)r(s,a) \right]. \tag{2}$$

Thereby, the problem of OPE is transformed into estimating the marginalized importance ratio – also known as stationary distribution correction (Nachum et al. 2019) – with a plethora of approaches developed covering various settings (Xie et al. 2019, Kallus and Uehara 2020, 2022, Uehara et al. 2020a, Tang et al. 2019, Chen et al. 2020b, Zhang et al. 2020a).

With a slight abuse of notation, we use $d_\pi(s,a,s')$ to denote the average visitation probability for the state-action-state pair $(s,a,s')$ of a policy $\pi$, i.e., $d_\pi(s,a,s') = d_\pi(s)\pi(a \mid s)P(s' \mid s,a)$. Using the stationary equation under policy $\pi$, it can be easily shown that $w$ and $\beta$ satisfy the following system of stationary equations, whose proof can be found in Appendix EC.2.

Lemma 1. *Let $\gamma \in (0,1]$. Then it holds that*

$$w(s')d_{\pi_b}(s') = (1-\gamma)d_0(s') + \gamma \sum_{s\in\mathcal{S},a\in\mathcal{A}} d_{\pi_b}(s,a,s')\beta_s(a)w(s), \quad \forall s' \in \mathcal{S}. \tag{3}$$

This lemma helps to develop an estimation of $w$ using the logged trajectory data. Indeed, $d_{\pi_b}$ can be estimated directly from the trajectories, thereby $w$ can be obtained by solving (3), and thus the value $R_\pi$ can be computed using (2). Typically in the literature, it is assumed that the empirical estimation of $d_{\pi_b}$ is a good surrogate. However in reality, the empirical estimate from the trajectory may deviate from the deploying environment since we only have access to trajectories with limited length and may face the issue of changing environments. The rest of this paper is devoted to addressing these issues.

## 3. Distributionally Robust and Optimistic Off-Policy Evaluation

As previously mentioned, the finite-length Markovian trajectory data and potential shifts of MDP environments may both lead to potential estimation error in the importance ratio $w$ and thus the value $R_\pi$. In this section, we develop a distributionally robust/optimistic framework that takes these issues into account.

We make the following assumption on the logged trajectory data generation process throughout the remainder of this paper.

ASSUMPTION 3. *The trajectories are generated according to the behavior policy $\pi_b$ and the transition dynamics $P$. Namely, given current state $s$, the action is generated according to $\pi_b(\cdot|s)$; and given the current state $s$ and action $a$, the next state is generated independently according to $P(\cdot|s,a)$.*

Note that this is a rather mild assumption requiring nothing more than that the trajectories are consistent with the MDP environment under policy $\pi_b$. This implies that conditioning on the current state $s$, we can extract i.i.d. samples $(a,s')$ from the conditional distribution $\pi_b(a|s)P(s'|a,s)$ from the trajectories. This is a much weaker assumption than the common i.i.d. assumption of $(s,a,s')$ in the existing literature that is usually not satisfied by the logged trajectories.

Leveraging ideas from distributionally robust MDP (Wiesemann et al. 2013, Yang 2017), we consider an s-rectangular Wasserstein distributional uncertainty set. Denote by $W$ the Wasserstein metric associated with a metric (transport cost function) $c:(\mathcal{S}\times\mathcal{A})^2\to\mathbb{R}_+$:

$$W(\mu,\nu):=\min_{\gamma\in\Gamma(\mu,\nu)}\mathbb{E}_{((a,s),(a',s'))\sim\gamma}\Big[c\Big((a,s),(a',s')\Big)\Big],$$

with $\Gamma(\mu,\nu)$ represents the joint distribution on $(\mathcal{A}\times\mathcal{S})^2$ with marginals $\mu$ and $\nu$. For any probability distribution $\mu$ on the state-action-next-state space $\mathcal{S}\times\mathcal{A}\times\mathcal{S}$, let $\mu(\cdot,\cdot|s)$ be the conditional probability distribution on the action-next-state space $\mathcal{A}\times\mathcal{S}$ conditioning on the state $s$ induced from $\mu$. Denote by $\mathcal{D}=\{(s_t^j,a_t^j,r_t^j)\}_{1\le t\le T_j,1\le j\le J}$ the collected samples under the behavior policy $\pi_b$, where $\mathcal{D}$ contains $J$ trajectories with the $j$-th trajectory being $(s_1^j,a_1^j,r_1^j,\ldots,s_{T_j}^j,a_{T_j}^j,r_T^j,s_{T_j+1}^j)$. Let $\rho=(\rho_s)_{s\in\mathcal{S}}$, where $\rho_s$ is the radius for the Wasserstein ball associated with state $s$, and let $\hat{\mu}=(\hat{\mu}_s)_{s\in\mathcal{S}}$, where $\hat{\mu}_s$ is the empirical distributions of the conditional action-next-state visitation distributions $d_{\pi_b}(\cdot,\cdot\mid s)$ constructed from tuples $\mathcal{D}$:

$$\hat{\mu}_s:=\frac{1}{n_s}\sum_{1\le t\le T_j,1\le j\le J}\delta_{(a_t^j,s_{t+1}^j)}1\{(s,a_t^j,r_t^j)\in\mathcal{D}\},$$

and $n_s=\sum_{1\le t\le T_j,1\le j\le J}1\{(s,a_t^j,r_t^j)\in\mathcal{D}\}$ denotes the number of state-action-next-state tuples starting with $s$. It is easy to verify that $\sum_{s\in\mathcal{S}}n_s=\sum_{1\le j\le J}T_j$. The s-rectangular Wasserstein distributional uncertainty set is defined as

$$\mathfrak{M}_{\hat{\mu}}(\rho):=\Big\{\mu\in\mathcal{P}(\mathcal{S}\times\mathcal{A}\times\mathcal{S}):\ W\big(\mu(\cdot,\cdot\mid s),\hat{\mu}_s\big)\le\rho_s,\forall s\in\mathcal{S}\Big\}.$$

We propose the following distributionally robust and optimistic formulations, respectively:

$$\mathcal{L}_{\hat{\mu}}(\rho)\Big(\text{resp. }\mathcal{U}_{\hat{\mu}}(\rho)\Big):=\min_{w,\mu}\Big(\text{resp. }\max_{w,\mu}\Big)\sum_{s\in\mathcal{S},a\in\mathcal{A}}\mu(s)\pi_b(a\mid s)w(s)\beta_s(a)r(s,a) \tag{P-a}$$

$$\text{s.t. }\ w(s')\mu(s')=(1-\gamma)d_0(s')+\gamma\sum_{s\in\mathcal{S},a\in\mathcal{A}}\mu(s,a,s')\beta_s(a)w(s),\ \ \forall s'\in\mathcal{S}, \tag{P-b}$$

$$\mu\in\mathfrak{M}_{\hat{\mu}}(\rho),\ w\in\mathbb{R}_+^{|\mathcal{S}|}. \tag{P-c}$$

Via the system of stationary equations (P-b), every $\mu$ determines a set of compatible marginalized importance ratio functions. In particular, the true average visitation distribution under the behavior policy $d_{\pi_b}$ satisfies (P-b) with $\mu = d_{\pi_b}$. When $\rho_s = 0$ for all $s \in \mathcal{S}$, the optimization problem (P) reduces to the typical sample average formulation studied in the literature (Liu et al. 2018b). Otherwise, the distributional uncertainty set $\mathfrak{M}_{\hat{\mu}}(\rho)$ induces an uncertainty set $\{w : \exists \mu \in \mathfrak{M}_{\hat{\mu}}(\rho),\ s.t.,\ \text{(P-b) holds}\}$ for the importance weight function $w$, which can be viewed as a confidence region for $w$. Under proper selection of the radius $\{\rho_s\}_{s\in\mathcal{S}}$ that will be specified in Section 5, the optimal values of (P) provides a CI estimate $[\mathcal{L}_{\hat{\mu}}(\rho), \mathcal{U}_{\hat{\mu}}(\rho)]$ for $R_\pi$. This framework is an end-to-end approach in the sense that it jointly learns the importance weight function, the worst-case/best-case distribution and computing the CI using the logged trajectory data, as opposed to a separate approach that builds the CI based on the estimated importance weight function $w$ using a separate procedure (Mousavi et al. 2020), whereby the estimation error of $w$ may propagate to the CI resulting a larger variance.

The advantage of using the s-rectangular Wasserstein distributional uncertainty set is four-fold. First, Wasserstein metric naturally incorporates the geometry of the state-action space, and is suitable for distributions with non-overlapping supports and for hedging against adversarial data perturbations (Gao and Kleywegt 2022). It is purely data-driven whereby $\hat{\mu}$ can be directly constructed from the logged trajectories, as opposed to the classical robust MDP literature (see, e.g. Wiesemann et al. (2013)) in which the nominal transition dynamics are estimated from data using a separate statistical procedure such as maximum likelihood. Second, from the optimization point of view, the s-rectangularity enables a tractable reformulation of (P) that will be presented in Section 4. Third, from the statistical point of view, the s-rectangularity facilitates us to establish a confidence interval with provable guarantees by observing that for any trajectory data, conditioning on the current state $s$, we have i.i.d. samples of action-next-state pairs $(a, s')$ as long as the trajectories are generated according to the behavior policy $\pi_b(a|s)$ and the transition dynamics $P(s'|s,a)$; see Assumption 3. Fourth, for batch RL setting in Section 6, the $s$-rectanguarity is consistent with the pessimism principle with respect to the state-wise Bellman uncertainty that is introduced in Buckman et al. (2021).

## 4. Tractable Reformulations and Algorithms

Problem (P) is not immediately tractable because it involves an optimization over probability distributions as well as a simultaneous optimization over $w$ and $\mu$. In this section, we provide exact tractable reformulations and algorithms for solving (P), whose proofs are provided in Appendix EC.3.

To simplify the presentation, we state only the results on the distributionally robust OPE, i.e., the minimization problem in (P), and the counterpart for distributionally optimistic OPE can be found in Appendix A. For a function $f : \mathcal{S} \to \mathbb{R}$, we define the *global slope* (Ambrosio et al. 2008) of $f$ at $s \in \mathcal{S}$ as

$$\mathrm{I}_f(s) = \max_{\tilde{s} \neq s} \frac{f(\tilde{s}) - f(s)}{c(\tilde{s}, s)}, \tag{4}$$

and for any probability measure $\nu \in \mathcal{P}(\mathcal{S})$ we define the Lipschitz norm of $f$ with respect to $\nu$ as

$$\|f\|_{\mathrm{Lip},\nu} = \max_{s \in \mathrm{supp}(\nu)} \mathrm{I}_f(s). \tag{5}$$

Define the following optimization problem on the value function $v$:

$$\begin{aligned}
\max_{v \in \mathbb{R}^{|\mathcal{S}|}} \quad & (1-\gamma) \sum_s v(s) d_0(s) \\
\text{s.t.} \quad & v(s) \le \sum_a \pi(a \mid s) r(s,a) + \gamma V(s),\ \forall s \in \mathcal{S}, \\
\text{where} \quad & V(s) := \max_{\lambda \ge 0} \left\{ -\lambda \rho_s + \frac{1}{n_s} \sum_{i=1}^{n_s} \min_{a \in \mathcal{A}, s' \in \mathcal{S}} \left\{ v(s') \beta_s(a) + \lambda c((a,s'),(a_i,s_i')) \right\} \right\},
\end{aligned} \tag{V}$$

We have the following result showing the equivalence between (P) and (V).

THEOREM 1. *For every* $s \in \mathcal{S}$, *let* $\epsilon_s \in (0, \frac{1-\gamma}{2\gamma})$ *and* $M_s := \max_{a\in\mathcal{A}} \beta_s(a) - \min_{a\in\mathcal{A}} \beta_s(a)$, *and assume* $\rho_s$ *satisfies* $\rho_s \|\beta_s\|_{\mathrm{Lip},\hat{\mu}_s} \le \frac{1-\gamma}{2\gamma} - \epsilon_s$. *Then with probability at least* $1 - \sum_{s\in\mathcal{S}} \exp(-2n_s\epsilon_s^2/M_s^2)$, *it holds that*

(I) *The vector-valued mapping*

$$v \mapsto \left[\sum_a \pi(a \mid s) r(s,a) + \gamma V(s)\right]_{s\in\mathcal{S}}$$

*is contractive with Lipschitz constant* $\frac{1+\gamma}{2}$;

(II) *The optimal values of* (P) *and* (V) *coincide.*

Problem (V) can be viewed as a robust Bellman recursion, in which $V(s)$ is the dual reformulation of the robust reward-to-go function $\min_{\mu\in\mathfrak{M}_{\hat{\mu}}(\rho)} \sum_{a\in\mathcal{A},s'\in\mathcal{S}} \mu(a,s' \mid s) v(s') \beta_s(a)$. One notable difference between Theorem 1 and standard results on s-rectangular robust MDP is that its statement is not deterministic, as the mapping in (I) is contractive only probabilistically. For on-policy evaluation, i.e., $\pi = \pi_b$, we have $\beta_s = 1$ and $\|\beta_s\|_{\mathrm{Lip},\hat{\mu}_s} = M_s = 0$, (V) becomes deterministic and provides a robust counterpart result of Puterman (1994). To obtain negligible error probability $\tau$ in Theorem 1, for fixed $\gamma \in (0,1)$, the sample size should be sufficiently large so that

$$\min_{s\in\mathcal{S}} \frac{n_s}{M_s^2} = \Omega\left(\frac{\gamma^2}{(1-\gamma)^2} \log \frac{|\mathcal{S}|}{\tau}\right). \tag{6}$$

Therefore, the required sample size is *nearly state space size-independent* as the lower bound only has a logarithmic dependence on the the number of states. In addition, taking discounted factor $\gamma$ close to 1 will require the sample size lower bound in (6) grow linearly with respect to $1/(1-\gamma)^2$. The proof of 1 is based on techniques from robust MDP with s-rectangular sets (Wiesemann et al. 2013) and Wasserstein distributionally robust optimization (Esfahani and Kuhn 2018, Blanchet and Murthy 2019, Gao and Kleywegt 2022).

It is clear from Theorem 1 that $v$ satisfies the robust counterpart of the fixed-point condition

$$v(s) = \sum_a \pi(a \mid s) r(s,a) + \gamma \min_{\mu\in\mathfrak{M}_{\hat{\mu}}(\rho)} \sum_{a\in\mathcal{A},s'\in\mathcal{S}} \mu(a,s' \mid s) v(s') \beta_s(a), \quad \forall s \in \mathcal{S}, \tag{7}$$

whence $v$ can be solved by value iteration, which is summarized in Algorithm 1.

---

**Algorithm 1** Value Iteration Algorithm for Robust Reward Evaluation

---

1: **Input**: $\gamma$, $\hat{\mu}$, $\rho$. Initialize $v \in \mathbb{R}_+^{|\mathcal{S}|}$.
2: **while** not converge **do**
3:     For each $s \in \mathcal{S}$, compute $V(s)$ defined in (V).
4:     For each $s \in \mathcal{S}$, update

$$v(s) \leftarrow \sum_a \pi(a \mid s) r(s,a) + \gamma \cdot V(s)$$

5: **end while**

---

The following proposition shows that the iterates in Algorithm 1 are guaranteed to converge into the optimal solution of (V).

PROPOSITION 1. *Under the setting of Theorem 1, the iterate* $v$ *in Algorithm 1 converges to the optimal solution to* (V).

### 4.1. A Regularized Lagarangian Perspective

In this subsection, we provide a different interpretation of (P) and (V) from the perspective of regularization. To this end, we first write the dual form of (V).

PROPOSITION 2. *Problem* (V) *admits a strong dual formulation*

$$
\begin{aligned}
\min_{\kappa \in \mathbb{R}_+^{|\mathcal{S}|}, q \in \mathbb{R}_+^{\sum_{s\in\mathcal{S}} n_s |\mathcal{A}||\mathcal{S}|}} \quad & \sum_{s\in\mathcal{S}, a\in\mathcal{A}} \pi(a \mid s)\kappa(s) r(s,a) \\
\text{s.t.} \quad & (1-\gamma) d_0(s') + \gamma \sum_{s\in\mathcal{S}, a\in\mathcal{A}} \beta_s(a)\kappa(s) \sum_{i=1}^{n_s} q_{i,(a,s')}^{(s)} = \kappa(s'),\ \forall s' \in \mathcal{S}, \\
& \sum_{i=1}^{n_s} \sum_{a\in\mathcal{A}, s'\in\mathcal{S}} q_{i,(a,s')}^{(s)} c((a,s'),(a_i,s_i')) \le \rho_s,\ \forall s \in \mathcal{S}, \\
& \sum_{a\in\mathcal{A}, s'\in\mathcal{S}} q_{i,(a,s')}^{(s)} = \frac{1}{n_s},\ \forall i = 1,2,\ldots,n_s,\ \forall s \in \mathcal{S}.
\end{aligned} \tag{8}
$$

This formulation can be interpreted as follows. Note that it can be easily verified that $\sum_{s\in\mathcal{S}} \kappa(s) = 1$ by adding up the first set of equality constraints in (8). Thereby the decision variable $\kappa$ can be viewed as the average visitation distribution under the target policy $\pi$. The decision variable $q^{(s)}$ can be viewed as a transport plan that transports probability mass from the empirical distribution $\hat{\mu}_s$ to a new distribution $\mu_s$, served as a surrogate for the underlying average state-visitation distribution $d_{\pi_b}$. The first set of constraint in (8) describes the system of stationary equation under the policy $\pi$. The second and third sets of inequality constraints confines the distribution $\mu_s$ to be within the Wasserstein ball. When $\mu = d_{\pi_b}$ and $\rho_s = 0$ for all $s \in \mathcal{S}$, (8) reduces to the bilevel feasibility formulation (2)(3). Thus, (8) is a robust counterpart that outputs the worst-case average visitation distribution $\kappa$ under the target policy $\pi$.

Introducing a Lagrangian multiplier $f$ for the first set of constraints in (8), we can obtain an equivalent Lagrangian reformulation of (8).

PROPOSITION 3. *Let $\Delta^{|\mathcal{S}|}$ be the probability simplex in $\mathbb{R}^{|\mathcal{S}|}$. There exists $\{\overline{\rho}_s\}_{s\in\mathcal{S}} \subset \mathbb{R}_+^{|\mathcal{S}|}$ such that for all $\{\rho_s\}_{s\in\mathcal{S}}$ satisfying $\rho_s < \overline{\rho}_s$, $\forall s \in \mathcal{S}$, problem* (8) *is equivalent to*

$$
\begin{aligned}
\min_{\kappa\in\Delta^{|\mathcal{S}|}} \max_{f\in\mathbb{R}^{|\mathcal{S}|}} \ & \mathbb{E}_{s\sim\kappa, a\sim\pi(\cdot|s)}[r(s,a)] + (1-\gamma)\mathbb{E}_{s\sim d_0}[f(s)] + \gamma \mathbb{E}_{s\sim\kappa,(a,s')\sim\hat{\mu}_s}[\beta_s(a) f(s')] - \mathbb{E}_{s\sim\kappa}[f(s)] \\
& - \gamma \mathbb{E}_{s\sim\kappa}[\rho_s \|\beta_s f\|_{\mathrm{Lip},\hat{\mu}_s}].
\end{aligned}
$$

Proposition 3 indicates that the distributionally robust formulation (P) is equivalent to a weighted Lipschitz regularization on $f$ for sufficiently small Wasserstein radii $\{\rho_s\}_{s\in\mathcal{S}}$, where the expression of the threshold $\{\bar{\rho}_s\}_{s\in\mathcal{S}}$ can be found in the proof of Proposition 3 in Appendix EC.3.2. For general radii $\{\rho_s\}_{s\in\mathcal{S}} \subset \mathbb{R}_+^{|\mathcal{S}|}$, the regularized minimax formulation above constitutes a lower bound for the problem (8). As a consequence, Proposition 3 uncovers the connection between distributionally robust formulation and the Lipschitz regularization formulation. This formulation falls into the family of regularized Lagrangian formulation described in Yang et al. (2020), but it provides a principled Lipschitz regularization resulting from the distributionally robust formulation (P). This min-max formulation is reminiscent of the generative adversarial network (GAN) (Goodfellow et al. 2014) in the deep learning literature, where $\kappa$ and $f$ are referred to as the generator and discriminator, respectively, and are often parameterized by neural networks. Thus, Proposition 3 suggests an alternative way to solving (P) using deep learning techniques. Study on this approach is beyond the scope of this paper, but below we detail several theoretical problems worth investigation. First, it is of research interest to investigate how large the network size is enough to approximate the underlying optimal generator and discriminator functions. Recent work suggests that neural networks can efficiently approximate functions with special data structures such as low-dimensional manifolds (Hanin 2019, Chen et al. 2020a, 2022, 2019a). It is desirable to generalize those analyses on our formulation. Second, it is important to understand the landscape of the non-convex training problem with neural networks. Both empirical and theoretical studies reveal that over-parameterized neural networks can benefit the optimization training (Mei et al. 2018, Du et al. 2019, Mei et al. 2019, Sun 2019, Zhang et al. 2020b, Sirignano and Spiliopoulos 2020, Jacot et al. 2018), but the analysis specialized to the off-policy evaluation problem remains unknown.

Substituting $\kappa(s)$ with $\hat{\mu}(s)w(s)$ for $s \in \mathcal{S}$ yields

$$\min_{w\geq 0} \max_{f\in\mathbb{R}^{|\mathcal{S}|}} (1-\gamma)\mathbb{E}_{s\sim d_0}[f(s)] \\ +\mathbb{E}_{(s,a,s')\sim\hat{\mu}}\Big[w(s)(\gamma\beta_s(a)f(s') - f(s) + \mathbb{E}_{a_0\sim\pi(\cdot|s)}[r(s,a_0)] - \gamma\rho_s\|\beta_s f\|_{\mathrm{Lip},\hat{\mu}_s}])\Big].$$

Define the function

$$L(w,f) := (1-\gamma)\mathbb{E}_{s\sim d_0}[f(s)] + \mathbb{E}_{(s,a,s')\sim\hat{\mu}}\Big[(\gamma\beta_s(a)f(s') - f(s))w(s)\Big].$$

When $\rho_s = 0, \forall s \in \mathcal{S}$, the problem above becomes

$$\min_{w\geq 0} \mathbb{E}_{s\sim\hat{\mu},a_0\sim\pi(\cdot|s)}[w(s)r(s,a_0)] + \max_{f\in\mathbb{R}^{|\mathcal{S}|}} L(w,f).$$

The problem above reduces to finding an importance weight function $w$ such that

$$L(w,f) = 0, \quad \forall f \in \mathbb{R}^{|\mathcal{S}|},$$

since otherwise, due to the linearity of $L(w,f)$ in $f$, one can always pick the function $f$ to make the objective value grow to infinity. In comparison, the algorithm developed in Liu et al. (2018b) solves

$$\tilde{L}(w,f) := (1-\gamma)\mathbb{E}_{s\sim d_0}[(1-w(s))f(s)] + \gamma\mathbb{E}_{(s,a,s')\sim\hat{\mu}}\Big[(w(s)\beta_s(a) - w(s'))f(s')\Big] = 0, \quad \forall f \in \mathcal{F},$$

where $\mathcal{F} \subset \mathbb{R}^{|\mathcal{S}|}$ is a suitable family of discriminators. Observe that

$$L(w,f) - \tilde{L}(w,f) = (1-\gamma)\mathbb{E}_{s\sim d_0}[w(s)f(s)] + \mathbb{E}_{(s,a,s')\sim\hat{\mu}}[\gamma f(s')w(s') - w(s)f(s)]. \tag{9}$$

Recall that for any $s' \in \mathcal{S}$,

$$(1-\gamma)d_0(s') - d_{\pi_b}(s') + \gamma\sum_{(s,a)}\pi_b(a\mid s)P(s'\mid s,a)d_{\pi_b}(s) = 0.$$

Multiplying both sides with the test function $w(s')\cdot f(s')$ for $s' \in \mathcal{S}$ implies that

$$(1-\gamma)\mathbb{E}_{s\sim d_0}[w(s)f(s)] - \mathbb{E}_{s\sim d_{\pi_b}}[w(s)f(s)] + \gamma\mathbb{E}_{(s,a,s')\sim d_{\pi_b}}[w(s')f(s')] = 0.$$

Equivalently,

$$(1-\gamma)\mathbb{E}_{s\sim d_0}[w(s)f(s)] + \mathbb{E}_{(s,a,s')\sim d_{\pi_b}}[\gamma f(s')w(s') - w(s)f(s)] = 0. \tag{10}$$

Comparing the identity (9) with (10), we can see the two objective functions $L(w,f)$ and $\tilde{L}(w,f)$ coincide when $\hat{\mu}$ is replaced with $d_{\pi_b}$.Hence, our formulation reduces to the formulation in Liu et al. (2018b) when $\hat{\mu} = d_{\pi_b}$ and $\rho_s = 0$ for all $s \in \mathcal{S}$.

## 5. Uncertainty Quantification

In this section, we provide uncertainty quantification for OPE under the two situations described in the introduction. In Section 5.1, we develop asymptotic and non-asymptotic confidence intervals when the logged trajectory data set contains one or more finite-length (possibly non-stationary) trajectories under the behavior policy. In Section 5.2, we derive the confidence interval estimates of the adversarial reward under changing MDP environments. Asymptotic result provides a characterization of the distributional behavior for a large sample size and are practically useful. In contrast, non-asymptotic result provides confidence bounds that are valid for any sample size and reveals the impact of parameters such as data dimension and problem hyper-parameters on the confidence interval. The results in this section are based on the finite-sample performance guarantees for Wasserstein DRO on the discrete space (Appendix B), which may be of independent interest.

Throughout this section and the next section, we define the following parameters for notational simplicity:

$$\delta_s = \min_{(a,s')\in \text{supp}\, d_{\pi_b}(\cdot,\cdot|s)} d_{\pi_b}(a,s' \mid s), \quad \Delta_s = \frac{1}{\left(\frac{1-2\delta_s}{1-\delta_s} \vee \frac{2\delta_s - 1}{\delta_s}\right)},$$
$$\delta'_s = \min_{a\in \text{supp}\, \pi_b(\cdot|s)} \pi_b(a \mid s), \quad M_s := \max_{a\in\mathcal{A}} \beta_s(a) - \min_{a\in\mathcal{A}} \beta_s(a), \quad s\in\mathcal{S}.$$

In addition, we assume that the value function $|v(s)| \le M$ for any $s\in\mathcal{S}$ and some constant $M>0$.

### 5.1. Confidence Interval using Finite-length Trajectories

In this subsection, we consider a stochastic setting where trajectory data are Markovian. Proofs for results in this section are provided in Appendix EC.4.1.

Recall that $R_\pi$ is the true value of the target policy $\pi$, and $\mathcal{L}_{\hat{\mu}}(\rho), \mathcal{U}_{\hat{\mu}}(\rho)$ are, respectively, the robust and optimistic value estimates defined in (P). In this subsection, we explicitly use subscript $n$ to denote the dependence on sample size $n=(n_s)_{s\in\mathcal{S}}$ for relevant quantities. The next theorem establishes non-asymptotic confidence bounds on $R_\pi$ based on $\mathcal{L}_{\hat{\mu}_n}(\rho_n)$ and $\mathcal{U}_{\hat{\mu}_n}(\rho_n)$.

Theorem 2. *For $s\in\mathcal{S}$, let $\tau_s>0$, and set $\rho_{n,s} = \sqrt{\frac{2\tau_s}{n_s}}\text{diam}(\mathcal{A}\times\mathcal{S})$, and $\epsilon_{n,s} = \frac{6}{n_s}$. Let $\epsilon_n = (\epsilon_{n,s})_{s\in\mathcal{S}}$. Then with probability at least $1-2\sum_{s\in\mathcal{S}}\alpha_s$, where*

$$\alpha_s := \exp\left(-\tau_s + \log(2n_s M)\right) + \exp\left(-n_s\delta_s + \log(2n_s M)\right) + \exp\Big(-n_s \log\Delta_s + \log(2n_s M)\Big),$$

*it holds that*

$$\mathcal{L}_{\hat{\mu}_n}(\rho_n) - d_0^\top (I-\gamma P^{\text{true}})^{-1}\epsilon_n \le R_\pi \le \mathcal{U}_{\hat{\mu}_n}(\rho_n) + d_0^\top (I-\gamma P^{\text{true}})^{-1}\epsilon_n.$$

Theorem 2 shows that $[\mathcal{L}_{\hat{\mu}_n}(\rho_n) - d_0^\top (I-\gamma P^{\text{true}})^{-1}\epsilon_n,\ \mathcal{U}_{\hat{\mu}_n}(\rho_n) + d_0^\top (I-\gamma P^{\text{true}})^{-1}\epsilon_n]$ can be served as a confidence interval for $R_\pi$. Each row of the term $d_0^\top (I-\gamma P^{\text{true}})^{-1}\epsilon_n$ is of high-order $O(1/n_s)$, as compared to the length of $[\mathcal{L}_{\hat{\mu}_n}(\rho_n),\ \mathcal{U}_{\hat{\mu}_n}(\rho_n)]$ that has a square-root dependence on $1/n_s$ established in Proposition 4 below. The proof of Theorem 2 is based on a perturbation analysis on the Bellman operator (Lemma EC.6) and a finite-sample performance bound on Wasserstein DRO (Lemma EC.8). In particular, our finite-sample guarantee is based on covering number arguments as opposed to the VC dimension used in Dai et al. (2020). Unlike Dai et al. (2020) and Feng et al. (2020), we do not require the assumption that the value function belongs to a reproducing kernel Hilbert space. To prove Theorem 2, we leverage the out-of-sample performance guarantee for Wasserstein DRO in Proposition 8, which depends on a fractional parameter $\eta\in(0,1]$. When presenting the main result in Theorem 2, we select the fraction parameter $\eta=1$ for simplicity. One can tune this parameter to balance between the error probability and the radius. For problems where the global slope (see (4) for definition) of the value function is much smaller than its Lipschitz norm (see (5) for definition) for many states, one can reduce the fraction $\eta$ to obtain a tighter confidence interval.

Remark 1 (Asymptotic CI). Under the setting of Theorem 2, if we choose the Wasserstein radius $\rho_{n,s} = \sqrt{\frac{2\tau_s}{n_s}}\mathrm{diam}(\mathcal{A}\times\mathcal{S})$ with $\tau_s = \tau + \log(2n_s M)$, then with probability at least

$$1-2|\mathcal{S}|e^{-\tau} - 2\sum_s \left[\exp\Big(-n_s\delta_s + \log(2n_s M)\Big) + \exp\Big(-n_s\log\Delta_s + \log(2n_s M)\Big)\right],$$

the underlying reward $R_\pi \in [\mathcal{L}_{\hat{\mu}_n}(\rho_n) - d_0^\top (I-\gamma P^{\mathrm{true}})^{-1}\epsilon_n,\ \mathcal{U}_{\hat{\mu}_n}(\rho_n) + d_0^\top (I-\gamma P^{\mathrm{true}})^{-1}\epsilon_n]$. Moreover, the probability bound in Theorem 2 converges to $1-2|\mathcal{S}|e^{-\tau}$ asymptotically. As a consequence, the proposed CI is a $(1-\alpha)$-asymptotic confidence interval by choosing the parameter $\tau$ so that $e^{-\tau+\log(2|\mathcal{S}|)} = \alpha$.

The next proposition provides an upper bound on the length of our proposed CI.

Proposition 4. *For $s\in\mathcal{S}$, let $\tau_s, \tau_s' > 0$, $\rho_{n,s} = \sqrt{\frac{2\tau_s}{n_s}}\mathrm{diam}(\mathcal{A}\times\mathcal{S})$, $n_s \geq \frac{4\tau_s'^2\gamma^2}{(1-\gamma)^2}$, and set*

$$\epsilon_{n,s} := \gamma\rho_{n,s} \max_{v\in\mathbb{R}^{|\mathcal{S}|}:\ |v(s)|\leq M} \|\beta_s v\|_{\mathrm{Lip}, d_{\pi_b}(\cdot,\cdot|s)}.$$

*Then with probability at least $1-\sum_{s\in\mathcal{S}}\exp(-2\tau_s'^2/M_s^2)$, it holds that*

$$\mathcal{U}_{\hat{\mu}_n}(\rho_n) - \mathcal{L}_{\hat{\mu}_n}(\rho_n) \leq 2d_0^T (I-\gamma P_{\hat{\mu}_n})^{-1}\epsilon_n,$$

*where $P_{\hat{\mu}_n}(s,s') = \sum_a \hat{\mu}_n(a,s' \mid s)\beta_s(a)$.*

Proposition 4 reveals that the length of our established confidence interval depends inversely on $1/\sqrt{n_s}$. For small state spaces in which $|\mathcal{S}|$ is viewed as an irrelevant constant, this is optimal; while for large state spaces, the gap between $1/\sqrt{n_s}$ and $1/\sqrt{\sum_{s\in\mathcal{S}} n_s}$ roughly means that our bound has an extra factor of $1/\sqrt{|\mathcal{S}|}$. Such conservativeness arises from the s-rectangularity of the uncertainty set $\mathfrak{M}_{\hat{\mu}}(\rho)$. Nonetheless, the s-rectangularity is designed to maintain tractability for distributionally robust MDP and is seemingly unavoidable; see the NP-hardness discussion in Wiesemann et al. (2013). If we consider the regularized Lagrangian formulation in Section 4.1 and assume the global optimality of the involved nonconvex problem (as did in Dai et al. (2020)), then an extension of our formulation may achieve optimal sample rate, which is left to future work.

### 5.2. Confidence Interval under Changing Environments

In this subsection, we consider OPE under distribution shift. Most existing works on OPE for RL are based on a key assumption that the future environment in which the target policy is deployed is the same as the past environment from which the logged trajectory data are collected. As motioned in the introduction, such an assumption may not necessarily hold in practical scenarios. Under the changing environments, the average visitation distribution $d_{\pi_b}$ may be different in the future environment, which results in a different value $R_\pi$. This holds even when $d_{\pi_b}$ is exactly known for the past environment. Hence, it is important to understand the performance of the target policy under adversarial scenarios and quantify its uncertainty.

In the spirit of Si et al. (2020) which studies OPE for bandits under distribution shift, we refer to $\mathcal{L}^{\mathrm{adv}}(\rho)$ as the *adversarial value* under policy $\pi(s|a) = \pi_b(s|a)\beta_s(a)$, defined as the worst-case reward under an adversarial changing environment, with the radius $\rho$ capturing the discrepancy between the future and the past:

$$\begin{aligned}\mathcal{L}^{\mathrm{adv}}(\rho) := \min_{w,\mu} \ & \sum_{s\in\mathcal{S},a\in\mathcal{A}} \mu(s)\pi_b(a\mid s)w(s)\beta_s(a)r(s,a)\\ \text{s.t.}\ \ & w(s')\mu(s') = (1-\gamma)d_0(s') + \gamma\sum_{s\in\mathcal{S},a\in\mathcal{A}}\mu(s,a,s')\beta_s(a)w(s),\quad \forall s'\in\mathcal{S},\\ & \mu\in\mathfrak{M}_{d_{\pi_b}}(\rho).\end{aligned} \tag{11}$$

The difference between the equation above and (P) is that the center of the Wasserstein ball $\hat{\mu}$ in (P) is replaced by the true average visitation distribution $d_{\pi_b}$ in the past environment. Different from the previous subsection, the radius $\rho$ is *fixed* and not varying in the sample size, simply because the gap between the future and the past remains even when we have infinite amount of historical data. We are interested in developing a CI for $\mathcal{L}^{\mathsf{adv}}(\rho)$ using only the logged trajectory data, i.e., based on the empirical estimate $\mathcal{L}_{\hat{\mu}_n}(\rho)$. Proofs for results in this section are provided in Appendix EC.4.2.

The following theorem establishes the finite sample guarantee for estimating $\mathcal{L}^{\mathsf{adv}}(\rho)$ based on $\mathcal{L}_{\hat{\mu}_n}(\rho)$. Consider the function space $\mathcal{F}_s := \{(a,s') \mapsto \beta_s(a)v(s'), |v(s')| \le M, \forall s' \in \mathcal{S}\}$. Let $\mathbb{E}_{\otimes}[\mathfrak{R}_{n_s}(\mathcal{F}_s)]$ denote the Rademacher complexity of the space $\mathcal{F}_s$ with respect to $d_{\pi_b}(\cdot,\cdot \mid s)$ for sample size $n_s$.

THEOREM 3. *Let $\tau > 0$ and for $s \in \mathcal{S}$, set*

$$H_s = \max_{f \in \mathcal{F}_s} \|f\|_\infty, \quad \iota_{n,s} = 2\left(2\mathbb{E}_{\otimes}[\mathfrak{R}_{n_s}(\mathcal{F}_s)] + H_s\sqrt{\frac{\tau}{2n_s}}\right).$$

*Then there exists $\{\overline{\rho}_s\}_{s\in\mathcal{S}} \subset \mathbb{R}_+^{|\mathcal{S}|}$ such that for all $\{\rho_s\}_{s\in\mathcal{S}}$ satisfying $\rho_s < \overline{\rho}_s$ and $\rho_s\|\beta_s\|_{\mathrm{Lip},d_\pi(\cdot,\cdot|s)} \le \frac{1-\gamma}{2\gamma}$, $\forall s \in \mathcal{S}$, with probability at least $1 - \sum_s \alpha_s$, where*

$$\alpha_s := \exp\left(-n_s\delta_s + \log(2n_sM)\right) + 2e^{-\tau},$$

*it holds that*

$$\mathcal{L}_{\hat{\mu}_n}(\rho) - d_0^\top \iota_n \le \mathcal{L}^{\mathsf{adv}}(\rho) \le \mathcal{L}_{\hat{\mu}_n}(\rho) + d_0^\top \iota_n.$$

Note that the term $\iota_n$ in Theorem 3 and the term $\epsilon_n$ in Theorem 2 have different orders of the sample size. This is because the benchmarks are different under stochastic and adversarial settings. In the stochastic setting (Theorem 2), we would like to choose a radius $\rho_n$ such that the robust reward serves as a high-confidence lower bound, thus the residual $\epsilon_n$ is of higher-order than the length of the CI. Whereas in the adversarial setting, the radius of the distributional uncertainty set is fixed in advance, capturing the distribution shift under changing environments. The goal is to provide a CI for the adversarial reward $\mathcal{L}^{\mathsf{adv}}(\rho)$ using an empirical estimate $\mathcal{L}_{\hat{\mu}_n}(\rho)$, thus the term $\iota_n$ represents the half length of the CI. Similar to Proposition 4, the length of the confidence interval depends inversely on $1/\sqrt{n_s}$.

Below we also provide an asymptotic CI for $\mathcal{L}^{\mathsf{adv}}(\rho)$ based on $\mathcal{L}_{\hat{\mu}}(\rho)$. To ease the exposition, we only consider a single trajectory with length $T$, but the result is readily generalized to multiple (independent) trajectories. In the following theorem, we explicitly use subscript $T$ to denote the dependence of $\hat{\mu}$ on the truncation length $T$.

THEOREM 4. *Set*

$$\begin{aligned}
r_\pi(s) &= \sum_a r(s,a)\pi(a \mid s), \;\; s \in \mathcal{S}, \\
D_{(s,a,s'),(\tilde{s},\tilde{a},\tilde{s}')} &= 1\{s = \tilde{s}, a = \tilde{a}, s' = \tilde{s}'\}\frac{1}{\sqrt{d_{\pi_b}(s)}}, \quad s,\tilde{s},s',\tilde{s}' \in \mathcal{S}, a, \tilde{a} \in \mathcal{A}, \\
P_{\mu^*}(s,s') &= \sum_a \beta_s(a)\mu^*(a,s' \mid s), \quad s,s' \in \mathcal{S}, \\
y_{(s,a,s')} &= \gamma(1-\gamma)\left((I - \gamma P_{\mu^*}^T)^{-1} d_0 r_\pi^T (I - \gamma P_{\mu^*}^T)^{-1}\right)_{s,s'} \beta_s(a), \quad s,s' \in \mathcal{S}, a \in \mathcal{A},
\end{aligned}$$

*where $\mu^*$ is the optimal solution to (11), and $\Lambda \in \mathbb{R}_+^{|\mathcal{S}||\mathcal{A}||\mathcal{S}|\times|\mathcal{S}||\mathcal{A}||\mathcal{S}|}$ is defined as*

$$\Lambda_{(s,(a,s')),(\tilde{s},(\tilde{a},\tilde{s}'))} = \begin{cases} d_{\pi_b}(a,s' \mid s)(1 - d_{\pi_b}(a,s' \mid s)), & \textit{if } (s,(a,s')) = (\tilde{s},(\tilde{a},\tilde{s}')), \\ -d_{\pi_b}(a,s' \mid s)d_{\pi_b}(\tilde{a},\tilde{s}' \mid s) & \textit{if } s = \tilde{s}, (a,s') \ne (\tilde{a},\tilde{s}'), \\ 0, & \textit{otherwise.} \end{cases}$$

*Assume that* $\rho_s \|\beta_s\|_{\mathrm{Lip}, d_\pi(\cdot,\cdot|s)} < \frac{1-\gamma}{\gamma}$, $s \in \mathcal{S}$. *Then it holds that*

$$\sqrt{T}\left(\mathcal{L}_{\hat{\mu}_T}(\rho) - \mathcal{L}^{\mathsf{adv}}(\rho)\right) \xrightarrow{\mathbf{d}} \mathcal{N}(0, y^T D \Lambda D y).$$

Recalling that $T = \sum_s n_s$, Theorem 4 provides an asymptotic CI with length $O(1/\sqrt{\sum_s n_s})$, which improves the order in the non-asymptotic CI derived in Theorem 3.

## 6. Distributionally Robust Batch Reinforcement Learning

Our distributionally robust framework can be easily leveraged for batch RL, whereby the decision-maker finds the optimal policy using a fixed set of logged trajectories generated from a behavior policy $\pi_b$, by solving the following max-min formulation

$$\mathcal{L}^*_{\hat{\mu}_n}(\rho_n) = \sup_{\pi \in \Pi} \inf_{\mu \in \mathfrak{M}_{\hat{\mu}_n}(\rho_n)} \mathbb{E}_{\pi,\mu}\left[\sum_{t=0}^{\infty} \gamma^t r_t\right]. \tag{12}$$

Let $R^{\mathsf{true}}$ be the optimal value under the true underlying MDP environment. Below, we develop a robust value iteration algorithm for solving (12) and provide its finite-sample performance guarantee. Proofs for results in this section are given in Appendix EC.5.

With a slight abuse of notation $\beta^\pi_s(a) := \frac{\pi(a|s)}{\pi_b(a|s)}$, define

$$\begin{aligned} \max_{\pi, v} \quad & (1-\gamma)\sum_s v(s) d_0(s) \\ \text{s.t.} \quad & v(s) \leq \sum_a \pi(a \mid s) r(s,a) + \gamma V(s), \ \forall s \in \mathcal{S}, \\ \text{where} \quad & V(s) := \min_{\mu \in \mathfrak{M}_{\hat{\mu}_n}(\rho_n)} \sum_{(a,s')} \mu(a, s' \mid s) v(s') \beta^\pi_s(a). \end{aligned} \tag{13}$$

Similar to Theorem 1, this optimization problem can be solved efficiently by robust value iteration.

THEOREM 5. *For* $s \in \mathcal{S}$, *let* $\epsilon_s \in \left(0, \frac{1-\gamma}{2\gamma}\right)$, *and suppose* $\rho_{n,s}$ *satisfies* $\rho_{n,s}\|\beta^\pi_s\|_{\mathrm{Lip}, \hat{\mu}_{n,s}} \leq \frac{1-\gamma}{2\gamma} - \epsilon_s$ *for every deterministic policy* $\pi$. *Then with probability at least* $1 - \sum_{s \in \mathcal{S}} \exp\left(-2 n_s \delta'^2_s \epsilon_s^2\right)$, *where* $\delta'_s = \min_{a \in \operatorname{supp} \pi_b(\cdot|s)} \pi_b(a \mid s)$, *it holds that*

(I) *The optimal values for problems* (12) *and* (13) *coincide;*

(II) *Let* $(v^*, \pi^*)$ *be the optimal solution to problem* (13). *Then* $v^*$ *solves the fixed point equation*

$$v(s) = \max_{\pi(\cdot|s)} \sum_a \pi(a \mid s) r(s,a) + \gamma V(s), \ \forall s \in \mathcal{S}, \tag{14}$$

*and* $\pi^*$ *solves the maximization on the right-hand side in* (14)*:*

$$\pi^*(\cdot \mid s) = \arg\max_{\pi(\cdot|s)} \sum_a \pi(a \mid s) r(s,a) + \gamma V(s), \ \forall s \in \mathcal{S}.$$

By Theorem 5, we can solve the batch reinforcement learning problem via robust value iteration, and the detailed procedure is presented in Algorithm 2. In each iteration we first obtain the worst-case average visitation distribution of behavior policy conditioned on states (see line 3 of Algorithm 2), which can be solved in similar manner as in (7), and then perform the policy improvement step. In particular, the policy improvement part will result in a deterministic optimal policy since the objective function is linear with respect to the policy. Similar to the analysis in (6), to achieve an error probability $\tau$, the sample size in Theorem 5 should be such that $\min_s n_s = \Omega\left(\frac{\gamma^2}{\delta'^2_s(1-\gamma)^2} \log \frac{|\mathcal{S}|}{\tau}\right)$. Compared with

**Algorithm 2** Value Iteration Algorithm for Robust Batch Reinforcement Learning

1: **Input**: $\hat{\mu}_n$, $\rho_n$, $\gamma$. Initialize $v \in \mathbb{R}_+^{|\mathcal{S}|}$.
2: **while** not converge **do**
3: For each $s \in \mathcal{S}$, compute

$$\mu^*(\cdot,\cdot \mid s) = \arg\min_{\mu \in \mathfrak{M}_{\hat{\mu}_n}(\rho_n)} \sum_{a \in \mathcal{A}, s' \in \mathcal{S}} \mu(a, s' \mid s) v(s') \beta_s^\pi(a).$$

4: For each $s \in \mathcal{S}$, update

$$v(s) \leftarrow \max_{\pi(\cdot|s)} \sum_a \pi(a \mid s) r(s,a) + \gamma \sum_{a \in \mathcal{A}, s' \in \mathcal{S}} \mu^*(a, s' \mid s) v(s') \beta_s^\pi(a).$$

5: **end while**

the robust OPE reformulation in Theorem 1, the required sample size $n_s$ depends on an additional factor $1/\delta_s'^2$. This results from that the value iteration algorithm for batch RL optimizes worst-case distribution and target policy jointly, and the contractive property of the robust Bellman recursion depends on the factor $\delta_s'$, the minimum entry of the probability vector of the behavior policy. The next theorem establishes performance guarantees when focusing on the collection of deterministic policies.

Theorem 6. *For $s \in \mathcal{S}$, let $\tau_s > 0$, $\epsilon_{n,s} = \frac{6}{n_s}$, and $\rho_{n,s} = \sqrt{\frac{2\tau_s}{n_s}} \operatorname{diam}(\mathcal{A} \times \mathcal{S})$. Then with probability at least $1 - \sum_{s \in \mathcal{S}} \alpha_s$, where*

$$\alpha_s := \exp\left(-\tau_s + \log(2|\mathcal{A}| n_s M)\right) + \exp\left(-n_s \delta_s + \log(2|\mathcal{A}| n_s M)\right) + \exp\Big(-n_s \log \Delta_s + \log\left(2|\mathcal{A}| n_s M\right)\Big),$$

*it holds that*

$$R^{\mathsf{true}} \geq \mathcal{L}^*_{\hat{\mu}_n}(\rho_n) - d_0^\top (I - \gamma P^{\mathsf{true}})^{-1} \epsilon_n.$$

This theorem shows that when choosing the radius $\rho_s = O(1/\sqrt{n_s})$, the optimal reward is lower bounded by the estimated robust reward up to a higher order residual. Similar to the discussion in Remark 1, $\mathcal{L}^*_{\hat{\mu}_n}(\rho_n)$ is an $(1-\alpha)$-asymptotic lower bound on the true optimal value $R^{\mathsf{true}}$ by taking $\rho_{n,s} = \sqrt{\frac{2\tau_s}{n_s}} \operatorname{diam}(\mathcal{A} \times \mathcal{S})$ with $\tau_s = \tau + \log(2|\mathcal{A}| n_s M)$, and the parameter $\tau$ is chosen such that $e^{-\tau + \log|\mathcal{S}|} = \alpha$.

Some recent works including MOReL (Kidambi et al. 2020) and MoPo (Yu et al. 2020) learn pessimistic MDPs based on offline data and then solve the batch reinforcement learning problem. However, their confidence bounds massively rely on the discrepancy between the learned MDP environment and the underlying true MDP environment, which are often conservative in practical settings. Most recently, Kumar et al. (2020) proposes a conservative Q-learning framework for robust batch framework by penalizing Q-values. Their work serves as a counter-part of distributionally robust framework from the perspective of regularization.

## 7. Numerical Simulations

In this section, we conduct numerical experiments in two discrete MDP environments to show the performance of the algorithms based on our framework for OPE. The description of two MDP environments is as follows:

**Machine Replacement Problem.** This MDP environment (Wiesemann et al. 2013) has 10 states and 2 actions: *Repair* and *Do Nothing*. States 1 to 8 model the states of deterioration of a machine and there are two repair states $R_1$ and $R_2$. The state $R_1$ is a normal repair state with reward 18, and the state $R_2$ is a long repair state with reward 10. The reward for states 1 to 7 is 20, and the reward for state 8 is 0. When the action *Do Nothing* is performed, for $i = 1, 2, \ldots, 7$, the state $S_i$ will remain in its current state with probability $p = 0.2$, and move to the state $S_{i+1}$ with probability $q = 0.8$. States $S_8$ and $R_1$ will remain in its current state with probability 1. The state $R_2$ will remain in its current state with probability $p = 0.2$, and move to the state $S_1$ with probability $q = 0.8$. When the action *Repair* is performed, for $i = 1, 2, \ldots, 8$, the state $S_i$ will move to repair states $R_1, R_2$ with probability $0.1, 0.6$, respectively, and move to the state $S_{\min\{i+1,8\}}$ with probability 0.3.

**Healthcare Management Problem.** This MDP (Goyal and Grand-Clement 2022) has six states $\{1, 2, 3, 4, 5, 6\}$ to model the physical conditions of patients, in which the state 6 is an absorbing mortality state. Three actions are available under this setting: *Do Nothing* ($a_1$), *Prescribe Low Drug Level* ($a_2$), and *Prescribe High Drug Level* ($a_3$). The goal of the agent is to minimize the mortality rate of patients and reduce the drug level to lower the harm of treatment. The reward at the state 6 is 0, while at remaining states, taking the action $a_1, a_2, a_3$ receives reward $10, 6, 2$, respectively. When the action $a_1$ is taken, for $i = 1, 2, \ldots, 5$, the state $S_i$ will transit into state $S_i, S_{i+1}, S_{\max\{1,i-1\}}$ with probability $p_1 = 0.4, p_2 = 0.3, p_3 = 0.3$, respectively. When the action $a_2$ or $a_3$ is taken, the values of $[p_1, p_2, p_3]$ are replaced with transition probabilities $[0.4, 0.2, 0.4]$ or $[0.4, 0.1, 0.5]$, respectively.

In order to simulate the task for OPE numerically, we set the target policy to be the optimal one trained by the $Q$-learning algorithm. For MRP scenario, we set the behavior policy to be the random policy such that conditioned on any state, the action follows uniform distribution. For HMP scenario, we set the behavior policy to be the one after running Q-learning for 5 iterations, since setting random policy will make $d_{\pi_b}(s) < 7 \times 10^{-3}$ for $s \in \{S_4, S_5\}$ and Assumption 2 will be approximately violated. The collected samples of state-action-state pairs are generated under the behavior policy. The cost function of the Wasserstein metric is set to be $c((a,s),(a',s')) = [|s - s'| + |a - a'|]/(|\mathcal{S}| + |\mathcal{A}|)$. The computed radii are recorded in Figure EC.1, EC.2, and EC.3, respectively.

### 7.1. Confidence Intervals for Non-stationary Trajectory Data

In this subsection we show the numerical performance for our interval estimates based on non-stationary trajectory data. In particular, we choose the radius size to be the one discussed in Remark 1 to realize the asymptotic 95% coverage rate. Recall that $T$ is the truncation length, and $J$ is the number of trajectories for observed samples. The default parameters are $T = 300$ and $J = 300$, unless varying them for performance comparison. The evaluation criterion is our estimated reward normalized by the underlying true reward under the target policy. Figure 1 reports the 95% confidence interval of the normalized reward across different choices of $T$ and $J$. The values of radii establishing confidence intervals are shared in Figure EC.1. As the number of trajectories or truncation length increases, both upper and lower confidence bounds become tighter, which suggests that our algorithm is able to give a sample-efficient confidence interval for off-policy evaluation.

Figure 2 shows the empirical coverage rate for the 95% confidence interval generated by our algorithm with 200 independent trials. The $y$-axis represents empirical error rates in which the corresponding confidence intervals do not cover the underlying true reward. While the asymptotic confidence regions may be overly optimistic for small number of trajectories (e.g., $J \le 200$ for MRP and $J \le 220$ for HMP) or short truncation length (e.g., $T \le 280$ for MRP and $T \le 400$ for HMP) cases, the figure shows the 5% error rate is controlled when the number of observations increases, justifying that the asymptotic confidence interval constructed in Remark 1 can roughly achieve the 95% coverage probability.

### 7.2. Confidence Intervals for Changing Environment

Next, we study the uncertainty quantification for OPE under changing environments. Assume that there exist experiment errors during the past MDP environments, and parameters for the transition dynamics will be perturbed a little bit. In particular, the parameters for the transition dynamics $(p,q),(p_1,p_2,p_3)$ in the two MDP environments above are replaced with $(p' = p+0.1, q' = q-0.1)$, $(p_1' = p_1 + 0.05, p_2' = p_2, p_3' = p_3 - 0.05)$ during the data collection phase. Moreover, during the data collection phase when the action $a_3$ is performed at state $S_6$, the state will transit into $S_5, S_6$ with probability $0.05, 0.95$, respectively. Once the perturbation of environment is high, the confidence interval will be overly conservative so that the off-policy evaluation task becomes meaningless. We thereby choose small perturbations to two MDP environments. In both scenarios, OPE based on the data for past environment over-estimates the underlying true reward under future MDP environment. We test the convergence of $\mathcal{L}_{\hat{\mu}_T}(\rho)$ into $\mathcal{L}^{\mathsf{adv}}(\rho)$ as the trajectory length increases, in which $\rho$ is chosen to be a constant radius size. We tune the radii $\rho$ such that the adversarial value $\mathcal{L}^{\mathsf{adv}}(\rho)$ serves as the lower bound for the underlying true reward under future MDP environment. Values of the radii are shared in Figure EC.2. Figure 3 shows the estimation of the adversarial reward normalized by its exact value across different number of truncation lengths, where we only use one trajectory to collect samples. Each data point in the plot represents the values of $\mathcal{L}_{\hat{\mu}_T}(\rho)$, and the error bars represent the asymptotic 95% confidence intervals discussed in Theorem 4. Note that the variance stated in Theorem 4 depends on $d_{\pi_b}$ and $\mu^*$, which cannot be obtained exactly. Therefore, we use the approximate variance instead by replacing these terms with $\hat{\mu}$ and the optimal solution to $\mathcal{L}_{\hat{\mu}_T}(\rho)$. From the plot we can see that as the sample size increases, both bias and variance for estimated adversarial reward values decrease, which indicates that $\mathcal{L}_{\hat{\mu}_T}(\rho)$ converges into $\mathcal{L}^{\mathsf{adv}}(\rho)$ well.

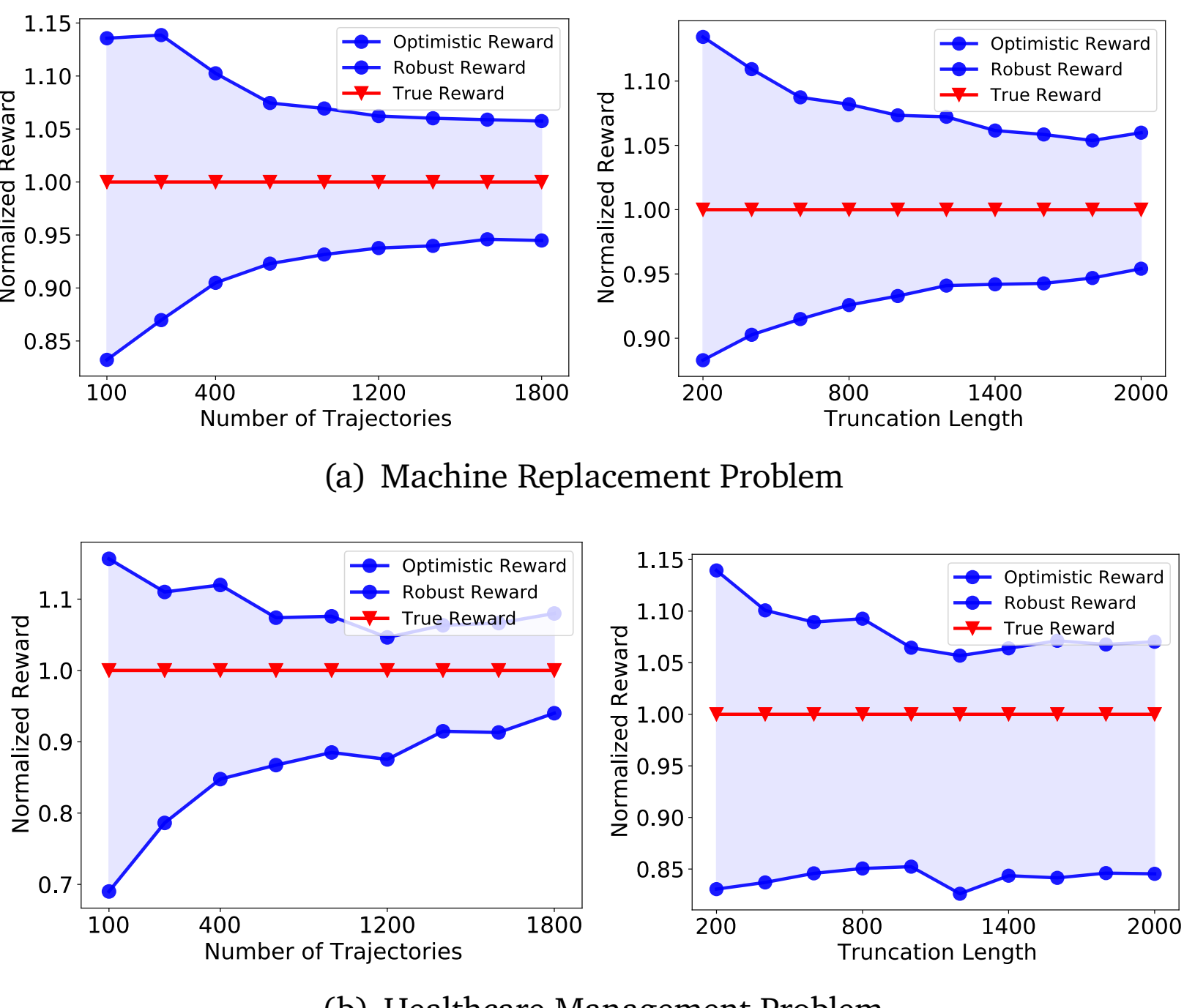

(a) Machine Replacement Problem

(b) Healthcare Management Problem

**Figure 1** Results for two MDP environments with discounted reward ($\gamma = 0.95$), in which the $y$-axis represents the estimated reward normalized by the underlying true reward under the target policy. The plots show the 95% confidence interval of the normalized reward generated by our algorithm across different numbers of trajectories and different truncation lengths.

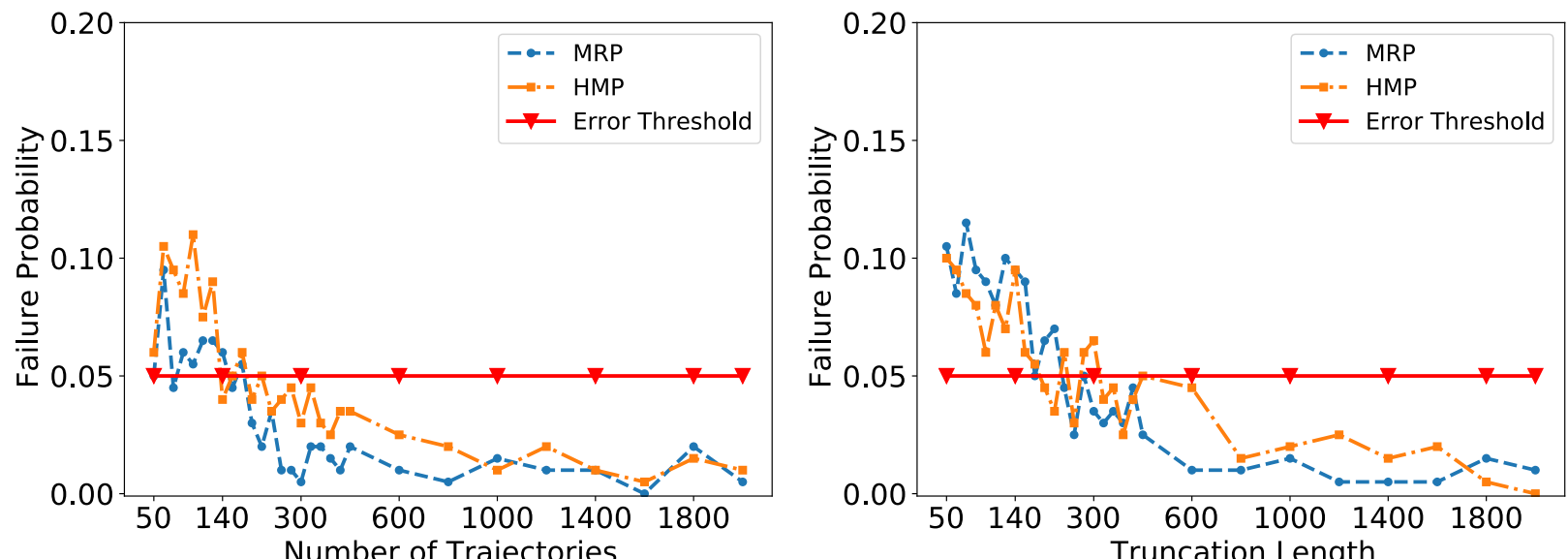

**Figure 2** Results for two MDP environments with discounted reward ($\gamma = 0.95$) on the empirical coverage rate of the constructed 95% confidence intervals. The plots show the empirical error rate generated by 200 independent trials across different numbers of trajectories and different truncation lengths.

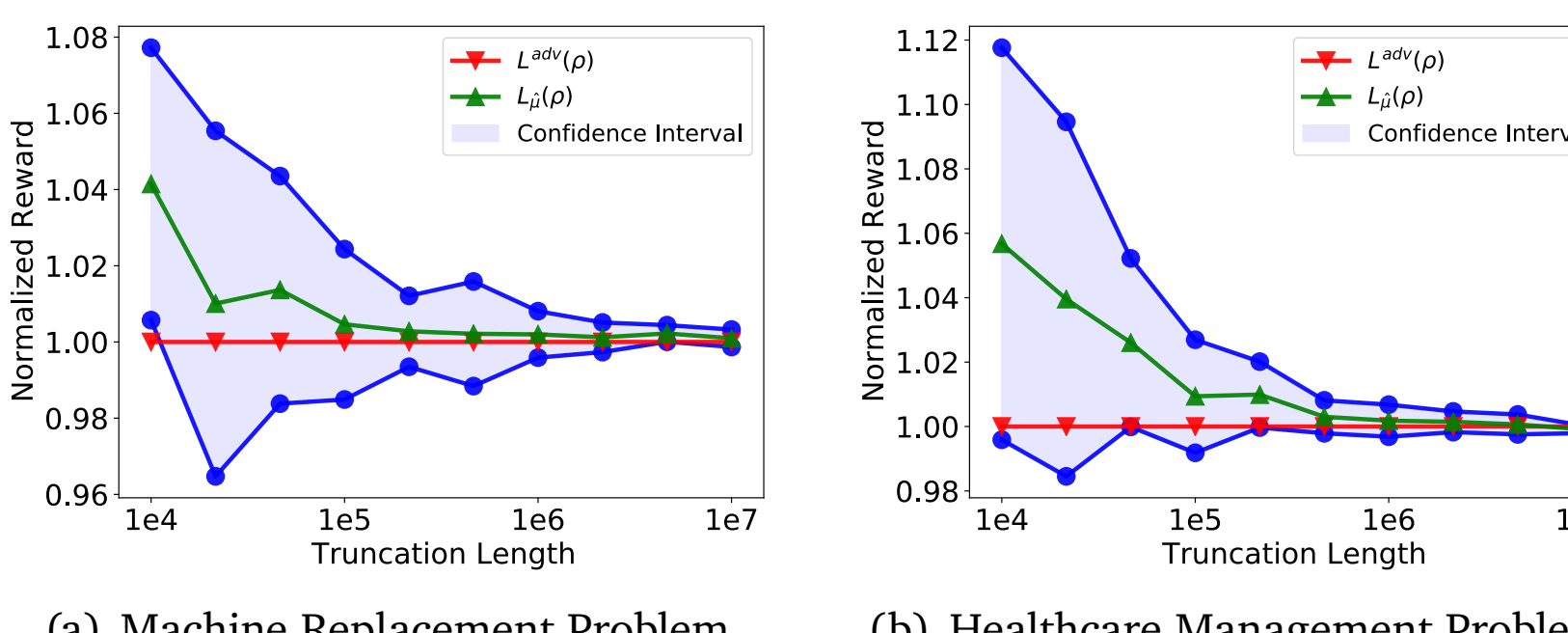

(a) Machine Replacement Problem (b) Healthcare Management Problem

**Figure 3** Results on the 95% confidence bound estimate for $\mathcal{L}^{\mathsf{adv}}$. The plots show the robust reward estimates across different truncation lengths within a single trajectory, where error bars are generated based on the asymptotic uncertainty quantification result in Theorem 4.

### 7.3. Distributionally Robust Batch Reinforcement Learning

Finally, we run experiments on the task of distributionally robust batch reinforcement learning, based on the historical data induced by a single behavior policy. We compare the performance of our robust algorithm with the algorithm based on the sample average approximation (SAA) approach within two MDP environments. The performance of an obtained policy $\pi$ is measured by its relative performance gap $100\times\frac{J^*-J(\pi)}{|J^*|}$, where $J^*$ denotes the underlying true reward when the MDP environment is exactly known, and $J(\pi)$ denotes the reward of the policy $\pi$ under the true MDP environment. Thus, the smaller the relative performance gap is, the better performance the policy has. Figure 4 reports the box plots for the relative performance gap of these two algorithms across different numbers of trajectories and truncation lengths. From the plot we can see that generally our algorithm reduces relative performance gap faster than the SAA approach as the sample size increases. Therefore, we conclude that our algorithm outperforms the SAA approach for the batch reinforcement learning task.

## 8. Concluding Remarks

In this paper, we develop a novel framework for computing either non-asymptotic or asymptotic confidence interval estimates for off-policy evaluation in infinite-horizon RL. Unlike existing approaches, we do not assume the restrictive i.i.d. or mixing conditions on the transition tuples and consider both Markovian and adversarial settings. When applying our framework to on-policy problems, our theory provide an end-to-end approach to statistical inference for robust MDP using trajectory data without explicitly estimating the transition probabilities. The length of our proposed CI estimates has an optimal sample rate for small state spaces, in which case we proposed efficient algorithms for both OPE and batch RL. Our formulation can be naturally extended to the behavior-agnostic setting, in which the behavior policy is not known to the decision maker. Our regularized Lagrangian formulation can be tailored to large or continuous state space by solving a minimax saddle point problem with the Lipschitz regularization, which is left for future work.

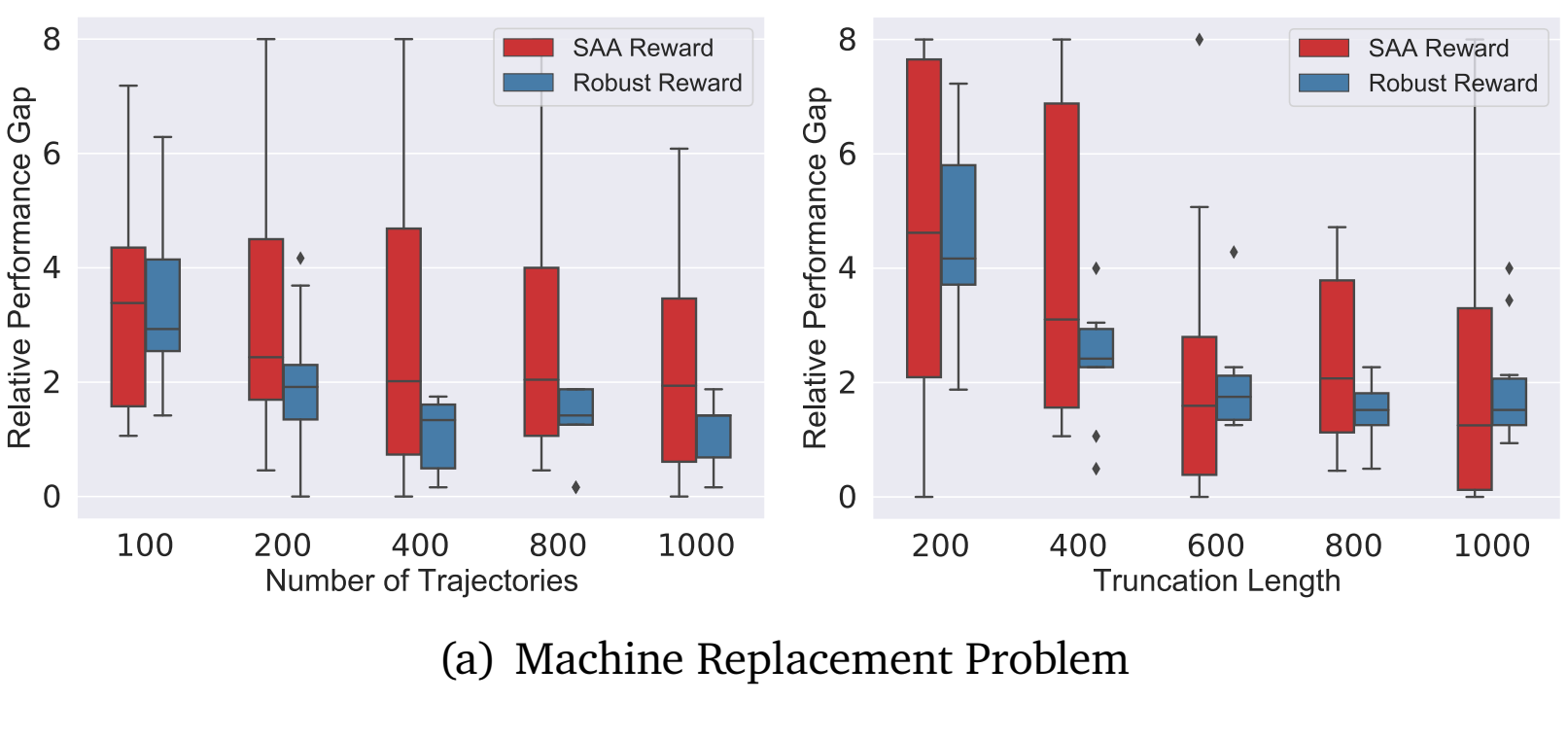

(a) Machine Replacement Problem

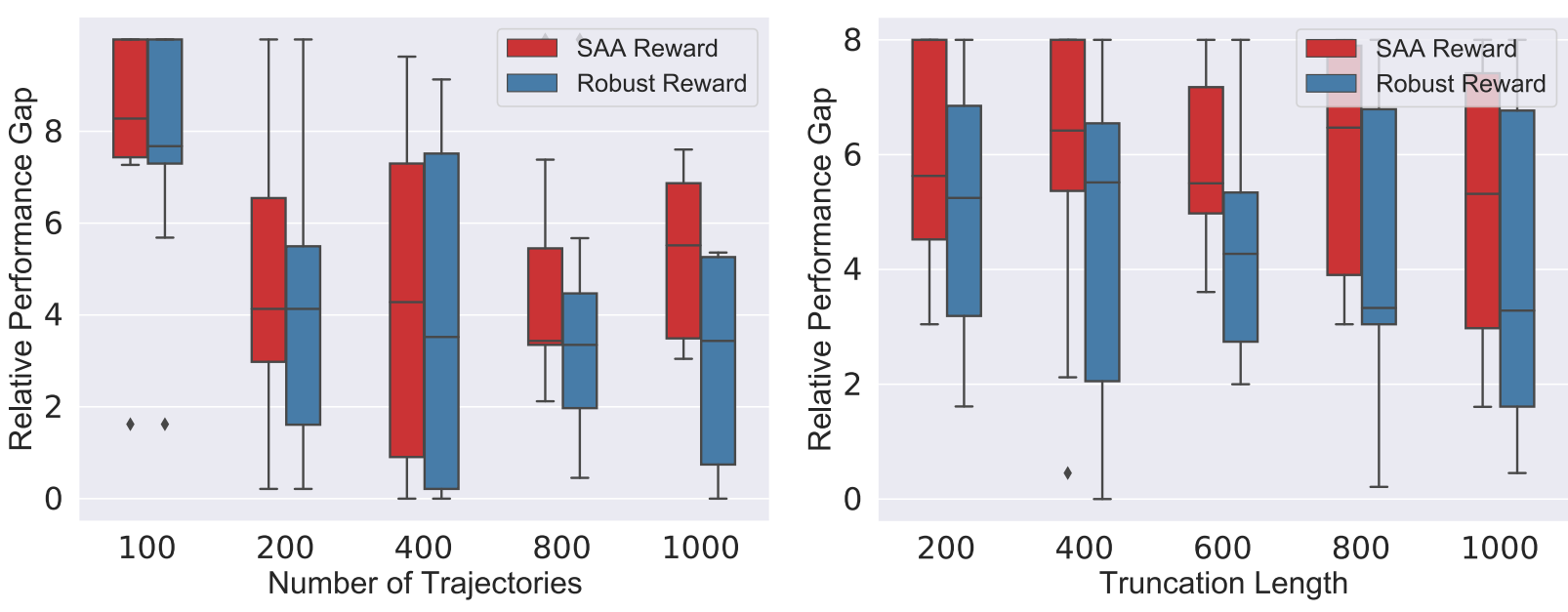

(b) Healthcare Management Problem

**Figure 4** Results for two MDP environments with discounted reward ($\gamma = 0.95$), where the evaluation criterion is chosen to be the relative performance gap between the estimated optimal reward and the underlying true reward when the MDP environment is exactly known. Experiments are conducted for different number of trajectories and different truncation lengths, where the box plots are generated based on 10 independent trials.

## 9. Acknowledgements

The third author's research is partially supported by a grant from the Shenzhen Science and Technology Program (JCYJ20210324120011032) and Shenzhen Institute of Artificial Intelligence and Robotics for Society. The authors would like to thank the referees and the Editorial team at *Operations Research* for their extensive feedback in improving this manuscript.

## 10. Brief Author Biographies

**Jie Wang** received the B.Sc. degree in pure mathematics from the Chinese University of Hong Kong, Shenzhen, in 2020. He is currently pursuing the Ph.D. degree with the H. Milton Stewart School of Industrial and Systems Engineering, Georgia Institute of Technology. His research interests include statistics, optimization, and network information theory.

**Rui Gao** is an assistant professor in the McCombs School of Business at the University of Texas at Austin. He received his B.S. degree in Mathematics and Applied Mathematics from Xi'an Jiaotong University in 2013, and his Ph.D. degree in Operations Research from Georgia Institute of Technology in 2018. His main research studies data-driven decision making under uncertainty and prescriptive data analytics.

**Hongyuan Zha** is a X.Q. Deng Presidential Chair Professor of the Chinese University of Hong Kong, Shenzhen. He received his B.S. degree in Mathematics from Fudan University in 1984, and his Ph.D. in Scientific Computing from Stanford University in 1993. He was a faculty member of College of Computing at Georgia Institute of Technology from 2006 to 2020, and the Department of Computer

Science and Engineering at Pennsylvania State University from 1992 to 2006. His research interest is machine learning.

### Appendix A: Results for Optimistic OPE

This section provides tractable formulations for optimistic OPE. It can be reformulated as the following minimization problem:

$$
\begin{aligned}
\min_{v} \quad & (1-\gamma)\sum_{s} v(s) d_0(s) \\
\text{s.t.} \quad & v(s) \geq \sum_{a} \pi(a \mid s) r(s,a) + \gamma V(s), \ \forall s \in \mathcal{S}, \\
\textit{where} \quad & V(s) := \min_{\lambda \geq 0} \left\{ \lambda \rho_s + \frac{1}{n_s} \sum_{i=1}^{n_s} \max_{a \in \mathcal{A}, s' \in \mathcal{S}} \left\{ v(s')\beta_s(a) - \lambda c((a,s'),(a_i,s_i')) \right\} \right\}.
\end{aligned}
$$

Theorem 7. *Under the same conditions of Theorem 1, with probability at least*

$$
1 - \sum_{s \in \mathcal{S}} \exp(-2 n_s \epsilon_s^2 / M_s^2),
$$

*the optimal values for the maximization problem* (P) *and the minimization problem presented above coincide.*

Hence, the optimistic reward can be evaluated by solving for the fixed-point equation

$$
v(s) = \sum_{a} \pi(a \mid s) r(s,a) + \gamma \max_{\mu \in \mathfrak{M}_{\hat{\mu}}(\rho)} \sum_{a \in \mathcal{A}, s' \in \mathcal{S}} \mu(a, s' \mid s) v(s') \beta_s(a), \ \forall s \in \mathcal{S}.
$$

With almost the same arguments presented in Section 4.1, we can develop a regularization counterpart for optimistic OPE. This problem with discriminator function constraints can be formulated as

$$
\begin{aligned}
\max_{\kappa \in \mathbb{R}_+^{|\mathcal{S}|}, q \in \mathbb{R}_+^{\sum_{s\in\mathcal{S}} n_s |\mathcal{A}||\mathcal{S}|}} \quad & \sum_{(s,a)} \pi(a \mid s) \kappa(s) r(s,a) \\
s.t. \quad & (1-\gamma)\mathbb{E}_{s\sim d_0}[f(s)] + \gamma \mathbb{E}_{s\sim\kappa}\left[ \mathbb{E}_{(a,s') \sim \sum_{i=1}^{n_s} q^{(s)}_{i,(a,s')}} [\beta_s(a) f(s')] \right] = \mathbb{E}_{s\sim\kappa}[f(s)], \ \forall f \in \mathbb{R}^{|\mathcal{S}|} \\
& \sum_{i=1}^{n_s} \sum_{a \in \mathcal{A}, s' \in \mathcal{S}} q^{(s)}_{i,(a,s')} c((a,s'),(a_i,s_i')) \leq \rho_s, \ \forall s \in \mathcal{S} \\
& \sum_{a \in \mathcal{A}, s' \in \mathcal{S}} q^{(s)}_{i,(a,s')} = \frac{1}{n_s}, \forall i = 1, 2, \ldots, n_s, \forall s \in \mathcal{S}
\end{aligned}
$$

The following proposition reveals that the optimistic OPE is equivalent to the variation regularization problem.

Proposition 5. *Under the same conditions of Proposition 3, when $\rho_s < \overline{\rho}_s, s \in \mathcal{S}$, the optimistic OPE is equivalent to*

$$
\begin{aligned}
\max_{\kappa \geq 0} \min_{f \in \mathbb{R}^{|\mathcal{S}|}} \ & \mathbb{E}_{s\sim\kappa}[\mathbb{E}_{a\sim\pi(\cdot|s)}[r(s,a)]] + (1-\gamma)\mathbb{E}_{s\sim d_0}[f(s)] + \gamma \mathbb{E}_{s\sim\kappa}\left[ \mathbb{E}_{(a,s')\sim\hat{\mu}_s}[\beta_s(a) f(s')] \right] - \mathbb{E}_{s\sim\kappa}[f(s)] \\
& + \gamma \mathbb{E}_{s\sim\kappa(s)} \rho_s \cdot \|\beta_s f\|_{\mathrm{Lip},\hat{\mu}_s}.
\end{aligned}
$$

Similarly, for general radii $\{\rho_s\}_{s\in\mathcal{S}} \subset \mathbb{R}_+^{|\mathcal{S}|}$, the regularized maximin formulation above constitutes an upper bound for the optimistic OPE problem. Using the importance sampling technique, we solve the following problem instead:

$$\max_{w\geq 0}\ \min_{f\in\mathbb{R}^{|\mathcal{S}|}}\ (1-\gamma)\mathbb{E}_{s\sim d_0}[f(s)] \\ +\mathbb{E}_{(s,a,s')\sim\hat{\mu}}\big[w(s)(\gamma\beta_s(a)f(s')-f(s)+\mathbb{E}_{a_0\sim\pi(\cdot|s)}[r(s,a_0)]+\gamma\rho_s\cdot\|\beta_s f\|_{\mathrm{Lip},\hat{\mu}_s}])\big].$$

## Appendix B: Finite-sample Guarantees for Wasserstein DRO in Discrete Space

The results in this section are parallel to the results in Gao et al. (2020) which focuses on continuous space. Proofs are given in Appendix B.

We begin by introducing some notations in a general setup. Let $\mathbb{P}_n$ denote the empirical distribution constructed from $n$ i.i.d. samples from some underlying true distribution $\mathbb{P}_{\mathsf{true}}$ on a finite discrete metric space $\mathcal{Z}$ associated with a metric $c:\mathcal{Z}^2\to\mathbb{R}_+$. Let $\mathcal{F}$ be a class of functions on $\mathcal{Z}$. We define the *Wasserstein regularizer* for a function $f\in\mathcal{F}$ as

$$\mathcal{R}_n(\rho;f) := \sup_{\mathbb{P}}\big\{\mathbb{E}_{z\sim\mathbb{P}}[f(z)]:\ W(\mathbb{P},\mathbb{P}_n)\leq\rho\big\} - \mathbb{E}_{z\sim\mathbb{P}_n}[f(z)].$$

The *global slope* of a function $f\in\mathcal{F}$ at $z\in\mathcal{Z}$ is defined as

$$\mathfrak{l}_f(z) = \max_{\tilde{z}\neq z}\frac{f(\tilde{z})-f(z)}{c(\tilde{z},z)},$$

and the Lipschitz norm of $f$ with respect to $\mathbb{P}$ is defined as

$$\|f\|_{\mathrm{Lip},\mathbb{P}} = \max_{z\in\operatorname{supp}\mathbb{P}}\mathfrak{l}_f(z).$$

The following result shows the relationship between Wasserstein DRO in discrete space and Lipschitz regularization.

Proposition 6. *Define* $\mathcal{Z}_\infty := \{z\in\mathcal{Z}:\ \mathfrak{l}_f(z)=\|f\|_{\mathrm{Lip},\mathbb{P}_n}\}$ *and*

$$\bar{\rho} := \sup_{\mathfrak{T}:\mathcal{Z}\to\mathcal{Z}}\left\{\mathbb{E}_{\mathbb{P}_n}[c(\mathfrak{T}(z),z)]:\ \mathfrak{T}(z)=z,\forall z\notin\mathcal{Z}_\infty,\ \frac{f(\mathfrak{T}(z))-f(z)}{c(\mathfrak{T}(z),z)}=\mathfrak{l}_f(z),\forall z\in\mathcal{Z}_\infty\right\}.$$

*For any $\rho\geq 0$, it holds that*

$$\mathcal{R}_n(\rho;f)\leq\rho\cdot\|f\|_{\mathrm{Lip},\mathbb{P}_n}.$$

*For any $\rho<\bar{\rho}$, it holds that*

$$\mathcal{R}_n(\rho;f)=\rho\cdot\|f\|_{\mathrm{Lip},\mathbb{P}_n}.$$

Next, we study the finite-sample performance guarantee for Wasserstein DRO in discrete space. We introduce a metric $\mathsf{d}$ on $\mathcal{F}$

$$\mathsf{d}(\tilde{f}, f) := \|\tilde{f} - f\|_\infty \vee \|\tilde{f} - f\|_{\mathrm{lip}}, \quad \forall \tilde{f}, f \in \mathcal{F},$$

where $\|\cdot\|_\infty$ and $\|\cdot\|_{\mathrm{Lip}}$ denotes the sup-norm and Lipschitz norm respectively. Denote by $\mathcal{N}(\epsilon, \mathcal{F}, \mathsf{d})$ the *$\epsilon$-covering number* of $\mathcal{F}$ under the metric $\mathsf{d}$, defined as the smallest cardinality of an $\epsilon$-cover of $\mathcal{F}$, where the collection of functions $\mathcal{F}_\epsilon$ is called an $\epsilon$-cover of $\mathcal{F}$ if for any $f \in \mathcal{F}$, there exists $f' \in \mathcal{F}_\epsilon$ so that $\mathsf{d}(f, f') \leq \epsilon$. Finally, let $H(a\|b)$ denote the binary relative entropy function $H(a\|b) := a \log \frac{a}{b} + (1-a) \log \frac{1-b}{1-a}$.

Assumption 4. *Suppose $\mathcal{F}$ satisfies the following conditions:*

(I) *There exists $L > 0$ so that $f(\tilde{z}) - f(z) \leq Lc(\tilde{z}, z)$ for all $z, \tilde{z} \in \mathcal{Z}$ and all $f \in \mathcal{F}$.*

(II) *There exists $\eta \in (0, 1]$ such that $\delta := \inf_{f \in \mathcal{F}} \mathbb{P}_{\mathsf{true}}\big\{z \in \mathcal{Z} :\ \mathfrak{l}_f(z) \geq \eta \|f\|_{\mathrm{Lip}, \mathbb{P}_{\mathsf{true}}}\big\} \in (0, 1)$.*

Proposition 7. *Assume Assumption 4 holds. Suppose that $\rho_n = \rho_0/\sqrt{n}$ for some $\rho_0 > 0$. Let $c < \frac{\delta}{1-\delta} \wedge \frac{1-\delta}{\delta}$. Then with probability at least $1 - \exp\big(-nH\big(c\| \frac{\delta}{1-\delta} \wedge \frac{1-\delta}{\delta}\big) + \log \mathcal{N}(\frac{1}{n}, \mathcal{F}, \mathsf{d})\big)$, simultaneously for all $f \in \mathcal{F}$,*

$$\eta \rho_n \|f\|_{\mathrm{Lip}, \mathbb{P}_n} - \tfrac{3}{n} \leq \mathcal{R}_n(\rho_n; f) \leq \rho_n \|f\|_{\mathrm{Lip}, \mathbb{P}_n}$$

Note that this result appears in (Gao et al. 2020, Theorem 2) in a slightly different form for a continuous space.

Next, we establish the out-of-sample performance guarantee for Wasserstein DRO on a discrete space. Comparing with the discussion in Gao et al. (2020), our analysis is easier since the variation only relies on the global slope of a function. Define $\mathrm{diam}(\mathcal{Z}) := \max_{\tilde{z}, z \in \mathcal{Z}} c(\tilde{z}, z)$.

Proposition 8. *Assume Assumption 4 holds. Let $\tau > 0$ and $\Delta = \frac{1}{\left(\frac{1-2\delta}{1-\delta} \vee \frac{2\delta-1}{\delta}\right)}$. Set $\rho_n = \frac{\mathrm{diam}(\mathcal{Z})}{\eta^2} \sqrt{2\tau/n}$. Then with probability at least*

$$1 - \mathcal{N}\left(\tfrac{1}{n}, \mathcal{F}, \mathsf{d}\right) \Big[ e^{-\tau} + \exp\left(-n \log(1/(1-\delta))\right) + \exp\left(-n \log \Delta\right) \Big],$$

*simultaneously for all $f \in \mathcal{F}$, it holds that*

$$\mathbb{E}_{z \sim \mathbb{P}_{\mathsf{true}}}[f(z)] \leq \mathbb{E}_{z \sim \mathbb{P}_n}[f(z)] + \mathcal{R}_n\left(\rho_n; f\right) + \frac{6}{n}.$$

This indicates that with high probability, by setting the radius in the order of $O(\mathrm{diam}(\mathcal{Z})/\sqrt{n})$, the Wasserstein robust loss dominates the true loss up to a higher order remainder. There is a trade-off between $\eta$, which affects the radius, and $\delta$, which affects the error probability bound, in Proposition 8

— one can reduce $\eta$ to achieve a larger $\delta$ (thus a smaller error probability bound). When the global slope of loss functions are much smaller than their Lipschitz norm with high probability, to keep the error probability bound at a reasonably small level, we would like to reduce the fraction $\eta$ so that $\delta$ in Assumption 4(II) would not be too small. Note that if one choose $\rho_n$ using the principle in Esfahani and Kuhn (2018), namely, a high confidence bound on the Wasserstein distance between $\mathbb{P}_n$ and $\mathbb{P}_{\text{true}}$, then one would only obtain a much worse bound $O(|\mathcal{Z}|/\sqrt{n})$ that linearly depends on the carnality of $\mathcal{Z}$ (Singh and Póczos 2019). This is the curse of dimension in discrete settings: imagine $\mathcal{Z}$ is an $\epsilon$-covering of a $k$-dimensional unit box, then $|\mathcal{Z}| = O(\epsilon^{-k})$. Our bound does not suffer from the curse of dimensionality which, to the best of knowledge, is the first result of this kind for Wasserstein DRO on a discrete space. The difference between this result and the result on a continuous space (Gao et al. 2020) is that local slope does not serve as the regularization term, which simplifies the analysis on the out-of-sample performance guarantee. We will instantiate this result on OPE and show that the tail probability has a mild dependence on $n$.

Finally, we discuss the generalization bound for Wasserstein DRO with fixed radius. This generalization bound is developed by selecting the fraction parameter $\eta = 1$ in Assumption 4. Define the following two risk functions in which the radius is $\rho$ and the nominal distributions are different:

$$U^*(\rho; f) = \sup_{\mathbb{P}} \left\{ \mathbb{E}_{\mathbb{P}}[f(z)] : \ W(\mathbb{P}, \mathbb{P}_{\text{true}}) \le \rho \right\}, \quad U_n(\rho; f) = \sup_{\mathbb{P}} \left\{ \mathbb{E}_{\mathbb{P}}[f(z)] : \ W(\mathbb{P}, \mathbb{P}_n) \le \rho \right\}.$$

Proposition 9. *Let $\tau > 0$ and $\delta = \min_{z \in \mathcal{Z}} \mathbb{P}_{\text{true}}(z)$. Suppose the function $f$ satisfies $0 \le f(z) \le M, \forall z \in \mathcal{Z}$ and the radius $\rho$ satisfies*

$$\rho < \overline{\rho} = \sup_{\mathfrak{T}: \mathcal{Z} \to \mathcal{Z}} \left\{ \mathbb{E}_{\mathbb{P}_n}[c(\mathfrak{T}(z), z)] : \ \mathfrak{T}(z) = z, \forall z \notin \mathcal{Z}_\infty, \ \frac{f(\mathfrak{T}(z)) - f(z)}{c(\mathfrak{T}(z), z)} = \mathfrak{l}_f(z), \forall z \in \mathcal{Z}_\infty \right\}.$$

*Then with probability at least $1 - \mathcal{N}\left(\frac{1}{n}, \mathcal{F}, \mathsf{d}\right) e^{-n \log\left(\frac{1}{1-\delta}\right)} - 2e^{-\tau}$, simultaneously for all $f \in \mathcal{F}$, it holds that*

$$|U^*(\rho; f) - U_n(\rho; f)| \le 2\mathbb{E}_{\otimes}[\mathfrak{R}_n(\mathcal{F})] + M\sqrt{\frac{\tau}{2n}}.$$

## References

Abdullah MA, Ren H, Ammar HB, Milenkovic V, Luo R, Zhang M, Wang J (2019) Wasserstein robust reinforcement learning. *arXiv preprint arXiv:1907.13196* .

Agrawal A, Verschueren R, Diamond S, Boyd S (2018) A rewriting system for convex optimization problems. *Journal of Control and Decision* 5(1):42–60.

Ambrosio L, Gigli N, Savaré G (2008) *Gradient flows: in metric spaces and in the space of probability measures* (Springer Science & Business Media).

Billingsley P (1961) Statistical methods in markov chains. *The Annals of Mathematical Statistics* 12–40.

Blanchet J, Murthy K (2019) Quantifying distributional model risk via optimal transport. *Mathematics of Operations Research* 44(2):565–600.

Buckman J, Gelada C, Bellemare MG (2021) The importance of pessimism in fixed-dataset policy optimization. *International Conference on Learning Representations*.

Chen M, Jiang H, Liao W, Zhao T (2019a) Efficient approximation of deep relu networks for functions on low dimensional manifolds. *Advances in Neural Information Processing Systems*, volume 32, 8172–8182.

Chen M, Jiang H, Liao W, Zhao T (2022) Nonparametric regression on low-dimensional manifolds using deep relu networks: function approximation and statistical recovery. *Information and Inference: A Journal of the IMA* .

Chen M, Liao W, Zha H, Zhao T (2020a) Statistical guarantees of generative adversarial networks for distribution estimation. *arXiv preprint arXiv: 2002.03938* .

Chen X, Wang L, Hang Y, Ge H, Zha H (2020b) Infinite-horizon off-policy policy evaluation with multiple behavior policies. *International Conference on Learning Representations*.

Chen Z, Yu P, Haskell WB (2019b) Distributionally robust optimization for sequential decision-making. *Optimization* 68(12):2397–2426.

Dai B, Nachum O, Chow Y, Li L, Szepesvari C, Schuurmans D (2020) Coindice: Off-policy confidence interval estimation. *Advances in Neural Information Processing Systems*, volume 33, 9398–9411.

Diamond S, Boyd S (2016) CVXPY: A Python-embedded modeling language for convex optimization. *Journal of Machine Learning Research* 17(83):1–5.

Du SS, Zhai X, Poczos B, Singh A (2019) Gradient descent provably optimizes over-parameterized neural networks. *International Conference on Learning Representations*.

Duan Y, Jia Z, Wang M (2020) Minimax-optimal off-policy evaluation with linear function approximation. *International Conference on Machine Learning*, volume 119, 2701–2709.

Esfahani PM, Kuhn D (2018) Data-driven distributionally robust optimization using the wasserstein metric: Performance guarantees and tractable reformulations. *Mathematical Programming* 171(1-2):115–166.

Farajtabar M, Chow Y, Ghavamzadeh M (2018) More robust doubly robust off-policy evaluation. *International Conference on Machine Learning*, volume 80, 1447–1456.

Feng Y, Ren T, Tang Z, Liu Q (2020) Accountable off-policy evaluation with kernel bellman statistics. *International Conference on Machine Learning*, volume 119, 3102–3111.

Gao R (2022) Finite-sample guarantees for wasserstein distributionally robust optimization: Breaking the curse of dimensionality. *Operations Research* .

Gao R, Chen X, Kleywegt AJ (2020) Wasserstein distributionally robust optimization and variation regularization. *arXiv preprint arXiv:1712.06050* .

Gao R, Kleywegt A (2022) Distributionally robust stochastic optimization with wasserstein distance. *Mathematics of Operations Research* .

Goodfellow I, Pouget-Abadie J, Mirza M, Xu B, Warde-Farley D, Ozair S, Courville A, Bengio Y (2014) Generative adversarial nets. *Advances in neural information processing systems*, volume 27, 2672–2680.

Gottesman O, Johansson F, Komorowski M, Faisal A, Sontag D, Doshi-Velez F, Celi LA (2019) Guidelines for reinforcement learning in healthcare. *Nat Med* 25(1):16–18.

Goyal V, Grand-Clement J (2022) Robust markov decision processes: Beyond rectangularity. *Mathematics of Operations Research* .

Gurobi Optimization, LLC (2022) Gurobi Optimizer Reference Manual. URL https://www.gurobi.com.

Hanin B (2019) Universal function approximation by deep neural nets with bounded width and relu activations. *Mathematics* 7(10).

Hanna JP, Stone P, Niekum S (2017) Bootstrapping with models: Confidence intervals for off-policy evaluation. *Thirty-First AAAI Conference on Artificial Intelligence*, volume 31.

Ho CP, Petrik M, Wiesemann W (2021) Partial policy iteration for l1-robust markov decision processes. *Journal of Machine Learning Research* 22(275):1–46.

Hou L, Pang L, Hong X, Lan Y, Ma Z, Yin D (2020) Robust reinforcement learning with wasserstein constraint. *arXiv preprint arXiv:2006.00945* .

Iyengar GN (2005) Robust dynamic programming. *Mathematics of Operations Research* 30(2):257–280.

Jacot A, Gabriel F, Hongler C (2018) Neural tangent kernel: Convergence and generalization in neural networks. *Advances in Neural Information Processing Systems*, volume 31.

Jiang N, Huang J (2020) Minimax value interval for off-policy evaluation and policy optimization. *Advances in Neural Information Processing Systems* 33:2747–2758.

Jiang N, Li L (2016) Doubly robust off-policy value evaluation for reinforcement learning. *International Conference on Machine Learning*, volume 48, 652–661.

Jin Y, Yang Z, Wang Z (2021) Is pessimism provably efficient for offline rl? *International Conference on Machine Learning*, volume 139, 5084–5096.

Kallus N, Uehara M (2020) Double reinforcement learning for efficient off-policy evaluation in markov decision processes. *Journal of Machine Learning Research*, volume 21, 1–63.

Kallus N, Uehara M (2022) Efficiently breaking the curse of horizon: Double reinforcement learning in infinite-horizon processes. *Operations Research* .

Kidambi R, Rajeswaran A, Netrapalli P, Joachims T (2020) Morel: Model-based offline reinforcement learning. *Advances in neural information processing systems* 33:21810–21823.

Kober J, Bagnell JA, Peters J (2013) Reinforcement learning in robotics: A survey. *The International Journal of Robotics Research* 32(11):1238–1274.

Kosorok MR (2008) *Introduction to empirical processes and semiparametric inference.* (Springer).

Kostrikov I, Nachum O (2020) Statistical bootstrapping for uncertainty estimation in off-policy evaluation. *arXiv preprint arXiv:2007.13609* .

Kumar A, Zhou A, Tucker G, Levine S (2020) Conservative q-learning for offline reinforcement learning. *Advances in Neural Information Processing Systems* 33:1179–1191.

Lam H, Qian H (2017) Optimization-based quantification of simulation input uncertainty via empirical likelihood. *arXiv preprint arXiv:1707.05917* .

Lam H, Zhou E (2017) The empirical likelihood approach to quantifying uncertainty in sample average approximation. *Operations Research Letters* 45(4):301–307.

Liu F, Tang R, Li X, Ye Y, Chen H, Guo H, Zhang Y (2018a) Deep reinforcement learning based recommendation with explicit user-item interactions modeling. *arXiv preprint arXiv:1810.12027* .

Liu Q, Li L, Tang Z, Zhou D (2018b) Breaking the curse of horizon: Infinite-horizon off-policy estimation. *Advances in Neural Information Processing Systems*, volume 31, 5356–5366.

Mandel T, Liu YE, Levine S, Brunskill E, Popovic Z (2014) Offline policy evaluation across representations with applications to educational games. *Proceedings of the 13th International Conference on Autonomous Agents and Multiagent Systems*, volume 13, 1077–1084.

Mannor S, Mebel O, Xu H (2016) Robust mdps with k-rectangular uncertainty. *Mathematics of Operations Research* 41:1484–1509.

Matsushima T, Furuta H, Matsuo Y, Nachum O, Gu S (2021) Deployment-efficient reinforcement learning via model-based offline optimization. *International Conference on Learning Representations*.

Mei S, Misiakiewicz T, Montanari A (2019) Mean-field theory of two-layers neural networks: dimension-free bounds and kernel limit. *Proceedings of the Thirty-Second Conference on Learning Theory*, volume 99, 2388–2464.

Mei S, Montanari A, Nguyen PM (2018) A mean field view of the landscape of two-layer neural networks. *Proceedings of the National Academy of Sciences* 115(33):7665–7671.

Mnih V, Kavukcuoglu K, Silver D, Rusu AA, Veness J, Bellemare MG, Graves A, Riedmiller M, Fidjeland AK, Ostrovski G, Petersen S, Beattie C, Sadik A, Antonoglou I, King H, Kumaran D, Wierstra D, Legg S, Hassabis D (2015) Human-level control through deep reinforcement learning. *Nature* 518(7540):529–533.

Mousavi A, Li L, Liu Q, Zhou D (2020) Black-box off-policy estimation for infinite-horizon reinforcement learning. *International Conference on Learning Representations*.

Munos R (2014) From bandits to monte-carlo tree search: The optimistic principle applied to optimization and planning. *Foundations and Trends in Machine Learning* 7(1):1–129.

Nachum O, Chow Y, Dai B, Li L (2019) Dualdice: Behavior-agnostic estimation of discounted stationary distribution corrections. *Advances in Neural Information Processing Systems* 32.

Nilim A, El Ghaoui L (2005) Robust control of markov decision processes with uncertain transition matrices. *Operations Research* 53(5):780–798.

Nilim A, Ghaoui LE (2005) Robust control of markov decision processes with uncertain transition matrices. *Operations Research* 53(5):780–798.

OpenAI, Berner C, Brockman G, Chan B, Cheung V, Dębiak P, Dennison C, Farhi D, Fischer Q, Hashme S, Hesse C, Józefowicz R, Gray S, Olsson C, Pachocki J, Petrov M, de Oliveira Pinto HP, Raiman J, Salimans T, Schlatter J, Schneider J, Sidor S, Sutskever I, Tang J, Wolski F, Zhang S (2019) Dota 2 with large scale deep reinforcement learning. *arXiv preprint arXiv:1912.06680* .

Petrik M, Russel RH (2019) Beyond confidence regions: Tight bayesian ambiguity sets for robust mdps. *Advances in Neural Information Processing Systems*, volume 32, 7049–7058.

Precup D (2000) Eligibility traces for off-policy policy evaluation. *International Conference on Machine Learning*, 759–766.

Puterman ML (1994) *Markov Decision Processes: Discrete Stochastic Dynamic Programming* (John Wiley & Sons, Inc.).

Raghu A, Komorowski M, Ahmed I, Celi L, Szolovits P, Ghassemi M (2017) Deep reinforcement learning for sepsis treatment. *arXiv preprint arXiv:1711.09602* .

Sallab AE, Abdou M, Perot E, Yogamani SK (2017) Deep reinforcement learning framework for autonomous driving. *Electronic Imaging* 2017(19):70–76.

Shapiro A, Dentcheva D, Ruszczynski A (2021) *Lectures on stochastic programming: modeling and theory* (Society for Industrial and Applied Mathematics), 3 edition.

Shi C, Zhang S, Lu W, Song R (2020) Statistical inference of the value function for reinforcement learning in infinite horizon settings. *arXiv preprint arXiv:2001.04515* .

Si N, Zhang F, Zhou Z, Blanchet J (2020) Distributional robust batch contextual bandits. *International Conference on Machine Learning*, volume 119, 8884–8894.

Singh S, Póczos B (2019) Minimax distribution estimation in wasserstein distance. *arXiv preprint arXiv:1802.08855* .

Sion M (1958) On general minimax theorems. *Pacific Journal of mathematics* 8(1):171–176.

Sirignano J, Spiliopoulos K (2020) Mean field analysis of neural networks: A law of large numbers. *SIAM Journal on Applied Mathematics* 80(2):725–752.

Smirnova E, Dohmatob E, Mary J (2019) Distributionally robust reinforcement learning. *arXiv preprint arXiv:1902.08708* .

Sonabend A, Lu J, Celi LA, Cai T, Szolovits P (2020) Expert-supervised reinforcement learning for offline policy learning and evaluation. *Advances in Neural Information Processing Systems* 33:18967–18977.

Song J, Zhao C (2020) Optimistic distributionally robust policy optimization. *arXiv preprint arXiv:2006.07815* .

Sun R (2019) Optimization for deep learning: theory and algorithms. *arXiv preprint arXiv:1912.08957* .

Tang Z, Feng Y, Li L, Zhou D, Liu Q (2019) Doubly robust bias reduction in infinite horizon off-policy estimation. *International Conference on Learning Representations*.

Thomas PS, Theocharous G, Ghavamzadeh M (2015) High-confidence off-policy evaluation. *AAAI Conference on Artificial Intelligence*, volume 29.

Thomas PS, Theocharous G, Ghavamzadeh M, Durugkar I, Brunskill E (2017) Predictive off-policy policy evaluation for nonstationary decision problems, with applications to digital marketing. *AAAI Conference on Artificial Intelligence*, volume 31, 4740–4745.

Tirinzoni A, Chen X, Petrik M, Ziebart BD (2018) Policy-conditioned uncertainty sets for robust markov decision processes. *Advances in neural information processing systems* 31:8953–8963.

Uehara M, Huang J, Jiang N (2020a) Minimax weight and q-function learning for off-policy evaluation. *International Conference on Machine Learning*, 9659–9668.

Uehara M, Kato M, Yasui S (2020b) Off-policy evaluation and learning for external validity under a covariate shift. *Advances in Neural Information Processing Systems*, volume 33, 49–61.

Wang L, Zhang W, He X, Zha H (2018) Supervised reinforcement learning with recurrent neural network for dynamic treatment recommendation. *Proceedings of the 24th ACM SIGKDD International Conference on Knowledge Discovery & Data Mining*, 2447–2456.

Wiesemann W, Kuhn D, Rustem B (2013) Robust markov decision processes. *Mathematics of Operations Research* 38:153–183.

Xie T, Ma Y, Wang YX (2019) Towards optimal off-policy evaluation for reinforcement learning with marginalized importance sampling. *Advances in Neural Information Processing Systems*, volume 32, 9668–9678.

Xu H, Mannor S (2010) Distributionally robust markov decision processes. *Advances in Neural Information Processing Systems*, volume 23, 2505–2513.

Yang I (2017) A convex optimization approach to distributionally robust markov decision processes with wasserstein distance. *IEEE control systems letters* 1(1):164–169.

Yang M, Nachum O, Dai B, Li L, Schuurmans D (2020) Off-policy evaluation via the regularized lagrangian. *Advances in Neural Information Processing Systems* 33:6551–6561.

Yu T, Thomas G, Yu L, Ermon S, Zou JY, Levine S, Finn C, Ma T (2020) Mopo: Model-based offline policy optimization. *Advances in Neural Information Processing Systems* 33:14129–14142.

Zhang R, Dai B, Li L, Schuurmans D (2020a) Gendice: Generalized offline estimation of stationary values. *International Conference on Learning Representations*.

Zhang Y, Cai Q, Yang Z, Chen Y, Wang Z (2020b) Can temporal-difference and q-learning learn representation? a mean-field theory. *Advances in Neural Information Processing Systems* 33:19680–19692.

# Supplementary for "Reliable Off-policy Evaluation for Reinforcement Learning"

## Appendix EC.1: Detailed Experiment Setup

All the experiments are performed on a MacBook Pro laptop with 32GB of memory running python 3.7 with CVXPY optimization solver (Diamond and Boyd 2016, Agrawal et al. 2018) and Gurobi package (Gurobi Optimization, LLC 2022). Values of radii for DRO models are listed as follows. As we can see, the values of radii in Figure EC.1 generally scale in the order of $1/\sqrt{n}$ in terms of the sample size $n$, which is consistent with the choice of radii listed in Remark 1. Moreover, the width of intervals in Figure 1 decreases as the radii decrease, which is consistent with the theoretical upper bound on the length of proposed CI in Proposition 4.

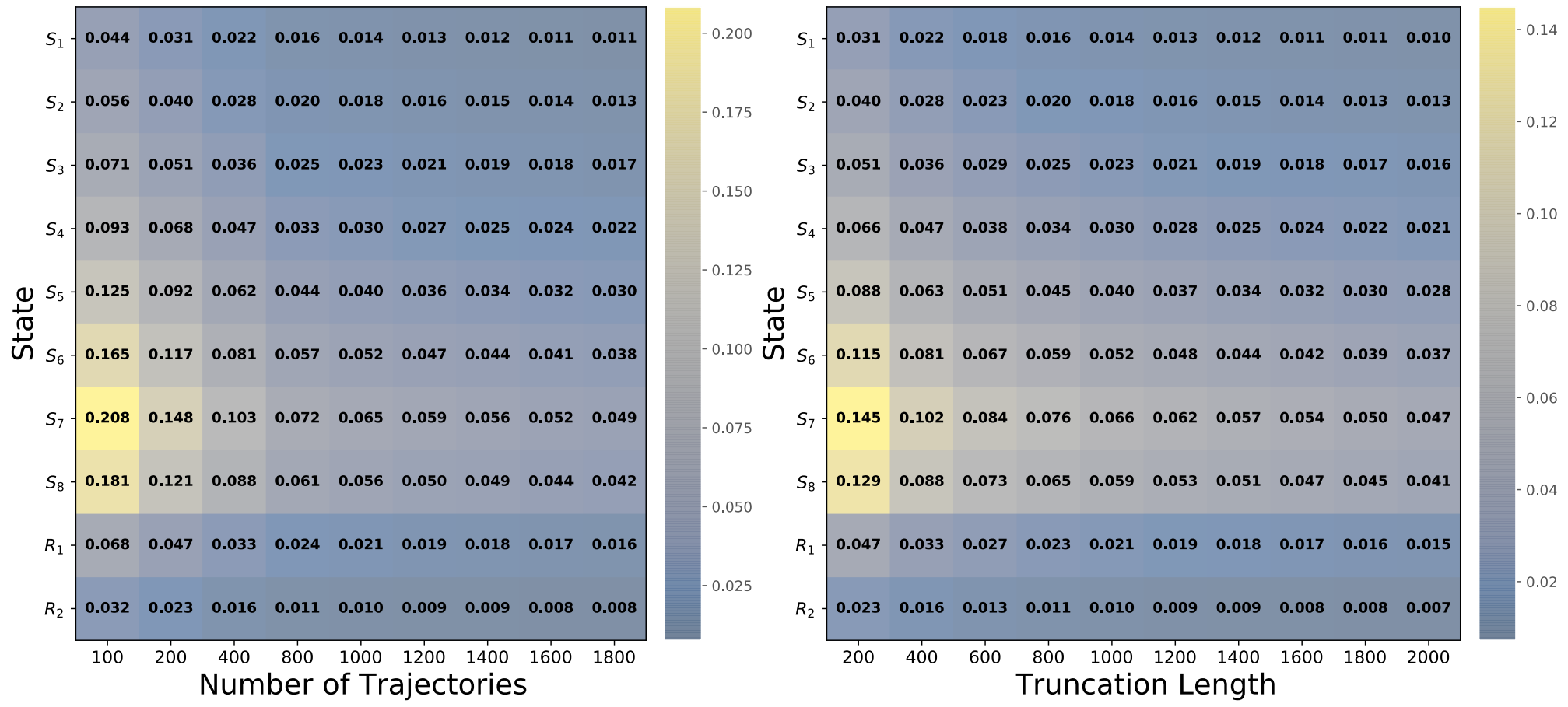

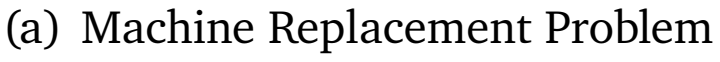
(a) Machine Replacement Problem

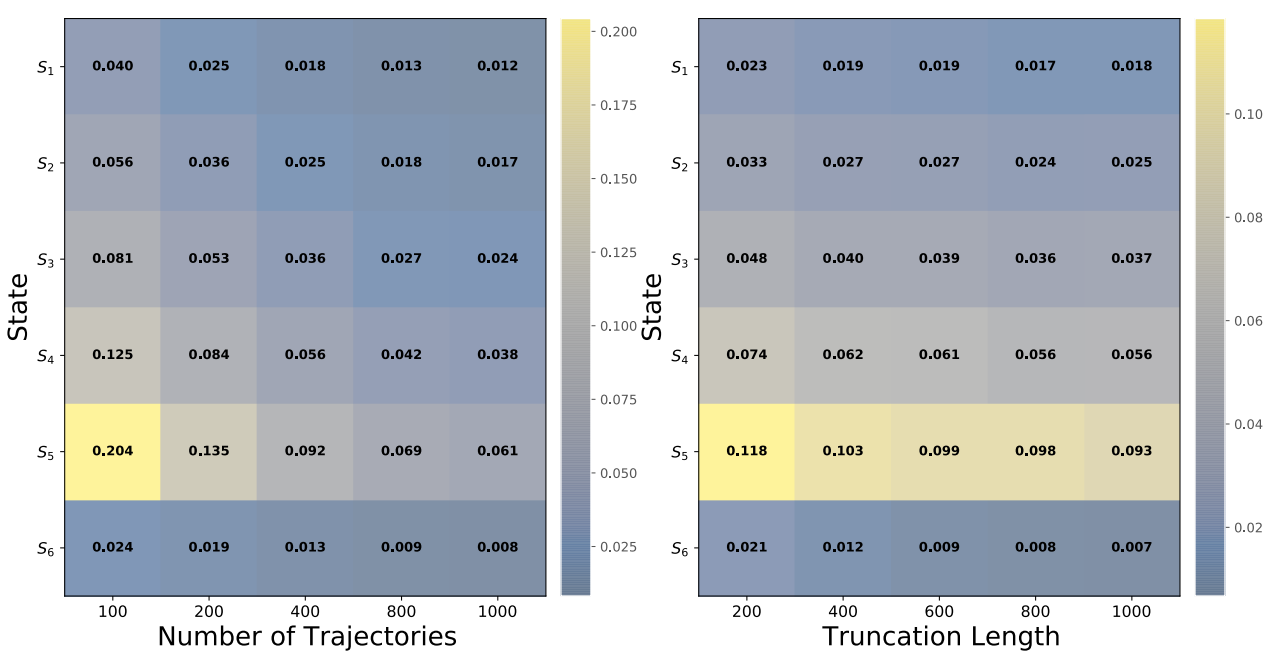

(b) Healthcare Management Problem

**Figure EC.1** Values of radii for constructing 95% confidence intervals with non-stationary trajectory data in Figure 1. The $y$-axis represents states of MDP environments, and the $x$-axis represents different numbers of trajectories or different truncation lengths.

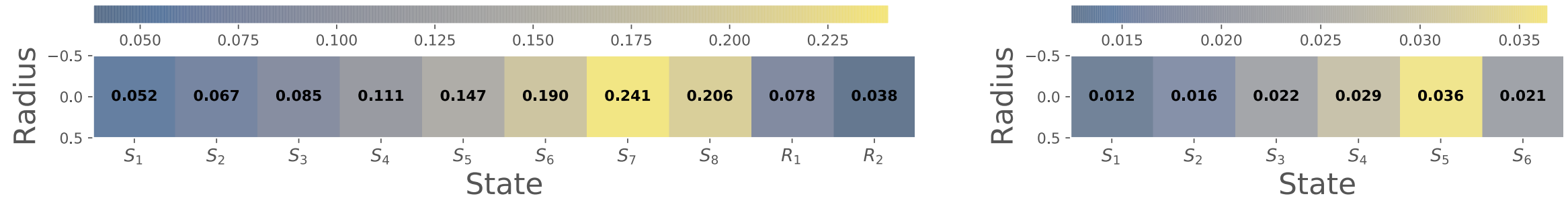

(a) Machine Replacement Problem (b) Healthcare Management Problem

**Figure EC.2** Values of radii for constructing 95% confidence intervals under changing MDP environments in Figure 3. The $x$-axis represents states of MDP environments, and the $y$-axis represents values of radii.

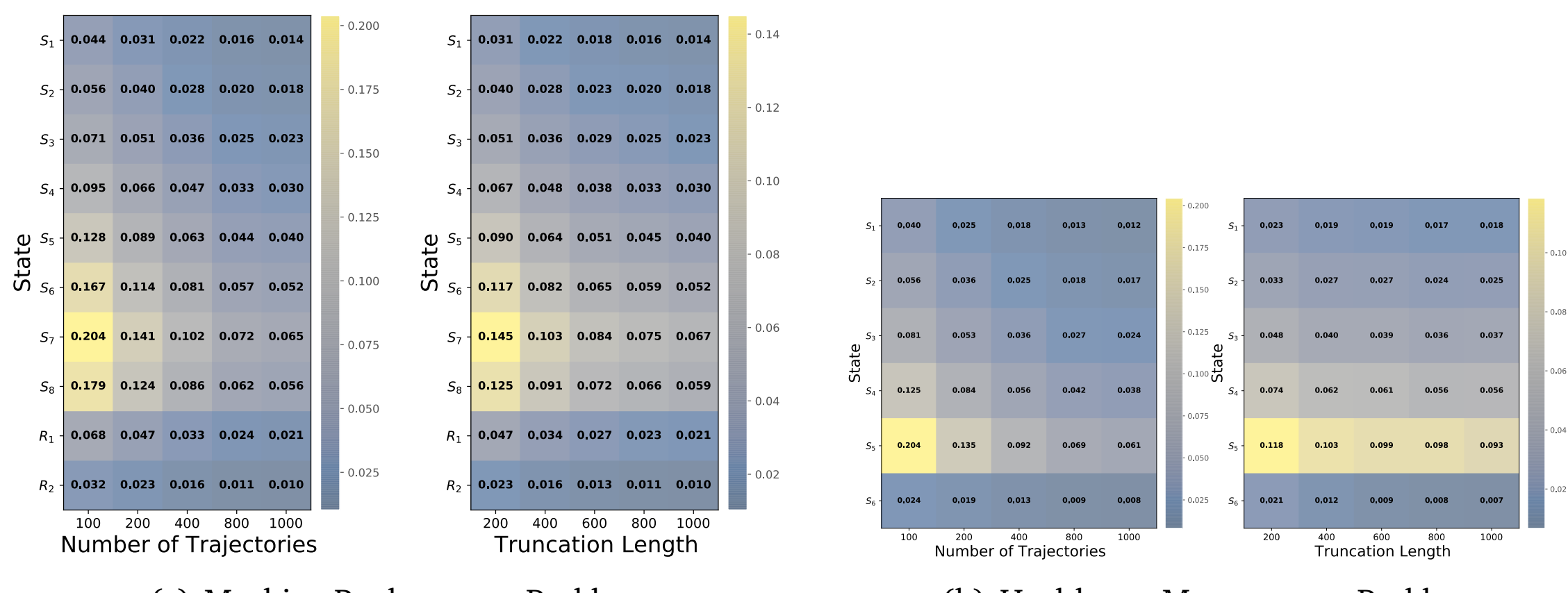

(a) Machine Replacement Problem (b) Healthcare Management Problem

**Figure EC.3** Values of radii for the task of distributionally robust batch reinforcement learning in Figure 4. The $y$-axis represents states of MDP environments, and the $x$-axis represents different numbers of trajectories or different truncation lengths.

## Appendix EC.2: Proofs for Section 2

*Proof of Lemma 1.* By the definition of average visitation distribution $d_\pi$ in (1), it can be shown that the following linear system holds (Liu et al. 2018b, Lemma 3):

$$d_\pi(s') = (1-\gamma)d_0(s') + \gamma \sum_{(s,a)} P(s' \mid s,a)\pi(a \mid s)d_\pi(s), \forall s' \in \mathcal{S}.$$

Substituting $d_\pi(s)$ with $w(s)d_{\pi_b}(s)$ in the equalities above gives

$$w(s')d_{\pi_b}(s') = (1-\gamma)d_0(s') + \gamma \sum_{(s,a)} P(s' \mid s,a)\pi(a \mid s)w(s)d_{\pi_b}(s), \quad \forall s' \in \mathcal{S}.$$

This, together with the relation $d_{\pi_b}(s,a,s') = P(s' \mid s,a)\pi_b(a \mid s)d_{\pi_b}(s)$ completes the proof. □

## Appendix EC.3: Proofs for Section 4

### EC.3.1. Proof of Theorem 1

We need several technical lemmas before the proof of Theorem 1.

Lemma EC.1. *For fixed state* $s \in \mathcal{S}$*, value function* $v$*, and the empirical distribution* $\hat{\mu}(\cdot,\cdot \mid s) = \frac{1}{n_s}\sum_{i=1}^{n_s} \delta_{(a_i,s_i')}$*, the optimization*

$$\min_{\mu \in \mathfrak{M}_{\hat{\mu}}(\rho)} \sum_{(a,s')} \mu(a,s' \mid s) v(s') \beta_s(a) \tag{EC.1}$$

*can be equivalently formulated as*

$$\min_{\mu_i(\cdot,\cdot),\ i=1,2,\ldots,n_s} \left\{ \frac{1}{n_s}\sum_{i=1}^{n_s} \mu_i(a,s')v(s')\beta_s(a) :\ \frac{1}{n_s}\sum_{i=1}^{n_s} \mu_i(a,s') c((a,s'),(a_i,s_i')) \le \rho \right\},$$

*where* $c((a_1,s_1'),(a_2,s_2'))$ *denotes the transportation cost between action-state pairs* $(a_1,s_1')$ *and* $(a_2,s_2')$*. The optimal value can also be computed from the one-dimensional dual problem:*

$$\max_{\kappa \ge 0} \left\{ -\kappa\rho + \frac{1}{n_s}\sum_{i=1}^{n_s} \min_{(a,s')} \left\{ v(s')\beta_s(a) + \kappa c((a,s'),(a_i,s_i')) \right\} \right\}.$$

*Proof.* The problem (EC.1) can be viewed as the minimization of expectation with respect to the probability measure within a Wasserstein ball. Applying duality results in Blanchet and Murthy (2019) and Gao and Kleywegt (2022) completes the proof. □

Lemma EC.2. *The min-min problem in* (P) *can be equivalently formulated as:*

$$\min_{\kappa,\mu} \quad \sum_s \kappa(s) \sum_a \pi_b(a \mid s)\beta_s(a) r(s,a) = \sum_s \kappa(s) \sum_a \pi(a \mid s) r(s,a) \tag{EC.2a}$$

$$\text{subject to} \quad \kappa(s') = (1-\gamma)d_0(s') + \gamma \cdot \sum_s \kappa(s) \left[ 1(\mu(s) > 0) \sum_a \frac{\mu(s,a,s')}{\mu(s)} \beta_s(a) \right], \quad \forall s' \in \mathcal{S}, \tag{EC.2b}$$

$$\mu \in \mathfrak{M}_{\hat{\mu}}(\rho). \tag{EC.2c}$$

*Note that this equivalence is independent of the structure of the ambiguity set.*

*Proof.* The result follows simply by the change-of-variable technique with $\kappa(s) = w(s)\mu(s)$. □

In particular, when $\mu(s_0) = 0$ for some $s_0 \in \mathcal{S}$, by the stationarity constraint (EC.2b) we can assert that $\kappa(s_0) = (1-\gamma)d_0(s_0)$. After substituting $\kappa(s) = (1-\gamma)d_0(s_0)$ for all $s$ in the set $\bar{\mathcal{S}} = \{s \in \mathcal{S} :\ \mu(s) = 0\}$, the problem is reduced into a problem with smaller size so that the decision variable becomes $\{\kappa(s), s \in \mathcal{S} \setminus \bar{\mathcal{S}}\}$. Without loss of generality, we can assume that for any $\mu \in \mathcal{P}$, the marginal distribution for state $\mu(s) > 0$ for any $s \in \mathcal{S}$. Then the indicator term in (EC.2b) can be omitted, and we denote the fraction $\frac{\mu(s,a,s')}{\mu(s)}$ as the conditional probability $\mu(a,s' \mid s)$ for simplicity.

Taking the duality for the minimization over $\mu$ in the problem (EC.2), we reformulate it as a min-max problem.

Lemma EC.3. *The min-min problem in* (EC.2) *can be equivalently formulated as:*

$$\min_{\mu} \max_{v} \quad (1-\gamma)\sum_s v(s) d_0(s) \tag{EC.3a}$$

$$s.t. \quad v(s) \le \sum_a \pi(a \mid s) r(s,a) + \gamma \sum_{(a,s')} \mu(a,s' \mid s) v(s') \beta_s(a), \quad \forall s \in \mathcal{S}, \tag{EC.3b}$$

$$\mu \in \mathfrak{M}_{\hat{\mu}}(\rho). \tag{EC.3c}$$

The following lemma from Nilim and El Ghaoui (2005) is useful for the reformulation of the min-max problem (EC.3).

Lemma EC.4. *Let $c \in \mathbb{R}_+^n$ and $f:\ \mathbb{R}_+^n \to \mathbb{R}_+^n$ be a component-wise non-decreasing contraction mapping. Then*

$$\left\{\begin{array}{cc} \max & c^T x \\ \textit{subject to} & x \le f(x) \end{array}\right\} = c^T x^*,$$

*where $x^*$ is the fixed point of the contraction mapping $f$, i.e. $x^* = f(x^*)$.*

Define a mapping $\mathcal{T}_{\mathsf{rob}} : \mathbb{R}^{|\mathcal{S}|} \to \mathbb{R}_+^{|\mathcal{S}|}$ as

$$\mathcal{T}_{\mathsf{rob}}[v] = \left( \min_{\mu\in\mathfrak{M}_{\hat{\mu}}(\rho)} \textstyle\sum_a \pi(a \mid s) r(s,a) + \gamma \sum_{(a,s')} \mu(a,s' \mid s) v(s') \beta_s(a) \right)_{s\in\mathcal{S}}, \quad v \in \mathbb{R}^{|\mathcal{S}|}. \tag{EC.4}$$

Lemma EC.5. *Under the setting of Theorem 1, $\mathcal{T}_{\mathsf{rob}}$ is a component-wise non-decreasing, and is contractive with constant $\frac{1+\gamma}{2}$ with probability at least $1 - \sum_{s\in\mathcal{S}} \exp\left(-\frac{2n_s\epsilon_s^2}{M_s^2}\right)$.*

*Proof.* The mapping $\mathcal{T}_{\mathsf{rob}}$ is component-wise non-decreasing because of the non-negativity of $\{\mu(a,s' \mid s)\beta_s(a)\}_{a,s'}$ for any fixed $s$. For any $v_1, v_2 \in \mathbb{R}_+^{|\mathcal{S}|}$ and $s \in \mathcal{S}$, it holds that

$$\begin{aligned}
\mathcal{T}_{\mathsf{rob}}(v_1)_s &= \sum_a \pi(a \mid s) r(s,a) + \gamma \min_{\mu\in\mathfrak{M}_{\hat{\mu}}(\rho)} \sum_{a,s'} \mu(a,s' \mid s) v_1(s') \beta_s(a) \\
&= \sum_a \pi(a \mid s) r(s,a) + \gamma \min_{\mu\in\mathfrak{M}_{\hat{\mu}}(\rho)} \left\{ \sum_{a,s'} \mu(a,s' \mid s) v_2(s') \beta_s(a) + \sum_{a,s'} \mu(a,s' \mid s) [v_1(s') - v_2(s')] \beta_s(a) \right\} \\
&\ge \sum_a \pi(a \mid s) r(s,a) + \gamma \min_{\mu\in\mathfrak{M}_{\hat{\mu}}(\rho)} \sum_{a,s'} \mu(a,s' \mid s) v_2(s') \beta_s(a) \\
&\qquad + \gamma \min_{\mu\in\mathfrak{M}_{\hat{\mu}}(\rho)} \sum_{a,s'} \mu(a,s' \mid s) [v_1(s') - v_2(s')] \beta_s(a) \\
&= \mathcal{T}_{\mathsf{rob}}(v_2)_s + \gamma \min_{\mu\in\mathfrak{M}_{\hat{\mu}}(\rho)} \sum_{a,s'} \mu(a,s' \mid s) [v_1(s') - v_2(s')] \beta_s(a).
\end{aligned}$$

It follows that

$$\begin{aligned}
\mathcal{T}_{\mathsf{rob}}(v_2)_s - \mathcal{T}_{\mathsf{rob}}(v_1)_s &\le \gamma \max_{\mu\in\mathfrak{M}_{\hat{\mu}}(\rho)} \sum_{a,s'} \mu(a,s' \mid s) [v_2(s') - v_1(s')] \beta_s(a) \\
&\le \gamma \|v_1 - v_2\|_\infty \cdot \max_{\mu\in\mathfrak{M}_{\hat{\mu}}(\rho)} \sum_{a,s'} \mu(a,s' \mid s) \beta_s(a).
\end{aligned}$$

Applying Proposition 6, we can assert that

$$\max_{\mu\in\mathfrak{M}_{\hat{\mu}}(\rho)} \sum_{a,s'} \mu(a,s' \mid s) \beta_s(a) \le \sum_{a,s'} \hat{\mu}(a,s' \mid s) \beta_s(a) + \rho_s \cdot \|\beta_s\|_{\mathrm{Lip},\hat{\mu}_s}.$$

Applying the Hoeffding upper bound implies that for any $\epsilon_s > 0$,

$$\mathbb{P}\left( \sum_{a,s'} \hat{\mu}(a,s' \mid s) \beta_s(a) \ge 1 + \epsilon_s \right) \le \exp\left( -\frac{2n_s\epsilon_s^2}{M_s^2} \right), \quad \text{where } M_s = \max_a \beta_s(a) - \min_a \beta_s(a). \tag{EC.5}$$

Therefore, with probability at least $1 - \exp\left(-\frac{2n_s\epsilon_s^2}{M_s^2}\right)$,

$$\max_{\mu\in\mathfrak{M}_{\hat{\mu}}(\rho)} \sum_{a,s'} \mu(a,s' \mid s) \beta_s(a) \le 1 + \epsilon_s + \rho_s \|\beta_s\|_{\mathrm{Lip},\hat{\mu}_s} \le \frac{1+\gamma}{2\gamma},$$

which implies that $\mathcal{T}_{\mathsf{rob}}(v_2)_s - \mathcal{T}_{\mathsf{rob}}(v_1)_s \le \frac{1+\gamma}{2} \|v_1 - v_2\|_\infty$. We can exchange the role of $v_1$ and $v_2$ to show that $|\mathcal{T}_{\mathsf{rob}}(v_2)_s - \mathcal{T}_{\mathsf{rob}}(v_1)_s| \le \frac{1+\gamma}{2} \|v_1 - v_2\|_\infty$. Taking the union bound for all $s \in \mathcal{S}$ completes the proof. □

Now we are ready to prove Theorem 1.

*Proof of Theorem 1.* Applying Lemma EC.1, the mapping $\mathcal{T}_{\mathsf{rob}}$ becomes

$$\mathcal{T}_{\mathsf{rob}}[v] = \left[ \sum_a \pi(a \mid s) r(s,a) + \gamma \max_{\kappa \geq 0} \left\{ -\kappa \rho_s + \frac{1}{N_s} \sum_{i=1}^{N_s} \min_{(a,s')} \left\{ v(s')\beta_s(a) + \kappa c((a,s'),(a_i,s_i')) \right\} \right\} \right]_{s \in \mathcal{S}}.$$

Note that $\mathcal{T}_{\mathsf{rob}}[v]$ is an equivalent formulation of the right-hand side of (7). Applying Lemma EC.5, we can see that $\mathcal{T}_{\mathsf{rob}}$ is a component-wise non-decreasing contraction mapping with high probability. Whenever this holds, by Lemma EC.4, at optimality each of the constraint (7) is tight.

Let $v^*$ be the optimal solution to (V) and let $\{\mu^*(\cdot,\cdot \mid s)\}_{s\in\mathcal{S}}$ be the corresponding worst-case conditional distributions yielding from Lemma EC.1. Thus, for all $s \in \mathcal{S}$,

$$\begin{aligned} v^*(s) &= \min_{\mu \in \mathfrak{M}_{\hat{\mu}}(\rho)} \sum_a \pi(a \mid s) r(s,a) + \gamma \sum_{(a,s')} \mu(a,s' \mid s) v^*(s') \beta_s(a) \\ &= \sum_a \pi(a \mid s) r(s,a) + \gamma \sum_{(a,s')} \mu^*(a,s' \mid s) v^*(s') \beta_s(a). \end{aligned} \tag{EC.6}$$

Because of the rectangularity of $\mathfrak{M}_{\hat{\mu}}(\rho)$, the pair $(\mu^*, v^*)$ is feasible for (EC.3). Hence, the optimal value in (V) is lower bounded by the optimal value of (EC.3).

On the other hand, for fixed $\mu \in \mathfrak{M}_{\hat{\mu}}(\rho)$, the optimum $v_\mu$ of the inner maximization problem in (EC.3) satisfies $v_\mu(s) = \sum_a \pi(a \mid s) r(s,a) + \gamma \sum_{(a,s')} \mu(a,s' \mid s) v_\mu(s') \beta_s(a),\ \forall s \in \mathcal{S}$. Since $\mu$ is feasible for $\mathfrak{M}_{\hat{\mu}}(\rho)$, we have

$$v_\mu(s) \geq \min_{\mu \in \mathfrak{M}_{\hat{\mu}}(\rho)} \sum_a \pi(a \mid s) r(s,a) + \gamma \sum_{(a,s')} \mu(a,s' \mid s) v_\mu(s') \beta_s(a).$$

Since $v^*$ is the solution to the fixed point equation (EC.6), applying Theorem 6.2.2 in Puterman (Puterman 1994) gives $v_\mu(s) \geq v^*(s)$ for all $s \in \mathcal{S}$. Because of the non-negativity of $d_0$, we conclude that the optimal value in (EC.3) is lower bounded by that in (V). Therefore, the proof is completed. □

### EC.3.2. Proofs for Propositions

*Proof of Proposition 1.* Since the right-hand side of (7) is a contraction mapping with respect to the value function $v$, all constraints in (7) are tight at optimality. Because of the non-negativity of $d_0(s), \forall s$, solving the optimization problem (V) is equivalent to computing the unique fixed point of the equation (7). As a result, the fixed point iteration presented in Algorithm 1 is guaranteed to converge into the optimal solution because the right-hand side at (7) is a contraction mapping. □

*Proof of Proposition 2.* By assigning the dual multiplier $\{\kappa(s)\}$ for each constraint in (V), the dual problem becomes

$$\min_{\kappa \geq 0} \sum_{(s,a)} \pi(a \mid s) \kappa(s) r(s,a) + \max_v \sum_s v(s) \left[ (1-\gamma) d_0(s) - \kappa(s) \right] + \gamma \sum_s V(s) \kappa(s).$$

In particular, the inner maximization presented above can be reformulated as

$$\begin{aligned} &\max_{v, \lambda_s \geq 0} \sum_s v(s) \left[ (1-\gamma) d_0(s) - \kappa(s) \right] \\ &\qquad + \gamma \sum_s \kappa(s) \left[ -\lambda_s \rho_s + \frac{1}{n_s} \sum_{i=1}^{n_s} \min_{(a,s') \in \mathcal{A} \times \mathcal{S}} \left\{ v(s') \beta_s(a) + \lambda_s c((a,s'),(a_i,s_i')) \right\} \right] \\ &= \max_{\substack{v, \\ \lambda_s \geq 0, \forall s \in \mathcal{S} \\ \alpha_{s,i}, \forall s \in \mathcal{S}, \forall i}} \left\{ \begin{array}{l} \sum_s v(s) \left[ (1-\gamma) d_0(s) - \kappa(s) \right] + \gamma \sum_s \kappa(s) \left[ -\lambda_s \rho_s + \frac{1}{n_s} \sum_{i=1}^{n_s} \alpha_{s,i} \right] \\ \text{Subject to } \alpha_{s,i} \leq v(s') \beta_s(a) + \lambda_s c((a,s'),(a_i,s_i')), \forall i, \forall (a,s') \end{array} \right\}. \end{aligned}$$

By assigning the dual multiplier $\{h^{(s)}_{i,(a,s')}\}_{(a,s'),i}$, the dual of the maximization problem becomes

$$\begin{aligned}
&\min_{h\geq 0} \max_{\substack{v,\\ \lambda_s\geq 0,\forall s\in\mathcal{S}\\ \alpha_{s,i},\forall s\in\mathcal{S},\forall i}} \sum_s v(s)\Big[(1-\gamma)d_0(s)-\kappa(s)\Big]+\gamma\sum_s \kappa(s)\Big[-\lambda_s\rho_s+\frac{1}{n_s}\sum_{i=1}^{n_s}\alpha_{s,i}\Big]\\
&\qquad +\sum_s\sum_i\sum_{(a,s')} h^{(s)}_{i,(a,s')}\Big[v(s')\beta_s(a)+\lambda_s c((a,s'),(a_i,s'_i))-\alpha_{s,i}\Big]\\
&=\min_{h\geq 0} \max_{\substack{v,\\ \lambda_s\geq 0,\forall s\in\mathcal{S}\\ \alpha_{s,i},\forall s\in\mathcal{S},\forall i}} \sum_{s'} v(s')\left[(1-\gamma)d_0(s')-\kappa(s')+\sum_{(s,a)}\sum_i h^{(s)}_{i,(a,s')}\beta_s(a)\right]\\
&\quad+\sum_s\lambda_s\left[-\gamma\kappa(s)\rho_s+\sum_i\sum_{(a,s')} h^{(s)}_{i,(a,s')}c((a,s'),(a_i,s'_i))\right]+\sum_s\sum_i\alpha_{s,i}\left[\frac{1}{n_s}\gamma\kappa(s)-\sum_{(a,s')}h^{(s)}_{i,(a,s')}\right].
\end{aligned}$$

Therefore, the inner maximization problem is bounded if and only if the following conditions hold:

$$\begin{aligned}
(1-\gamma)d_0(s')-\kappa(s')+\sum_{(s,a)}\sum_i h^{(s)}_{i,(a,s')}\beta_s(a)&=0,\ \forall s'\in\mathcal{S},\\
\gamma\kappa(s)\rho_s-\sum_i\sum_{(a,s')}h^{(s)}_{i,(a,s')}c((a,s'),(a_i,s'_i))&\leq 0,\ \forall s\in\mathcal{S},\\
\frac{1}{n_s}\gamma\kappa(s)-\sum_{(a,s')}h^{(s)}_{i,(a,s')}&=0,\ \forall s\in\mathcal{S},\forall i\in[n_s].
\end{aligned}$$

Hence, the dual of problem (V) becomes

$$\begin{aligned}
\min_{\kappa\geq 0,h\geq 0}\quad & \sum_{(s,a)}\pi(a\mid s)\kappa(s)r(s,a)\\
\text{s.t.}\quad & (1-\gamma)d_0(s')+\sum_{(s,a)}\sum_i h^{(s)}_{i,(a,s')}\beta_s(a)=\kappa(s'),\ \forall s'\in\mathcal{S},\\
& \sum_i\sum_{(a,s')}h^{(s)}_{i,(a,s')}c((a,s'),(a_i,s'_i))\leq\gamma\kappa(s)\rho_s,\ \forall s\in\mathcal{S},\\
& \sum_{(a,s')}h^{(s)}_{i,(a,s')}=\frac{\gamma}{n_s}\kappa(s),\forall s\in\mathcal{S},\ \forall i\in[n_s].
\end{aligned}$$

By change of variable $h^{(s)}_{i,(a,s')}\leftarrow h^{(s)}_{i,(a,s')}/\gamma$, this dual problem can be formulated as

$$\begin{aligned}
\min_{\kappa\geq 0,h\geq 0}\quad & \sum_{(s,a)}\pi(a\mid s)\kappa(s)r(s,a)\\
\text{s.t.}\quad & (1-\gamma)d_0(s')+\gamma\sum_{(s,a)}\beta_s(a)\sum_i h^{(s)}_{i,(a,s')}=\kappa(s'),\ \forall s'\in\mathcal{S},\\
& \sum_i\sum_{(a,s')}h^{(s)}_{i,(a,s')}c((a,s'),(a_i,s'_i))\leq\kappa(s)\rho_s,\ \forall s\in\mathcal{S},\\
& \sum_{(a,s')}h^{(s)}_{i,(a,s')}=\frac{1}{n_s}\kappa(s),\forall s\in\mathcal{S},\ \forall i\in[n_s].
\end{aligned}$$

Or we make the change of variable $h^{(s)}_{i,(a,s')}=\kappa(s)q^{(s)}_{i,(a,s')}$ and consider solving

$$\min_{\kappa\geq 0,\ell\geq 0}\quad \sum_{(s,a)}\pi(a\mid s)\kappa(s)r(s,a)$$

$$
\begin{aligned}
\text{s.t.}\quad & (1-\gamma)d_0(s') + \gamma \sum_{(s,a)} \beta_s(a)\kappa(s) \sum_i q^{(s)}_{i,(a,s')} = \kappa(s'),\ \forall s' \in \mathcal{S}, \\
& \sum_i \sum_{(a,s')} q^{(s)}_{i,(a,s')} c((a,s'),(a_i,s'_i)) \le \rho_s,\ \forall s \in \mathcal{S}, \\
& \sum_{(a,s')} q^{(s)}_{i,(a,s')} = \frac{1}{n_s}, \forall s \in \mathcal{S},\ \forall i \in [n_s].
\end{aligned}
$$

□

*Proof of Proposition 3.* Define $\mathcal{Z}_{s,\infty} = \{(a,s') \in \mathcal{A}\times\mathcal{S} :\ \mathfrak{l}_{\beta_s f}(a,s') = \|\beta_s f\|_{\mathrm{Lip},\hat{\mu}_s}\}$, and

$$
\begin{aligned}
\overline{\rho}_s = \sup_{\mathfrak{T}:\ \mathcal{A}\times\mathcal{S}\to\mathcal{A}\times\mathcal{S}} \Bigg\{ & \mathbb{E}_{\hat{\mu}_s}[c(\mathfrak{T}(a,s'),(a,s'))] :\ \mathfrak{T}(a,s') = (a,s'), \forall (a,s') \neq \mathcal{Z}_{s,\infty}, \\
& \frac{\beta_s f(\mathfrak{T}(a,s')) - \beta_s f((a,s'))}{c(\mathfrak{T}(a,s'),(a,s'))} = \mathfrak{l}_{\beta_s f}(a,s'), \forall (a,s') \in \mathcal{Z}_{s,\infty} \Bigg\}.
\end{aligned}
$$

Consider the general case where the function space $\mathcal{F}$ is a subset of $\{f :\ \mathcal{S} \ni s \mapsto f(s) \in \mathbb{R}\}$. Then for fixed $\kappa$, the inner problem of robust OPE becomes

$$
\begin{aligned}
& \min_{\mu \in \mathfrak{M}_{\hat{\mu}}(\rho), \forall s}\ \max_{f \in \mathcal{L}[\mathcal{F}]}\ (1-\gamma)\mathbb{E}_{s\sim d_0}[f(s)] \\
& \qquad + \left\{ \sum_s \kappa(s) r_\pi(s) + \gamma \mathbb{E}_{s\sim\kappa(s)} \left[ \mathbb{E}_{(a,s')\sim\mu(\cdot,\cdot|s)}[\beta_s(a) f(s')] \right] - \mathbb{E}_{s\sim\kappa}[f(s)] \right\} \\
= & \max_{f\in\mathcal{L}[\mathcal{F}]}\ (1-\gamma)\mathbb{E}_{s\sim d_0}[f(s)] + \sum_s \kappa(s) r_\pi(s) - \mathbb{E}_{s\sim\kappa}[f(s)] \\
& \qquad + \min_{\mu \in \mathfrak{M}_{\hat{\mu}}(\rho), \forall s} \left\{ \gamma \mathbb{E}_{s\sim\kappa(s)} \left[ \mathbb{E}_{(a,s')\sim\mu(\cdot,\cdot|s)}[\beta_s(a) f(s')] \right] \right\} \\
= & \max_{f\in\mathcal{L}[\mathcal{F}]}\ (1-\gamma)\mathbb{E}_{s\sim d_0}[f(s)] + \sum_s \kappa(s) r_\pi(s) - \mathbb{E}_{s\sim\kappa}[f(s)] + \gamma \mathbb{E}_{s\sim\kappa(s)} \left\{ \min_{\mu\in\mathfrak{M}_{\hat{\mu}}(\rho)} \mathbb{E}_{(a,s')\sim\mu(\cdot,\cdot|s)}[\beta_s(a) f(s')] \right\},
\end{aligned}
$$

where the first equality is by applying the sion's minimax theorem (Sion 1958) based on the fact that $\mathcal{L}[\mathcal{F}]$ is a linear topological space and $\mathfrak{M}_{\hat{\mu}}(\rho)$ is a compact Wasserstein ball, and the second equality is because of the rectangular structure of the ambiguity set $\mathcal{P}$. Based on the assumption that $\rho_s < \overline{\rho}_s, \forall s$ and applying Theorem 6 on the inner minimization within the Wasserstein ball, this problem can be equivalently formulated as

$$
\begin{aligned}
& \max_{f\in\mathcal{F}}\ (1-\gamma)\mathbb{E}_{s\sim d_0}[f(s)] + \sum_s \kappa(s) r_\pi(s) \\
& \qquad -\mathbb{E}_{s\sim\kappa}[f(s)] + \gamma \mathbb{E}_{s\sim\kappa(s)} \left\{ \mathbb{E}_{(a,s')\sim\hat{\mu}_s}[\beta_s(a) f(s')] - \rho_s \cdot \|\beta_s(\cdot) f(\cdot)\|_{\mathrm{Lip},\hat{\mu}_s} \right\}.
\end{aligned}
$$

Combining this formulation with the outer minimization with respect to $\kappa$ completes the proof. For general radii $\{\rho_s\}$, the results can be derived using Proposition 6 in a similar manner. □

## Appendix EC.4: Proofs for Section 5

### EC.4.1. Proofs for Section 5.1

We begin with a lemma on the error bounds for the perturbed value iterations.

LEMMA EC.6. *Denote by $\mathcal{T}^{\mathsf{true}}$ the Bellman operator with the true conditional probability $d_{\pi_b}(a,s' \mid s)$:*

$$
\mathcal{T}^{\mathsf{true}}[v](s) = \textstyle\sum_a \pi(a \mid s) r(s,a) + \gamma \sum_{(a,s')} d_{\pi_b}(a,s' \mid s) v(s') \beta_s(a).
$$

*Denote by $\mathcal{T}^l$ and $\mathcal{T}^u$ perturbations of $\mathcal{T}$ so that*

$$\mathcal{T}^l[v](s) = T^{\mathsf{true}}[v](s) - \epsilon_v^l(s),$$
$$\mathcal{T}^u[v](s) = T^{\mathsf{true}}[v](s) + \epsilon_v^u(s).$$

*Assume that there exist $\epsilon^l = (\epsilon^l(s))_{s\in\mathcal{S}}$ and $\epsilon^u = (\epsilon^l(s))_{s\in\mathcal{S}}$ such that $\epsilon_v^l(s) \le \epsilon^l(s)$ and $\epsilon_v^u(s) \le \epsilon^u(s)$ for all $s\in\mathcal{S}$ and $v$. Let $v^{\mathsf{true}}, v^l, v^u$ be the solutions to the fixed point of $\mathcal{T}^{\mathsf{true}}, \mathcal{T}^l, \mathcal{T}^u$, respectively. Then*

$$v^{\mathsf{true}} - v^l \le \left(I - \gamma P^{\mathsf{true}}\right)^{-1}\epsilon^l,$$
$$v^{\mathsf{true}} - v^u \ge -\left(I - \gamma P^{\mathsf{true}}\right)^{-1}\epsilon^u,$$

*where the matrix $P^{\mathsf{true}} \in \mathbb{R}^{|\mathcal{S}|\times|\mathcal{S}|}$ is defined as $P^{\mathsf{true}}_{s,s'} := \sum_a d_{\pi_b}(a,s' \mid s)\beta_s(a)$, and the inequality is interpreted component-wise.*

*Proof of Lemma EC.6.* Define $v^l_{(k)}$ as the $k$-th iteration point for the value iteration algorithm with Bellman operator $\mathcal{T}^l$. Then we have the relation

$$\begin{aligned}
v_l^{(k+1)} - v^{\mathsf{true}} &= \mathcal{T}^l[v^l_{(k)}] - \mathcal{T}^{\mathsf{true}}[v^{\mathsf{true}}] \\
&= \mathcal{T}^{\mathsf{true}}[v^l_{(k)}] - \mathcal{T}^{\mathsf{true}}[v^{\mathsf{true}}] - \epsilon_{v^l_{(k)}} \\
&\ge \mathcal{T}^{\mathsf{true}}[v^l_{(k)}] - \mathcal{T}^{\mathsf{true}}[v^{\mathsf{true}}] - \epsilon^l \\
&= \gamma\left(\sum_{a,s'} d_{\pi_b}(a,s' \mid s)\beta_s(a)(v^l_{(k)} - v^{\mathsf{true}})\right)_{s\in\mathcal{S}} - \epsilon^l \\
&= \gamma P^{\mathsf{true}}(v^l_{(k)} - v^{\mathsf{true}}) - \epsilon^l.
\end{aligned}$$

Applying the relation inductively, we have

$$v^l_{(n)} - v^{\mathsf{true}} \ge \gamma^n (P^{\mathsf{true}})^n (v^l_{(0)} - v^{\mathsf{true}}) - \sum_{k=0}^{n-1} \gamma^{n-k-1}(P^{\mathsf{true}})^{n-k-1}\epsilon^l.$$

Taking the limit $n\to\infty$ and applying the identity $(I-A)^{-1} = \lim_{n\to\infty}\sum_{k=0}^{n-1} A^k$ gives the desired result. The other part of this Lemma follows the similar argument. □

Lemma EC.7. *For fixed $s\in\mathcal{S}$, we define*

$$\mathcal{F}_s := \{(a,s') \mapsto \beta_s(a)v(s') :\ a\in\mathcal{A}, s'\in\mathcal{S}\}.$$

*Define*

$$\mathsf{d}(\beta_s(\cdot)v(\cdot), \beta_s(\cdot)\tilde{v}(\cdot)) = |(1-\gamma)d_0^\top\left(I-\gamma P^{\mathsf{true}}\right)^{-1}(\tilde{v} - v)|.$$

*Then*

$$\mathcal{N}\left(\frac{1}{n}, \mathcal{F}_s, \mathsf{d}\right) \le 2nM.$$

*Proof of Lemma EC.7.* Denote $\mathcal{H}$ as the collection of all possible value functions $v$. Take the semi-metric of $\mathcal{H}$ as

$$\mathsf{d}(\tilde{v}, v) := |(1-\gamma)d_0^\top\left(I-\gamma P^{\mathsf{true}}\right)^{-1}(\tilde{v}-v)|.$$

For fixed $t$, consider the line set $\mathcal{H}_t = \{v :\ (1-\gamma)d_0^\top\left(I-\gamma P^{\mathsf{true}}\right)^{-1}v = t\}$. For any $\epsilon\ge 0$, we have $\mathsf{d}(v,v')\le\epsilon$ when $v\in\mathcal{H}_t$ and $v'\in\mathcal{H}_{t+\epsilon}$. In order to find the $\frac{1}{n}$-covering number of $\mathcal{H}$, we only need to find the covering number for the 1-dimensional parameter $t$ with $-M\le t\le M$, which is $2nM$. Since the function class $\mathcal{F}_s$ can be expressed as

$$\mathcal{F}_s = \{a\mapsto\beta_s(a)\}\times\mathcal{H},$$

the covering number of $\mathcal{F}_s$ can be upper bounded as $2nM$. □

Applying Proposition 8 with $\eta = 1$ to the right-hand side of constraint (7) gives the following result.

LEMMA EC.8. *Fix $s \in \mathcal{S}$ and define $\mathcal{F}_s = \{(a, s') \mapsto v(s')\beta_s(a), a \in \mathcal{A}, s' \in \mathcal{S}\}$. Let $\tau_s > 0$, and set*

$$\rho_{n,s} = \sqrt{\frac{2\tau_s}{n_s}} \mathrm{diam}(\mathcal{A} \times \mathcal{S}), \quad \delta_s = \min_{(a,s') \in \mathrm{supp}\, d_{\pi_b}(\cdot,\cdot|s)} d_{\pi_b}(a, s' \mid s), \quad \Delta_s = \frac{1}{\left(\frac{1-2\delta_s}{1-\delta_s} \vee \frac{2\delta_s - 1}{\delta_s}\right)}.$$

*With probability at least $1 - \alpha'_s$, where*

$$\begin{aligned} \alpha'_s := \exp\big(-\tau_s + \log(2n_s M)\big) + \exp\left(-n_s \log\left(1/(1-\delta_s)\right) + \log(2n_s M)\right) \\ + \exp\Big(-n_s \log \Delta_s + \log(2n_s M)\Big), \end{aligned} \tag{EC.7}$$

*simultaneously for every function $v$ with $|v(s)| \le M$, it holds that*

$$\mathbb{E}_{d_{\pi_b}}[v(s)\beta_s(a)] \ge \mathcal{T}_{\mathrm{rob}}[v] - \frac{6}{n_s}.$$

In particular, based on the identity $\exp(-c\log(1/(1-\delta))) \le \exp(-c\delta)$ for any $c > 0, \delta \in [0, 1)$, we can replace the error probability $\alpha'_s$ in (EC.7) with

$$\alpha_s := \exp\big(-\tau_s + \log(2n_s M)\big) + \exp\left(-n_s\delta_s + \log(2n_s M)\right) + \exp\Big(-n_s \log \Delta_s + \log(2n_s M)\Big).$$

This new error probability is slightly more conservative, but it highlights the linear dependence of $\delta_s$ on the sample size $n_s$.

*Proof of Theorem 2.* Taking the union bound of the probability presented in Lemma EC.8, with probability at least $1 - \sum_{s \in \mathcal{S}} \alpha_s$, the Bellman operator $\mathcal{T}_{\mathrm{rob}}$ for robust reward defined in (EC.4) admits the following error bound for any value function $v$ satisfying $|v(s)| \le M, \forall s \in \mathcal{S}$,

$$\mathcal{T}_{\mathrm{rob}}[v] = \mathcal{T}^*[v] + \epsilon_v, \quad \epsilon_v(s) \le \epsilon_{n,s}.$$

Recall that we have used $v_*$ to denote the fixed point of $\mathcal{T}_{\mathrm{rob}}$. When this upper bound holds, applying Lemma EC.6 implies

$$v^{\mathrm{true}} - v^* \ge -(I - \gamma P^{\mathrm{true}})^{-1}\epsilon_n.$$

Then the gap between the underlying reward and the robust reward is

$$R_\pi \ge \mathcal{L}_{\hat{\mu}_n}(\rho_n) - d_0^T (I - \gamma P^{\mathrm{true}})^{-1}\epsilon_n.$$

Similarly, with probability at least $1 - \sum_{s \in \mathcal{S}} \alpha_s$, it holds that

$$R_\pi \le \mathcal{U}_{\hat{\mu}_n}(\rho_n) + d_0^T (I - \gamma P^{\mathrm{true}})^{-1}\epsilon_n.$$

Taking the union bound implies the desired result. □

*Proof of Proposition 4.* Denote by $\hat{\mathcal{T}}$ the empirical Bellman operator such that

$$\hat{\mathcal{T}}[v](s) = \sum_s \pi(a \mid s) r(s, a) + \gamma \sum_{s' \in \mathcal{S}} P_{\hat{\mu}_n}(s, s') v(s') \;\forall s \in \mathcal{S}.$$

where $P_{\hat{\mu}_n} \in \mathbb{R}^{|\mathcal{S}| \times |\mathcal{S}|}$ denotes the transition kernel with $P_{\hat{\mu}_n}(s, s') = \sum_a \hat{\mu}_n(a, s' \mid s)\beta_s(a)$. Denote by $\{\hat{v}(s)\}_{s \in \mathcal{S}}$ the empirical value function, which is the fixed point for the empirical Bellman operator. Leveraging the concentration inequality in (EC.5) implies that with probability at least $1 - \exp\left(-\frac{2\tau_s'^2}{M_s^2}\right)$, it holds that

$$\sum_{s'} P_{\hat{\mu}_n}(s, s') = \sum_{a \in \mathcal{A}, s' \in \mathcal{S}} \hat{\mu}_n(a, s' \mid s)\beta_s(a) \le 1 + \frac{\tau_s'}{\sqrt{n_s}}$$

Therefore, with probability at least $1-\sum_{s\in\mathcal{S}}\exp\left(-\frac{2\tau_s'^2}{M_s^2}\right)$, for all $s\in\mathcal{S}$ and for any two value functions $v_1, v_2$, it holds that

$$
\begin{aligned}
\left|\hat{\mathcal{T}}[v_1](s)-\hat{\mathcal{T}}[v_2](s)\right| &= \gamma\left|\sum_{(a,s')}\hat{\mu}_n(a,s'\mid s)[v_1(s')-v_2(s')]\beta_s(a)\right| \\
&\leq \gamma\|v_1-v_2\|_\infty \sum_{(a,s')}\hat{\mu}_n(a,s'\mid s)\beta_s(a) \\
&\leq \left(\gamma+\frac{\gamma\tau_s'}{\sqrt{n_s}}\right)\|v_1-v_2\|_\infty \leq \frac{1+\gamma}{2}\|v_1-v_2\|_\infty,
\end{aligned}
$$

which means that $\hat{\mathcal{T}}$ is a contraction mapping, i.e., the matrix $I-\gamma P_{\hat{\mu}_n}$ is invertible. Denote by $T^l$ and $T^u$ the robust and optimistic Bellman operators, with the associated robust and optimistic value functions being $v^l$ and $v^u$, respectively. Note that the Bellman operator $T^l$ satisfies

$$
\hat{\mathcal{T}}(s)-T^l(s)\leq \epsilon^l(s) := \gamma\rho_{n,s}\cdot\|\beta_s v^l\|_{\mathrm{Lip},\hat{\mu}_{n,s}}.
$$

Applying the similar perturbation analysis as in Lemma EC.6 gives

$$
\hat{v}-v^l\leq (I-\gamma P_{\hat{\mu}})^{-1}\epsilon^l.
$$

Similarly, $v^u-\hat{v}\leq (I-\gamma P_{\hat{\mu}})^{-1}\epsilon^u$ with $\epsilon^u(s) := \gamma\rho_{n,s}\cdot\|\beta_s v^u\|_{\mathrm{Lip},\hat{\mu}_{n,s}}$.

In summary, with probability at least $1-\sum_{s\in\mathcal{S}}\exp\left(-\frac{2\tau_s'^2}{M_s^2}\right)$, the length of the confidence interval can be upper bounded by

$$
\begin{aligned}
\mathcal{U}_{\hat{\mu}_n}(\rho)-\mathcal{L}_{\hat{\mu}_n}(\rho) &\leq \left|\sum_s d_0(s)[v^u(s)-v^l(s)]\right| \\
&\leq \left|\sum_s d_0(s)[\hat{v}(s)-v^l(s)]\right| + \left|\sum_s d_0(s)[v^u(s)-\hat{v}(s)]\right| \\
&\leq d_0^T(I-\gamma P_{\hat{\mu}_n})^{-1}\epsilon^l + d_0^T(I-\gamma P_{\hat{\mu}_n})^{-1}\epsilon^u.
\end{aligned}
$$

According to the definition of function variation term, it holds that

$$
\|\beta_s v^l\|_{\mathrm{Lip},\hat{\mu}_{n,s}} \leq \|\beta_s v^l\|_{\mathrm{Lip},d_{\pi_b}(\cdot,\cdot\mid s)} \leq \max_{v\in\mathbb{R}^{|\mathcal{S}|},|v(s)|\leq M}\|\beta_s v\|_{\mathrm{Lip},d_{\pi_b}(\cdot,\cdot\mid s)}.
$$

The other function variation term can also be upper bounded similarly. □

### EC.4.2. Proofs for Section 5.2

*Proof of Theorem 3.* Define the adversarial Bellman operator $\mathcal{T}:\ \mathbb{R}^{|\mathcal{S}|}\to\mathbb{R}^{|\mathcal{S}|}$ as

$$
\mathcal{T}[v] = \left(\sum_a \pi(a\mid s)r(s,a)+\gamma\min_{\mu\in\mathfrak{M}_{d_{\pi_b}}(\rho)}\sum_{(a,s')}\mu(a,s'\mid s)v(s')\beta_s(a)\right)_{s\in\mathcal{S}},\quad v\in\mathbb{R}^{|\mathcal{S}|}.
$$

Define the empirical adversarial Bellman operator $\hat{\mathcal{T}}:\ \mathbb{R}^{|\mathcal{S}|}\to\mathbb{R}^{|\mathcal{S}|}$ as

$$
\hat{\mathcal{T}}[v] = \left(\sum_a \pi(a\mid s)r(s,a)+\gamma\min_{\mu\in\mathfrak{M}_{\hat{\mu}}(\rho)}\sum_{(a,s')}\mu(a,s'\mid s)v(s')\beta_s(a)\right)_{s\in\mathcal{S}},\quad v\in\mathbb{R}^{|\mathcal{S}|}.
$$

We realize the Lipschitz constant of the operator $\mathcal{T}$ can be bounded as

$$
\gamma\max_{\mu\in\mathfrak{M}_{d_{\pi_b}}(\rho)}\sum_{(a,s')}\mu(a,s'\mid s)\beta_s(a)\leq\gamma(1+\rho_s\|\beta_s\|_{\mathrm{Lip},d_\pi(\cdot,\cdot\mid s)})\leq\gamma' := \frac{1+\gamma}{2}.
$$

Therefore, the Bellman operator $\mathcal{T}$ is a contractive mapping. Define the following two thresholds

$$\overline{\rho}_s^{(1)} = \sup_{T:\ \mathcal{A}\times\mathcal{S}\to\mathcal{A}\times\mathcal{S}} \Big\{ \mathbb{E}_{d_{\pi_b}(\cdot,\cdot|s)}[c(\mathfrak{T}(a,s'),(a,s'))]:\ \mathfrak{T}(a,s') = (a,s'), \forall (a,s') \neq \mathcal{Z}_{s,\infty}, \\ \frac{\beta_s f(\mathfrak{T}(a,s')) - \beta_s f((a,s'))}{c(\mathfrak{T}(a,s'),(a,s'))} = \mathfrak{l}_{\beta_s f}(a,s'), \forall (a,s') \in \mathcal{Z}_{s,\infty} \Big\},$$

$$\overline{\rho}_s^{(2)} = \sup_{T:\ \mathcal{A}\times\mathcal{S}\to\mathcal{A}\times\mathcal{S}} \Big\{ \mathbb{E}_{\hat{\mu}_s}[c(\mathfrak{T}(a,s'),(a,s'))]:\ \mathfrak{T}(a,s') = (a,s'), \forall (a,s') \neq \mathcal{Z}_{s,\infty}, \\ \frac{\beta_s f(\mathfrak{T}(a,s')) - \beta_s f((a,s'))}{c(\mathfrak{T}(a,s'),(a,s'))} = \mathfrak{l}_{\beta_s f}(a,s'), \forall (a,s') \in \mathcal{Z}_{s,\infty} \Big\}.$$

Suppose $\rho_s < \overline{\rho}_s := \overline{\rho}_s^{(1)} \wedge \overline{\rho}_s^{(2)}$. By Proposition 9 with $\eta = 1$, the following relation holds with probability at least $1 - \exp(-n_s\delta_s + \log(2n_sM)) - 2e^{-\tau}$:

$$|\mathcal{T}[v](s) - \hat{\mathcal{T}}[v](s)| \le \iota'_{n_s} := 2\mathbb{E}_{\otimes}[\mathfrak{R}_{n_s}(\mathcal{F}_s)] + H_s\sqrt{\frac{\tau}{2n_s}}.$$

Define $\hat{v}_{(k)}$ as the $k$-th iteration point for the value iteration algorithm with Bellman operator $\hat{\mathcal{T}}$. Let $v$ be the fixed point solution of the Bellman operator $\mathcal{T}$. Then we have the relation

$$\begin{aligned} |\hat{v}_{(k+1)}(s) - v(s)| &= \left|\hat{\mathcal{T}}[\hat{v}_{(k)}](s) - \mathcal{T}[v](s)\right| \\ &\le \left|\mathcal{T}[\hat{v}_{(k)}](s) - \mathcal{T}[v](s)\right| + \left|\mathcal{T}[\hat{v}_{(k)}](s) - \hat{\mathcal{T}}[\hat{v}_{(k+1)}](s)\right| \\ &\le \gamma'|\hat{v}_{(k)}(s) - v(s)| + \iota'_{n_s}. \end{aligned}$$

Applying this relation inductively, it holds that

$$|\hat{v}(s) - v(s)| := \lim_{k\to\infty} |\hat{v}_{(k+1)}(s) - v(s)| \le \iota'_{n_s}\sum_{k=0}^{\infty}(\gamma')^k = \frac{\iota'_{n_s}}{1-\gamma'},$$

where $\hat{v}$ is the fixed point solution of the Bellman operator $\hat{\mathcal{T}}$. Taking the union bound implies that the relation

$$|\mathcal{L}^{\mathsf{adv}}(\rho) - \mathcal{L}_{\hat{\mu}_n}(\rho)| = \left|(1-\gamma)\sum_s [v(s) - \hat{v}(s)]d_0(s)\right| \le \frac{1-\gamma}{1-\gamma'}\sum_s d_0(s)\iota'_{n_s}$$

holds with probability at least $1 - \sum_s \alpha_s$. The proof is completed. □

To prove Theorem 4, We first establish the asymptotic convergence for the transition probabilities $\hat{\mu}_T$ into $d_{\pi_b}$, and then the convergence for $\mathcal{L}^{\mathsf{adv}}(\rho)$ can be built by applying the functional delta theorem.

Lemma EC.9. *Denote by* $\hat{\mu}_T = \mathrm{vec}(\{\hat{\mu}_T(a,s' \mid s)\})$, *and* $d_{\pi_b} = \mathrm{vec}(\{d_{\pi_b}(a,s' \mid s)\})$. *It holds that*

$$\sqrt{T}\big(\hat{\mu}_T - d_{\pi_b}\big) \xrightarrow{\mathbf{d}} \mathcal{N}(0, D\Lambda D),$$

*where* $\Lambda \in \mathbb{R}_+^{|\mathcal{S}||\mathcal{A}||\mathcal{S}|\times|\mathcal{S}||\mathcal{A}||\mathcal{S}|}$ *is defined as*

$$\Lambda_{(s,(a,s')),(\tilde{s},(\tilde{a},\tilde{s}'))} = \begin{cases} d_{\pi_b}(a,s' \mid s)(1 - d_{\pi_b}(a,s' \mid s)), & \text{if } (s,(a,s')) = (\tilde{s},(\tilde{a},\tilde{s}')), \\ -d_{\pi_b}(a,s' \mid s)d_{\pi_b}(\tilde{a},\tilde{s}' \mid s) & \text{if } s = \tilde{s}, (a,s') \neq (\tilde{a},\tilde{s}'), \\ 0, & \text{otherwise}, \end{cases}$$

*and* $D = \mathrm{diag}\left(\mathrm{vec}\left((d_{\pi_b})^{-\frac{1}{2}} \otimes 1_{|\mathcal{A}||\mathcal{S}|}\right)\right)$, $d_{\pi_b}^{-1/2} := \left\{\frac{1}{\sqrt{d_{\pi_b}(s)}}\right\}_{s\in\mathcal{S}}$, *and* $\mathrm{vec}(\cdot)$ *denotes the vectorization of a matrix by stacking the rows of the matrix on top of one another.*

*Proof of Lemma EC.9.* The proof essentially follows Billingsley (1961) on the maximum likelihood estimate for Markov chains. Denote by $X_s^{(m)}$ be the action-state pair directly after the $m$-th return to $s$. Define

$$
\begin{aligned}
p_{s,(a,s')} &:= \pi_b(a|s)P(s'|s,a),\\
\tau_{s,(a,s')}(T) &:= \sum_{t=1}^{T} 1\{s_t = s, a_t = a, s_{t+1} = s'\},\\
\tau_s(T) &:= \sum_{t=1}^{T} 1\{s_t = s\},\\
Q_{s,(a,s')}(T) &:= \sum_{m=1}^{\lfloor T d_{\pi_b}(s)\rfloor} 1\{X_s^{(m)} = (a,s')\}.
\end{aligned}
$$

It follows that

$$\hat{\mu}_T(a,s'|s) = \tau_{s,(a,s')}(T)/\tau_s(T).$$

From the Markov property, for each $s \in \mathcal{S}$, $\{Q_{s,(a,s')}\}_{a\in\mathcal{A},s'\in\mathcal{S}}$ is multinomially distributed with $\lfloor T d_{\pi_b}(s)\rfloor$ trials and success probability vector $p_{s,(\cdot,\cdot)}$. As a result, we have

$$\left[\frac{Q_{s,(a,s')}(T) - \lfloor T d_{\pi_b}(s)\rfloor p_{s,(a,s')}}{\sqrt{\lfloor T d_{\pi_b}(s)\rfloor}}\right]_{s\in\mathcal{S},(a,s')\in\mathcal{A}\times\mathcal{S}} \xrightarrow{\mathbf{d}} \mathcal{N}(0,\Lambda).$$

Next, we prove that

$$\left[\frac{\tau_{s,(a,s')}(T) - \tau_s(T) p_{s,(a,s')}}{\sqrt{\lfloor T d_{\pi_b}(s)\rfloor}}\right]_{s\in\mathcal{S},(a,s')\in\mathcal{A}\times\mathcal{S}} \xrightarrow{\mathbf{d}} \mathcal{N}(0,\Lambda).$$

To this end, it suffices to show for each $s \in \mathcal{S}$ and $(a,s') \in \mathcal{A}\times\mathcal{S}$,

$$\Delta_T := \frac{\tau_{s,(a,s')}(T) - \tau_s(T)p_{s,(a,s')} - Q_{s,(a,s')}(T) + \lfloor T d_{\pi_b}(s)\rfloor p_{s,(a,s')}}{\sqrt{T}} \xrightarrow{\mathbf{p}} 0.$$

Fixing $s \in \mathcal{S}$ and $(a,s') \in \mathcal{A}\times\mathcal{S}$, setting $Z_m := 1\{X_s^{(m)} = (a,s')\} - p_{s,(a,s')}$ and $Y_m := \sum_{j=1}^m Z_j$, we rewrite $\Delta_T$ as

$$\Delta_T = \frac{Y_{\tau_s(T)} - Y_{\lfloor T d_{\pi_b}(s)\rfloor}}{\sqrt{T}}.$$

Let $\epsilon > 0$. By consistency of the empirical frequency $\tau_s(T)$, there exists $T_0$ such that for all $T > T_0$,

$$\mathbb{P}\{|\tau_{s,(a,s')}(T) - T d_{\pi_b}(s)| > T\epsilon^3\} \le \epsilon.$$

Note that $Z_j$'s are i.i.d. with mean 0 and variance $p_{s,(a,s')}(1-p_{s,(a,s')})$. For $T > T_0$, we have from Chebyshev's inequality that

$$
\begin{aligned}
\mathbb{P}\{|\Delta_T| > \epsilon\} &\le \mathbb{P}\{|\tau_{s,(a,s')}(T) - T d_{\pi_b}(s)| > T\epsilon^3\} + \mathbb{P}\{|\Delta_T| > \epsilon, |\tau_s(T) - T d_{\pi_b}(s)| \le T\epsilon^3\}\\
&\le \epsilon + \mathbb{P}\left\{\max_{m\in\mathbb{N}:|m - T d_{\pi_b}(s)|\le T\epsilon^3} |Y_m - Y_{\lfloor T d_{\pi_b}(s)\rfloor}| > \epsilon\sqrt{T}\right\}\\
&\le \epsilon + 2\sum_{m\in\mathbb{N}:|m - T d_{\pi_b}(s)|\le T\epsilon^3} \mathbb{P}\{|Y_m| > \epsilon\sqrt{T}/2\}\\
&\le \epsilon + 8T\epsilon^3 p_{s,(a,s')}(1-p_{s,(a,s')})/(T\epsilon^2)\\
&= \epsilon(1 + 8p_{s,(a,s')}(1-p_{s,(a,s')})) \to 0.
\end{aligned}
$$

Finally, observe that

$$
\begin{aligned}
&\left[\sqrt{T}\left(\hat{\mu}_T(a,s'\mid s)-d_{\pi_b}(a,s'\mid s)\right)\right]_{s\in\mathcal{S},(a,s')\in\mathcal{A}\times\mathcal{S}}\\
=&\left[\frac{1}{\sqrt{d_{\pi_b}(s)}}\left(\sqrt{\tau_s(T)}\hat{\mu}_T(a,s'\mid s)-\sqrt{\tau_s(T)}d_{\pi_b}(a,s'\mid s)\right)\right]_{s\in\mathcal{S},(a,s')\in\mathcal{A}\times\mathcal{S}}\\
=&\left[\frac{1}{\sqrt{d_{\pi_b}(s)}}\left(\frac{\tau_{s,(a,s')}(T)}{\sqrt{\tau_s(T)}}-\frac{\tau_s(T)d_{\pi_b}(a,s'\mid s)}{\sqrt{\tau_s(T)}}\right)\right]_{s\in\mathcal{S},(a,s')\in\mathcal{A}\times\mathcal{S}}\\
=&\left[\frac{1}{\sqrt{d_{\pi_b}(s)}}\frac{\sqrt{\lfloor Td_{\pi_b}(s)\rfloor}}{\sqrt{\tau_s(T)}}\cdot\left[\frac{\tau_{s,(a,s')}(T)-\tau_s(T)p_{s,(a,s')}}{\sqrt{\lfloor Td_{\pi_b}(s)\rfloor}}\right]\right]_{s\in\mathcal{S},(a,s')\in\mathcal{A}\times\mathcal{S}}\\
\xrightarrow{\mathbf{P}}&\,\mathcal{N}(0,D\Lambda D),
\end{aligned}
$$

where the convergence follows using Slutsky's theorem and linear transformation of multivariate normal distribution. □

LEMMA EC.10. *Under the setting of Theorem 4, the matrix $I-\gamma P_\mu$ is invertible for any $\mu\in\mathfrak{M}_{d_{\pi_b}}(\rho)$.*

*Proof of Lemma EC.10* It suffices to show that the mapping $\mathcal{T}:\ v\mapsto\gamma P_\mu v$ is contractive. For any $v_1,v_2\in\mathbb{R}^{|\mathcal{S}|}$, we have

$$
\begin{aligned}
\|\mathcal{T}[v_1]-\mathcal{T}[v_2]\|_\infty&=\gamma\max_s\left|\sum_{(a,s')}\mu(a,s'\mid s)[v_1(s')-v_2(s')]\beta_s(a)\right|\\
&\le\gamma\|v_1-v_2\|_\infty\max_s\sum_{(a,s')}\mu(a,s'\mid s)\beta_s(a)\\
&\le\gamma\|v_1-v_2\|_\infty\max_s\max_{\mu\in\mathfrak{M}_{d_{\pi_b}}(\rho)}\sum_{(a,s')}\mu(a,s'\mid s)\beta_s(a).
\end{aligned}
$$

In particular, applying Proposition 6 implies that

$$
\max_{\mu\in\mathfrak{M}_{d_{\pi_b}}(\rho)}\sum_{(a,s')}\mu(a,s'\mid s)\beta_s(a)\le\sum_{a,s'}d_{\pi_b}(a,s'\mid s)\beta_s(a)+\rho_s\cdot\|\beta_s\|_{\mathrm{Lip},d_{\pi_b}(\cdot,\cdot\mid s)}.
$$

Therefore, as long as $1+\rho_s\cdot\|\beta_s\|_{\mathrm{Lip},d_{\pi_b}(\cdot,\cdot\mid s)}<1/\gamma$ for all $s\in\mathcal{S}$, the mapping $\mathcal{T}$ is contractive. □

*Proof of Theorem 4.* Define

$$
\begin{aligned}
\kappa_\mu(s)&=w(s)\mu(s),\\
P_\mu(s,s')&=\sum_a\beta_s(a)\mu(a,s'\mid s),\\
r_\pi&=\sum_a\pi_b(a\mid s)\beta_s(a)r(s,a),
\end{aligned}
$$

then the stationary constraint (P-b) can be reformulated as

$$
\kappa_\mu(s')=(1-\gamma)d_0(s')+\gamma\sum_s\kappa_\mu(s)P_\mu(s,s').
$$

To apply the delta method, let us compute $\nabla_\mu\mathcal{L}_{d_{\pi_b}}(\rho)$. From Lemma EC.10 we can see that the system of equations above has the unique solution, and we can reformulate $\mathcal{L}_{d_{\pi_b}}$ as the following:

$$
\mathcal{L}_{d_{\pi_b}}(\rho)=\min_{\mu\in\mathfrak{M}_{d_{\pi_b}}(\rho)}\sum_s\kappa_\mu(s)r_\pi(s)=\min_{\mu\in\mathfrak{M}_{d_{\pi_b}}(\rho)}\langle(1-\gamma)r_\pi,\left(I-\gamma P_\mu^T\right)^{-1}d_0\rangle.
$$

Denote the minimizer above as $\mu^*$. By the envelope theorem, the gradient of $\mathcal{L}_\mu(\rho)$ (with respect to $\mu$) at $\mu = d_{\pi_b}$ can be expressed as

$$\begin{aligned}
\nabla \mathcal{L}_{d_{\pi_b}}(\rho) &= \nabla \langle (1-\gamma) r_\pi, (I - \gamma P_{\mu^*}^T)^{-1} d_0 \rangle \\
&= \langle (1-\gamma) r_\pi, \nabla (I - \gamma P_{\mu^*}^T)^{-1} d_0 \rangle \\
&= \langle (1-\gamma) r_\pi, \gamma (I - \gamma P_{\mu^*}^T)^{-1} (\nabla P_{\mu^*}^T)(I - \gamma P_{\mu^*}^T)^{-1} d_0 \rangle \\
&= \gamma (1-\gamma) \langle \mathrm{vec}\left( (I - \gamma P_{\mu^*}^T)^{-1} d_0 r_\pi^T (I - \gamma P_{\mu^*}^T)^{-1} \right), \nabla \mathrm{vec}(P_{\mu^*}^T) \rangle \\
&= \gamma (1-\gamma) \left\{ \left( (I - \gamma P_{\mu^*}^T)^{-1} d_0 r_\pi^T (I - \gamma P_{\mu^*}^T)^{-1} \right)_{s,s'} \beta_s(a) : \ (s,a,s') \in \mathcal{S} \times \mathcal{A} \times \mathcal{S} \right\},
\end{aligned}$$

where the last inequality is because

$$\nabla \mathrm{vec}(P_{\mu^*}^T) = \left\{ 1(\bar{s} = s, \bar{s}' = s') \beta_s(\bar{a}) \right\}_{(s,s'),(\bar{s},\bar{a},\bar{s}')}.$$

Therefore, using the delta theorem and Lemma EC.9 gives the result. □

## Appendix EC.5: Proofs for Section 6

LEMMA EC.11. *Under the setting of Theorem 6, with probability at least* $1 - \exp\left(-2 n_s \delta_s'^2 \epsilon_s^2\right)$*, it holds that*

$$\max_a \frac{\hat{\mu}_n(a \mid s)}{\pi_b(a \mid s)} \le \frac{1+\gamma}{2\gamma} - \rho_{n,s} \max_{\pi(\cdot|s)} \|\beta_s^\pi\|_{\mathrm{Lip}, \hat{\mu}_{n,s}}.$$

*Proof.* First, we can see that the maximization over the variation term is bounded:

$$\begin{aligned}
\max_{\pi(\cdot|s)} \|\beta_s^\pi\|_{\mathrm{Lip}, \hat{\mu}_{n,s}} &= \max_{\pi(\cdot|s)} \max_{\substack{a \in \mathrm{supp}(\hat{\mu}(\cdot|s)), \\ \tilde{a} \neq a}} \frac{\frac{\pi(a|s)}{\pi_b(a|s)} - \frac{\pi(\tilde{a}|s)}{\pi_b(\tilde{a}|s)}}{c(a,\tilde{a})} \\
&= \max_{\substack{a \in \mathrm{supp}(\hat{\mu}(\cdot|s)), \\ \tilde{a} \neq a}} \frac{\frac{1}{\pi_b(a|s)}}{c(a,\tilde{a})} \\
&= \frac{1}{\min_{\substack{a \in \mathrm{supp}(\hat{\mu}(\cdot|s)), \\ \tilde{a} \neq a}} \pi_b(a \mid s) c(a,\tilde{a})} < \infty.
\end{aligned}$$

Therefore, the assumption in Theorem 6 is valid for sufficiently small $\rho_{n,s}$. Applying Dvoretzky–Kiefer–Wolfowitz inequality (Kosorok 2008) gives

$$\begin{aligned}
&\Pr\left( \max_a \frac{\hat{\mu}(a \mid s)}{\pi_b(a \mid s)} \le \frac{1+\gamma}{2\gamma} - \rho_{n,s} \max_{\pi(\cdot|s)} \|\beta_s^\pi\|_{\mathrm{Lip}, \hat{\mu}_s} \right) \\
=&\Pr\left( \max_a \frac{\hat{\mu}(a \mid s) - \pi_b(a \mid s)}{\pi_b(a \mid s)} \le \frac{1-\gamma}{2\gamma} - \rho_{n,s} \max_{\pi(\cdot|s)} \|\beta_s^\pi\|_{\mathrm{Lip}, \hat{\mu}_s} \right) \\
\ge&\Pr\left( \max_a \left[ \hat{\mu}(a \mid s) - \pi_b(a \mid s) \right] \le \delta_s' \left[ \frac{1-\gamma}{2\gamma} - \rho_{n,s} \max_{\pi(\cdot|s)} \|\beta_s^\pi\|_{\mathrm{Lip}, \hat{\mu}_s} \right] \right) \\
\ge&1 - \exp\left( -2 n_s \delta_s'^2 \left[ \frac{1-\gamma}{2\gamma} - \rho_{n,s} \max_{\pi(\cdot|s)} \|\beta_s^\pi\|_{\mathrm{Lip}, \hat{\mu}_s} \right]^2 \right) \\
\ge&1 - \exp\left( -2 n_s \delta_s'^2 \epsilon_s^2 \right).
\end{aligned}$$

□

*Proof of Theorem 5.* By exchanging the maximization over $v$ and $\pi$ in (13), we can reformulate this problem as the existence problem:

$$\begin{aligned}
\max_{v} \quad & (1-\gamma)\sum_{s} v(s)d_0(s) \\
\text{s.t.} \quad & \exists \pi, v(s) \le \sum_{a} \pi(a \mid s) r(s,a) + \gamma V(s),\ \forall s \in \mathcal{S}, \\
\text{where} \quad & V(s) := \min_{\mu \in \mathfrak{M}_{\hat{\mu}_n}(\rho_n)} \sum_{(a,s')} \mu(a,s' \mid s) v(s') \beta_s^{\pi}(a).
\end{aligned}$$

Or equivalently, we can reformulate this existence problem as

$$\begin{aligned}
\max_{v} \quad & (1-\gamma)\sum_{s} v(s)d_0(s) \\
\text{s.t.} \quad & v(s) \le \max_{\pi(\cdot|s)} \sum_{a} \pi(a \mid s) r(s,a) + \gamma V(s),\ \forall s \in \mathcal{S}, \\
\text{where} \quad & V(s) := \min_{\mu \in \mathfrak{M}_{\hat{\mu}_n}(\rho_n)} \sum_{(a,s')} \mu(a,s' \mid s) v(s') \beta_s^{\pi}(a).
\end{aligned} \tag{*}$$

Define the operator $\phi(v)$ as the right-hand side of the constraint (*), i.e.,

$$\phi(v) = \left( \max_{\pi(\cdot|s)} \sum_{a} \pi(a \mid s) r(s,a) + \gamma \min_{\mu \in \mathfrak{M}_{\hat{\mu}_n}(\rho_n)} \sum_{(a,s')} \mu(a,s' \mid s) v(s') \beta_s^{\pi}(a) \right)_{s \in \mathcal{S}}, \quad v \in \mathbb{R}^{|\mathcal{S}|}.$$

We claim that $\phi(v)$ is a contraction mapping: For any $v_1, v_2 \in \mathbb{R}_+^{|\mathcal{S}|}$,

$$\begin{aligned}
\phi(v_1)_s &= \max_{\pi(\cdot|s)} \sum_{a} \pi(a \mid s) r(s,a) + \gamma \min_{\mu \in \mathfrak{M}_{\hat{\mu}_n}(\rho_n)} \sum_{(a,s')} \mu(a,s' \mid s) v_1(s') \beta_s^{\pi}(a) \\
&\ge \max_{\pi(\cdot|s)} \sum_{a} \pi(a \mid s) r(s,a) + \gamma \min_{\mu \in \mathfrak{M}_{\hat{\mu}_n}(\rho_n)} \sum_{(a,s')} \mu(a,s' \mid s) v_2(s') \beta_s^{\pi}(a) \\
&\qquad + \gamma \min_{\mu \in \mathfrak{M}_{\hat{\mu}_n}(\rho_n)} \sum_{(a,s')} \mu(a,s' \mid s) [v_1(s') - v_2(s')] \beta_s^{\pi}(a) \\
&= \phi(v_2)_s + \gamma \min_{\mu \in \mathfrak{M}_{\hat{\mu}_n}(\rho_n), \pi(\cdot|s)} \sum_{(a,s')} \mu(a,s' \mid s) [v_1(s') - v_2(s')] \beta_s^{\pi}(a).
\end{aligned}$$

It follows that

$$\begin{aligned}
\phi(v_2)_s - \phi(v_1)_s &\le \gamma \max_{\mu \in \mathfrak{M}_{\hat{\mu}_n}(\rho_n), \pi(\cdot|s)} \sum_{(a,s')} \mu(a,s' \mid s) [v_2(s') - v_1(s')] \beta_s^{\pi}(a) \\
&\le \gamma \|v_2 - v_1\|_\infty \cdot \max_{\mu \in \mathfrak{M}_{\hat{\mu}_n}(\rho_n), \pi(\cdot|s)} \sum_{(a,s')} \mu(a,s' \mid s) \beta_s^{\pi}(a) \\
&\le \gamma \|v_2 - v_1\|_\infty \cdot \left[ \max_{\pi(\cdot|s)} \sum_{(a,s')} \hat{\mu}_n(a,s' \mid s) \beta_s^{\pi}(a) + \rho_{n,s} \max_{\pi(\cdot|s)} \|\beta_s^{\pi}\|_{\mathrm{Lip},\hat{\mu}_n} \right] \\
&= \gamma \|v_2 - v_1\|_\infty \cdot \left[ \max_{a} \frac{\hat{\mu}_n(a \mid s)}{\pi_b(a \mid s)} + \rho_{n,s} \max_{\pi(\cdot|s)} \|\beta_s^{\pi}\|_{\mathrm{Lip},\hat{\mu}_n} \right] \le \frac{1+\gamma}{2} \|v_2 - v_1\|_\infty.
\end{aligned}$$

We can exchange the role of $v_1$ and $v_2$ to show that $|\phi(v_2)_s - \phi(v_1)_s| \le \frac{1+\gamma}{2}\|v_1 - v_2\|_\infty$. Then applying Lemma EC.4 implies that $v^*$ and $\pi^*$ must be the optimal solution for (13). □

LEMMA EC.12. *For fixed $s \in \mathcal{S}$, we define*

$$\mathcal{F}_s := \left\{ (a,s') \mapsto \frac{v(s')\pi(a \mid s)}{\pi_b(a \mid s)}, \quad \pi(a \mid s) \text{ is deterministic} \right\}.$$

*Define* $\mathsf{d}(\tilde{v}, v) := |d_0^\top (I - \gamma P^{\mathsf{true}})^{-1} (\tilde{v} - v)|$, *then*

$$\mathcal{N}\left(\tfrac{1}{n}, \mathcal{F}_s, \mathsf{d}\right) \le 2|\mathcal{A}|nM.$$

*Proof of Lemma EC.12.* Define d and $\mathcal{H}$ following the setting in Lemma EC.7. Since $\pi(a \mid s)$ is deterministic, for fixed $s$, it can be expressed as

$$\pi(a \mid s) = 1\{a = a'\}, \quad \text{for some } a'.$$

It follows that

$$\mathcal{F}_s = \left\{ \bigcup_{a' \in \mathcal{A}} \left\{ a \mapsto \frac{1\{a = a'\}}{\pi_b(a \mid s)} \right\} \right\} \times \mathcal{H},$$

which implies the covering number of $\mathcal{F}_s$ can be upper bounded as $2|\mathcal{A}|nM$. □

As a result, Proposition EC.1 follows by applying the covering number argument similar as in Lemma EC.8. Next, we build the performance guarantee for robust batch reinforcement learning by applying the perturbation analysis on the Bellman operator and the generalization bound on Wasserstein DRO.

Proposition EC.1. *Fix* $s \in \mathcal{S}$ *and define*

$$\mathcal{F}_s := \left\{ (a, s') \mapsto \frac{v(s')\pi(a \mid s)}{\pi_b(a \mid s)}, \quad \pi(a \mid s) \text{ is deterministic} \right\}.$$

*Take* $\tau_s > 0$ *and* $\rho_{n,s} = \sqrt{\frac{2\tau_s}{n_s}} \operatorname{diam}(\mathcal{A} \times \mathcal{S})$,

$$\delta_s = \min_{(a,s') \in \operatorname{supp} d_{\pi_b}(\cdot,\cdot \mid s)} d_{\pi_b}(a, s' \mid s), \quad \Delta_s = \frac{1}{\left(\frac{1-2\delta_s}{1-\delta_s} \vee \frac{2\delta_s - 1}{\delta_s}\right)},$$

*then with probability at least* $1 - \alpha_s$, *where*

$$\alpha_s := \exp\left(-\tau_s + \log(2|\mathcal{A}|n_s M)\right) + \exp\left(-n_s\delta_s + \log(2|\mathcal{A}|n_s M)\right) + \exp\left(-n_s \log \Delta_s + \log\left(2|\mathcal{A}|n_s M\right)\right),$$

*simultaneously for every function* $f \in \mathcal{F}_s$, *it holds that*

$$\mathbb{E}_{z \sim \mathbb{P}_{\mathsf{true}}}[f(z)] \ge \min_{\mathbb{P}: W(\mathbb{P}, \mathbb{P}_n) \le \rho_{n,s}} \mathbb{E}_{\mathbb{P}}[f(z)] - \frac{6}{n_s}.$$

Now we give a proof of Theorem 6 by utilizing the perturbation analysis on the Bellman operator.

*Proof of Theorem 6.* Denote by $\mathcal{T}^{\mathsf{true}}$ the Bellman operator for the underlying true value function $v^{\mathsf{true}}$, and by $\mathcal{T}^*$ the Bellman operator for the robust value function $v^*$. Taking the union bound of the probability presented in Proposition EC.1 implies that, with probability at least $1 - \sum_s \alpha_s$, it holds that

$$\mathcal{T}^*[v] = \mathcal{T}^{\mathsf{true}}[v] + \epsilon_v, \quad \epsilon_v \le \epsilon_{n,s}.$$

Applying Lemma EC.6 implies

$$v^{\mathsf{true}} - v^* \ge -(I - \gamma P^{\mathsf{true}})^{-1} \epsilon_n.$$

Substituting $R^{\mathsf{true}}$ with $\sum_s d_0(s) v^{\mathsf{true}}(s)$ and $\mathcal{L}^*_{\hat{\mu}_n}(\rho_n)$ with $\sum_s d_0(s) v^*(s)$ completes the proof. □

## Appendix EC.6: Proofs of Appendix B

*Proof of Proposition 6.* Let $(\tilde{z}, z) \sim \Upsilon$ be taken from the set of joint distributions on $\mathcal{Z} \times \mathcal{Z}$ whose second marginal being $\mathbb{P}_n$, and denote by $\Upsilon_z$ the conditional distribution of $\tilde{z}$ given $z$. We have that

$$\begin{aligned}
\mathcal{R}_n(\rho; f) &= \sup_{\Upsilon} \inf_{\lambda \geq 0} \left\{\lambda\rho + \mathbb{E}_{(\tilde{z},z)\sim\Upsilon}[f(\tilde{z}) - f(z) - \lambda c(\tilde{z}, z)]\right\} \\
&\leq \inf_{\lambda \geq 0} \sup_{\Upsilon} \left\{\lambda\rho + \mathbb{E}_{(\tilde{z},z)\sim\Upsilon}[f(\tilde{z}) - f(z) - \lambda c(\tilde{z}, z)]\right\} \\
&= \inf_{\lambda \geq 0} \sup_{\Upsilon_z} \left\{\lambda\rho + \mathbb{E}_{z\sim\mathbb{P}_n}\left[\mathbb{E}_{\tilde{z}\sim\Upsilon_z}[f(\tilde{z}) - f(z) - \lambda c(\tilde{z}, z)|z]\right]\right\} \\
&= \inf_{\lambda \geq 0} \left\{\lambda\rho + \mathbb{E}_{z\sim\mathbb{P}_n}\left[\sup_{\Upsilon_z} \mathbb{E}_{\tilde{z}\sim\Upsilon_z}[f(\tilde{z}) - f(z) - \lambda c(\tilde{z}, z)|z]\right]\right\} \\
&= \inf_{\lambda \geq 0} \left\{\lambda\rho + \mathbb{E}_{z\sim\mathbb{P}_n}\left[\max_{\tilde{z}\in\mathcal{Z}}\{f(\tilde{z}) - f(z) - \lambda c(\tilde{z}, z)\}\right]\right\} \\
&\leq \rho \cdot \|f\|_{\mathrm{Lip},\mathbb{P}_n},
\end{aligned}$$

where the inequality holds due to Lagrangian relaxation; the second equality follows from the tower property of conditional expectation; the third equality follows from interchangeability principle (Shapiro et al. 2021); the fourth equality holds due to the fact that $\sup_{\Upsilon_z}$ is attained at a Dirac point mass; and the last equality follows by plugging in a feasible solution $\lambda = \|f\|_{\mathrm{Lip},\mathbb{P}_n}$.

To show the other direction, by definition of $\bar{\rho}$, there exists $\mathfrak{T} : \mathcal{Z} \to \mathcal{Z}$ such that $\mathfrak{T}(z) = z$ for $z \notin \mathcal{Z}_\infty$, $\frac{f(\mathfrak{T}(z)) - f(z)}{c(\mathfrak{T}(z), z)} = \mathfrak{l}_f(z) = \|f\|_{\mathrm{Lip},\mathbb{P}_n}$ for $z \in \mathcal{Z}_\infty$, and $\mathbb{E}_{\mathbb{P}_n}[c(\mathfrak{T}(z), z)] = \bar{\rho}$. Let $s = \rho/\bar{\rho}$. Then

$$\mathbb{E}_{\mathbb{P}_n}[(1-s)f(z) + sf(\mathfrak{T}(z))] - \mathbb{E}_{\mathbb{P}_n}[f] = s\mathbb{E}_{\mathbb{P}_n}[1\{z \in \mathcal{Z}_\infty\}(f(\mathfrak{T}(z)) - f(z))] = \rho\|f\|_{\mathrm{Lip},\mathbb{P}_n}.$$

Therefore, it holds that $\mathcal{R}_n(\rho; f) = \rho \cdot \|f\|_{\mathrm{Lip},\mathbb{P}_n}$. □

The proof of Proposition 8 relies on the finite-sample convergence of the Lipschitz norm with respect to $\mathbb{P}_n$ into the norm with respect to $\mathbb{P}_{\mathsf{true}}$, which is discussed in the following Lemma.

LEMMA EC.13. *Assume Assumption 4(II) holds. Let $\mathcal{F}$ be a collection of discrete functions $f : \mathcal{Z} \to \mathbb{R}$. With probability at least $1 - \exp\left(-n\log\left(\frac{1}{1-\delta}\right) + \log\mathcal{N}(\frac{1}{n}, \mathcal{F}, \mathsf{d})\right)$, the following relation holds uniformly for any discrete function $f \in \mathcal{F}$:*

$$\eta\|f\|_{\mathrm{Lip},\mathbb{P}_{\mathsf{true}}} \leq \|f\|_{\mathrm{Lip},\mathbb{P}_n} \leq \|f\|_{\mathrm{Lip},\mathbb{P}_{\mathsf{true}}}.$$

*Proof of Lemma EC.13.* By definition, it always holds that $\|f\|_{\mathrm{Lip},\mathbb{P}_n} \leq \|f\|_{\mathrm{Lip},\mathbb{P}_{\mathsf{true}}}$. On the other hand, for fixed $f \in \mathcal{F}$ define the event $\mathcal{E} = \{z : \mathfrak{l}_f(z) \geq \eta\|f\|_{\mathrm{Lip},\mathbb{P}_{\mathsf{true}}}\}$, then

$$\mathbb{P}\left\{\|f\|_{\mathrm{Lip},\mathbb{P}_n} \geq \eta\|f\|_{\mathrm{Lip},\mathbb{P}_{\mathsf{true}}}\right\} \geq 1 - \mathbb{P}\left\{z_1^n, \ldots, z_n^n \notin \mathcal{E}\right\} \geq 1 - (1-\delta)^n,$$

where $\operatorname{supp}\mathbb{P}_n := \{z_1^n, \ldots, z_n^n\}$. To bound the probability for a family of losses, using the definition of covering numbers, there exists a finite subset $\mathcal{F}_\epsilon \subseteq \mathcal{F}$ with $|\mathcal{F}_\epsilon| \leq \mathcal{N}(\epsilon, \mathcal{F}, \mathsf{d})$ so that for any $f \in \mathcal{F}$, there exists $\tilde{f} \in \mathcal{F}_\epsilon$ with $\mathsf{d}(f, \tilde{f}) \leq \epsilon$. It follows that

$$\begin{aligned}
&\mathbb{P}\left\{\forall f \in \mathcal{F}, \|f\|_{\mathrm{Lip},\mathbb{P}_n} \geq \eta\|f\|_{\mathrm{Lip},\mathbb{P}_{\mathsf{true}}} + 2\epsilon\right\} \\
\geq &\mathbb{P}\left\{\forall f \in \mathcal{F}_\epsilon, \|f\|_{\mathrm{Lip},\mathbb{P}_n} \geq \eta\|f\|_{\mathrm{Lip},\mathbb{P}_{\mathsf{true}}}\right\} \\
\geq &1 - \mathcal{N}(\epsilon, \mathcal{F}, \mathsf{d})(1-\delta)^n,
\end{aligned}$$

where the first inequality is based on the fact that $\mathsf{d}(f, \tilde{f}) \leq \epsilon$ implies $|\|f\|_{\mathrm{Lip},\mathbb{P}_n} - \|\tilde{f}\|_{\mathrm{Lip},\mathbb{P}_n}| \leq \epsilon$ and $|\|f\|_{\mathrm{Lip},\mathbb{P}_{\mathsf{true}}} - \|\tilde{f}\|_{\mathrm{Lip},\mathbb{P}_{\mathsf{true}}}| \leq \epsilon$, and the second inequality is by applying the union bound over $\mathcal{F}_\epsilon$. The proof is completed.

*Proof of Proposition 8.* Note that the underlying distribution $\mathbb{P}_{\mathsf{true}}$ satisfies the following transportation information inequality (Gao 2022, Definition 1):

$$\mathsf{W}(\mathbb{P},\mathbb{P}_{\mathsf{true}}) \le \mathrm{diam}(\mathcal{Z})\sqrt{2D_{\mathrm{KL}}(\mathbb{P}\|\mathbb{P}_{\mathsf{true}})},$$

where $D_{\mathrm{KL}}$ denotes the KL-divergence metric. Applying the concentration inequality presented in (Gao 2022, Theorem 1), with probability at least $1-e^{-\tau}$, for a single function $f\in\mathcal{F}$, it holds that

$$\mathbb{E}_{z\sim\mathbb{P}_{\mathsf{true}}}[f(z)] \le \mathbb{E}_{z\sim\mathbb{P}_n}[f(z)] + \mathcal{R}_{\mathbb{P}_{\mathsf{true}}}\left(\mathrm{diam}(\mathcal{Z})\sqrt{\frac{2\tau}{n}}; -f\right). \tag{EC.8}$$

Let $\mathcal{C}\left(\frac{1}{n},\mathcal{F},\mathsf{d}\right)$ be a $\frac{1}{n}$-cover of $\mathcal{F}$, then for any $f\in\mathcal{F}$, there exists $f'\in\mathcal{C}$ so that

$$\mathbb{E}_{\mathbb{P}_{\mathsf{true}}}[f] \le \mathbb{E}_{\mathbb{P}_{\mathsf{true}}}[f'] + \frac{1}{n}. \tag{EC.9}$$

Applying the union bound over all functions in $\mathcal{F}$ on the upper bound (EC.8), together with the relation (EC.9) and the upper bound in Theorem 6, implies that with probability at least $1-\mathcal{N}\left(\frac{1}{n},\mathcal{F},\mathsf{d}\right)e^{-\tau}$,

$$\begin{aligned}
\mathbb{E}_{z\sim\mathbb{P}_{\mathsf{true}}}[f(z)] &\le \mathbb{E}_{z\sim\mathbb{P}_n}[f'(z)] + \mathcal{R}_{\mathbb{P}_{\mathsf{true}}}\left(\mathrm{diam}(\mathcal{Z})\sqrt{\frac{2\tau}{n}}; -f'\right) + \frac{1}{n}\\
&\le \mathbb{E}_{z\sim\mathbb{P}_n}[f(z)] + \mathrm{diam}(\mathcal{Z})\sqrt{\frac{2\tau}{n}}\|f'\|_{\mathrm{Lip},\mathbb{P}_{\mathsf{true}}} + \frac{2}{n}\\
&\le \mathbb{E}_{z\sim\mathbb{P}_n}[f(z)] + \mathrm{diam}(\mathcal{Z})\sqrt{\frac{2\tau}{n}}\|f\|_{\mathrm{Lip},\mathbb{P}_{\mathsf{true}}} + \frac{3}{n},
\end{aligned} \tag{EC.10}$$

where the last inequality above is because

$$\big|\|f'\|_{\mathrm{Lip},\mathbb{P}_{\mathsf{true}}} - \|f\|_{\mathrm{Lip},\mathbb{P}_{\mathsf{true}}}\big| \le \|f-f'\|_{\mathbb{P}_{\mathsf{true}},\infty} \le \mathsf{d}(f,f') \le \frac{1}{n}.$$

By Lemma EC.13, with probability at least $1-\mathcal{N}\left(\frac{1}{n},\mathcal{F},\mathsf{d}\right)e^{-n\log\left(\frac{1}{1-\delta}\right)}$,

$$\mathrm{diam}(\mathcal{Z})\sqrt{\frac{2\tau}{n}}\|f\|_{\mathrm{Lip},\mathbb{P}_{\mathsf{true}}} \le \eta\rho_n\|f\|_{\mathrm{Lip},\mathbb{P}_n}. \tag{EC.11}$$

Applying Proposition 7 with $c=0$ implies that, with probability at least

$$1-\exp\left[-n\log\left(\frac{1}{\left(\frac{1-2\delta}{1-\delta}\vee\frac{2\delta-1}{\delta}\right)}\right)+\log\mathcal{N}\left(\frac{1}{n},\mathcal{F},\mathsf{d}\right)\right],$$

we have

$$\eta\rho_n\|f\|_{\mathrm{Lip},\mathbb{P}_n} \le \mathcal{R}_n(\rho_n;f) + \frac{3}{n}. \tag{EC.12}$$

Combining (EC.10)-(EC.12) completes the proof. □

*Proof of Proposition 9.* Since $\rho$ is sufficiently small, applying Theorem 6 gives

$$U_n(\rho;f) - U^*(\rho;f) = \left(\mathbb{E}_{\mathbb{P}_n}[f(z)] - \mathbb{E}_{\mathbb{P}_{\mathsf{true}}}[f(z)]\right) + \rho\left(\|f\|_{\mathrm{Lip},\mathbb{P}_n} - \|f\|_{\mathrm{Lip},\mathbb{P}_{\mathsf{true}}}\right).$$

On the one hand, applying Lemma EC.5 in Gao et al. (2020) implies that with probability at least $1-e^{-\tau}$, it holds that

$$\mathbb{E}_{\mathbb{P}_n}[f(z)] - \mathbb{E}_{\mathbb{P}_{\text{true}}}[f(z)] \leq 2\mathbb{E}_{\otimes}[\mathfrak{R}(\mathcal{F})] + M\sqrt{\frac{\tau}{2n}}.$$

This, together with the relation $\|f\|_{\text{Lip},\mathbb{P}_{\text{true}}} \leq \|f\|_{\text{Lip},\mathbb{P}_n}, \forall f \in \mathcal{F}$ implies that

$$U_n(\rho; f) - U^*(\rho; f) \leq 2\mathbb{E}_{\otimes}[\mathfrak{R}(\mathcal{F})] + M\sqrt{\frac{\tau}{2n}}. \tag{EC.13}$$

On the other hand, applying Lemma EC.5 in Gao et al. (2020) again implies that with probability at least $1-e^{-\tau}$, it holds that

$$\mathbb{E}_{\mathbb{P}_n}[f(z)] - \mathbb{E}_{\mathbb{P}_{\text{true}}}[f(z)] \geq -\left(2\mathbb{E}_{\otimes}[\mathfrak{R}(\mathcal{F})] + M\sqrt{\frac{\tau}{2n}}\right).$$

By Lemma EC.13, the relation $\|f\|_{\text{Lip},\mathbb{P}_{\text{true}}} = \|f\|_{\text{Lip},\mathbb{P}_n}, \forall f \in \mathcal{F}$ holds with probability at least $1 - \mathcal{N}\left(\frac{1}{n}, \mathcal{F}, \mathrm{d}\right) e^{-n\log\left(\frac{1}{1-\delta}\right)}$. Therefore, we conclude that with probability at least $1-\mathcal{N}\left(\frac{1}{n}, \mathcal{F}, \mathrm{d}\right) e^{-n\log\left(\frac{1}{1-\delta}\right)} - e^{-\tau}$, it holds that

$$U^*(\rho; f) - U_n(\rho; f) \leq 2\mathbb{E}_{\otimes}[\mathfrak{R}(\mathcal{F})] + M\sqrt{\frac{\tau}{2n}}. \tag{EC.14}$$

Taking the union bound in (EC.13) and (EC.14) completes the proof.

□